\documentclass{article}

\usepackage[final]{corl_2026} % Uncomment for the camera-ready ``final'' version.

\usepackage{graphicx}
\usepackage{amsmath}
\usepackage{amssymb}
\usepackage{amsthm}
\usepackage{array}
\usepackage{multirow}
\usepackage[capitalise]{cleveref}
\crefname{appendix}{Appendix}{Appendices}
\Crefname{appendix}{Appendix}{Appendices}
\usepackage{tabularx}
\usepackage{booktabs}
\usepackage{caption}

\newcolumntype{Y}{>{\centering\arraybackslash}X}

\title{VertexCBF: Improving Neural Control Barrier Functions via Vertex-Restricted Control Search}

\author{
Bojan Derajić$^{1}$,
Sebastian Bernhard$^{2}$,
Wolfgang Hönig$^{1,3}$ \\[4pt]
$^{1}$Technical University of Berlin, Germany \\
$^{2}$Technical University of Applied Sciences Augsburg, Germany \\
$^{3}$Robotics Institute Germany (RIG), Germany \\[6pt]
Contact: \texttt{bojan.derajic@campus.tu-berlin.de}
}

\begin{document}
\maketitle

%===============================================================================

\begin{abstract}
As the number of autonomous robots continues to grow, safety becomes increasingly important. Control barrier functions (CBFs) provide a theoretically grounded framework for ensuring safety, but existing design methods often face limitations in effectiveness, scalability, or interpretability, and may result in overly conservative safe sets. In this paper, we propose \emph{VertexCBF}, a framework for learning neural CBFs in a scalable, systematic, and explainable way. We approximate the stationary Hamilton--Jacobi value function using a neural network trained via a combination of physics-informed and sparsely supervised learning. By exploiting control-affine dynamics and a convex polytope control set, under which the Hamiltonian is maximized at the control vertices, we efficiently generate supervision points via GPU-parallel vertex-restricted tree search, while a residual architecture guarantees that the learned CBF is never larger than the specified constraint function. We evaluate the method on 15 systems and compare it against relevant baselines, showing that it reliably recovers large safe sets where the baselines are conservative or fail completely. In addition, we perform a hardware experiment in which a mobile robot safely avoids pedestrians using a neural CBF trained with our method.
\end{abstract}


%===============================================================================

\section{Introduction}
\label{sec:intro}

In recent years, robotics applications have expanded rapidly across many areas of society. Many of these advances are driven by learning-based algorithms and involve direct interaction between autonomous robots and their environments. However, ensuring safe and reliable operation in a systematic and explainable manner remains a key challenge in robot deployment.

One well-established approach is safety filtering, where the nominal control input is modified at runtime to preserve safety \cite{hsu_safety_2024}. In particular, control barrier functions (CBFs) provide a theoretically grounded and explainable way to design safety filters, and are especially efficient for control-affine systems \cite{ames_control_2014, ames_control_2019}. Nevertheless, designing CBFs for systems with state and input constraints is notoriously difficult and often results in a compromise between performance and safety \cite{wabersich_data-driven_2023}.

In this paper, we propose \emph{VertexCBF}, a learning-based framework for constructing neural CBFs that approximate the optimal safe set characterized by Hamilton--Jacobi (HJ) reachability. Specifically, we use a neural approximation of the stationary HJ safety value function as a CBF and train it using a combination of physics-informed and sparsely supervised losses. Following a residual parametrization, the learned function is constrained to be no larger than the specified constraint function, ensuring that learning cannot enlarge the safe set beyond the original state constraint. The key idea is to exploit the standard structure used by CBF-QP filters, namely control-affine dynamics and a convex polytope control set. Under these assumptions, the continuous-time Hamiltonian is maximized at a control vertex. We therefore generate supervision labels by solving finite-horizon optimization problems restricted to control vertices, enabling efficient GPU-parallel tree search while avoiding numerical PDE solvers and sampling-based trajectory optimization.

The main contributions of this paper are summarized as follows:
\begin{itemize}
    \item We introduce a systematic method for learning neural CBFs that approximate the optimal safe set. It is grounded in HJ reachability, combines physics-informed and supervised learning, and exploits control-affine dynamics with polytope input constraints to generate supervision points through highly parallel vertex-restricted optimization.
    \item We evaluate the proposed method on 15 dynamical systems and against relevant baselines. The results show that our approach consistently recovers large and reliable safe sets where the baselines are conservative or fail completely. We further demonstrate the method in a hardware experiment, where a mobile robot safely navigates among pedestrians.
\end{itemize}

%===============================================================================

\section{Related Work}
\label{sec:related_work}

Learning-based approaches to safe robot control have attracted considerable attention due to their flexibility and scalability \cite{wabersich_data-driven_2023, dawson_safe_2023}. Particularly relevant are methods that combine machine learning with CBFs and HJ reachability. Early work on learning CBFs used support vector machines \cite{srinivasan_synthesis_2020} and neural networks \cite{long_learning_2021}, while more recent approaches employ expressive representations such as Gaussian splats to construct distance-based CBFs \cite{chen_control_2025}. However, these methods often struggle with higher-order systems and limited actuation, where distance-based formulations become inadequate \cite{hsu_safety_2024}. Extensions that account for control limits \cite{liu_safe_2022, xiao_barriernet_2023} improve feasibility but typically yield conservative or suboptimal safe sets.

The maximal safe set can be characterized through HJ reachability, but solving the corresponding PDEs is generally intractable for real-time applications. Methods in \cite{derajic_orn-cbf_2025, lin_one_2026, derajic_cn-cbf_2026, derajic_residual_2025} learn approximations of the HJ value function but rely on numerically generated training data, which limits scalability. DeepReach \cite{bansal_deepreach_2021} improves scalability through self-supervised learning, with further refinements addressing boundary conditions and uncertainty \cite{singh_exact_2025, lin_verification_2024}. Nevertheless, these methods rely on carefully designed curriculum training procedures that can be difficult to tune and may exhibit instability.

The method in \cite{manda_learning_2025} trains neural CBFs by minimizing a loss derived from the stationary HJB variational inequality (HJB-VI). However, this objective alone admits multiple trivial or suboptimal solutions, leading to poor results for many relevant systems in practice, as we also demonstrate experimentally. This motivates augmenting physics-informed learning with additional supervision to guide optimization toward meaningful solutions.

Our approach combines physics-informed learning with efficiently generated, high-quality supervision by exploiting control-affine dynamics and convex control constraints. The closest work, \cite{feng_bridging_2025}, similarly combines PDE-based and data-driven losses, but targets the general time-dependent HJ reachability problem and relies on full-control sampling-based MPC for supervision, together with pretraining, curriculum training, and iterative refinement. In contrast, we solve the infinite-horizon stationary HJB-VI, eliminating the time variable and the need for curriculum learning, and assume control-affine dynamics with a convex polytope control set. These assumptions replace full-control trajectory optimization with a vertex-restricted tree search, while the residual parametrization makes pretraining unnecessary.

%===============================================================================

\section{Preliminaries and Problem Formulation}
\label{sec:preliminaries}

We consider control-affine system dynamics
\begin{equation} \label{eq:sys_dyn_c}
    \dot{x} = \phi(x, u) = f(x) + g(x)u,
\end{equation}
where $x \in \mathcal{X} \subseteq \mathbb{R}^{n_x}$ is the system state and $u \in \mathcal{U} \subset \mathbb{R}^{n_u}$ is the control input. We assume that $\mathcal{U}$ is a compact convex polytope, with the set of control vertices denoted by $\mathcal{V} = \operatorname{vert}(\mathcal{U})$. The system is required to remain within the constraint set $\mathcal C = \{ x : c(x) \geq 0 \}$, where $c : \mathcal{X} \to \mathbb{R}$ is a continuous constraint function.

Given the initial time $t = 0$, the infinite-horizon safety value function $V : \mathcal{X} \to \mathbb{R}$ is defined as
\begin{equation} \label{eq:safety_vf_c}
    V(x) = \sup_{u(\cdot) \in \mathcal{U}} \inf_{t \geq 0} c\left( x(t) \right),
\end{equation}
where $x(0)=x$ and $x(t)$ evolves according to \cref{eq:sys_dyn_c}. The value $V(x)$ is the largest worst-case constraint value that can be maintained from $x$, and the set ${\mathcal{S} = \{ x : V(x) \geq 0 \}}$ is the maximal controlled invariant subset of $\mathcal{C}$, termed the \textit{maximal safe set}\footnote{Also known as the \textit{viability kernel}.} \cite{bansal_hamilton-jacobi_2017, hsu_safety_2021}. We assume throughout the paper that the supremum in \cref{eq:safety_vf_c} is finite for every $x \in \mathcal{X}$, so that $V$ is well defined. This holds in all our experiments, since $\mathcal{U}$ is compact and $c$ is bounded on $\mathcal{X}$.

A common approach to enforcing safety is through CBFs. A continuously differentiable function $h : \mathcal{X} \to \mathbb{R}$ is a CBF if there exists an extended class-$\mathcal{K}$ function $\alpha$ such that, for all $x \in \mathcal{X}$,
\begin{equation} \label{eq:cbf_constr}
    \sup_{u \in \mathcal{U}} \nabla h(x)^\top \phi(x,u) + \alpha(h(x)) \geq 0.
\end{equation}
For an arbitrary nominal control $u_{\mathrm{nom}}$, the safe control $u^*$ is obtained by solving the following quadratic program (QP):
\begin{equation} \label{eq:cbf_qp}
\begin{aligned}
    u^* = {} &\operatorname*{argmin}_{u \in \mathcal{U}} \frac{1}{2} \lVert u - u_{\mathrm{nom}} \rVert^2 \\
    & \text{s.t.} \quad L_f h(x) + L_g h(x)u + \alpha(h(x)) \geq 0,
\end{aligned}
\end{equation}
where $L_f$ and $L_g$ are the Lie derivatives of $h$ with respect to $f$ and $g$, respectively \cite{ames_control_2014}. If the QP is feasible, by Nagumo's theorem \cite{nagumo_uber_1942} this CBF-QP safety filter renders the set $\mathcal{S}_h = \left\{ x : h(x) \geq 0 \right\}$ forward invariant. The central problem of this paper is to establish a systematic method for designing a CBF $h$ such that $\mathcal{S}_h \approx \mathcal{S}$. In the next section, we describe the proposed method.

%===============================================================================

\section{Methodology}
\label{sec:methodology}

VertexCBF aims to design a CBF whose resulting safe set coincides with the maximal safe set $\mathcal{S}$. Similar to \cite{derajic_orn-cbf_2025, lin_one_2026, derajic_cn-cbf_2026}, we use a neural approximation $V_\Theta$ of the safety value function \cref{eq:safety_vf_c} as a CBF, i.e., $h = V_\Theta$. The model parameters $\Theta$ are trained by combining a PDE-based loss with sparse supervision, as in \cite{feng_bridging_2025}. In contrast to that work, our approach exploits the assumptions on the system dynamics and control set to obtain more effective supervision.

The safety value function \cref{eq:safety_vf_c} is a solution of the stationary HJB-VI \cite{manda_learning_2025} and necessarily satisfies
\begin{equation} \label{eq:stationary_hjb_vi}
    \min \left\{ c(x) - V(x), \; H \left(x, \nabla V(x) \right) \right\} = 0,
\end{equation}
where the Hamiltonian is
\begin{equation} \label{eq:hamiltonian}
    H(x, p) = \max_{u \in \mathcal{U}} p^\top \phi(x,u) = p^\top f(x) + \max_{u \in \mathcal{U}} p^\top g(x) u.
\end{equation}
Since the second term is linear in $u$ and $\mathcal{U}$ is a compact convex polytope, the maximum is attained at a vertex $v \in \mathcal{V}$. Therefore, we can rewrite the stationary HJB-VI as
\begin{equation} \label{eq:stationary_hjb_vi_vert}
    \min \left\{ c(x) - V(x), \; \max_{u \in \mathcal{V}} \nabla V(x)^\top \phi(x, u) \right\} = 0.
\end{equation}
This variational inequality could be used as a PDE loss and optimized as a single self-supervised objective when training a neural approximation $V_\Theta(x)$, similar to \cite{manda_learning_2025}. However, \cref{eq:stationary_hjb_vi_vert} generally does not have a unique solution. In fact, any constant value function $V(x) = a$ such that ${a \leq c(x)}$ for all $x \in \mathcal{X}$ satisfies the stationary HJB-VI. Moreover, gradient-based optimization of the model parameters is prone to local optima, further undermining training performance. An effective alternative is to generate a (potentially sparse) dataset of state-value pairs and add a supervised loss term that guides training toward the true optimal solution. Nevertheless, generating such a dataset with existing methods is often ineffective without a strong nominal policy.

\subsection{Time Discretization of the Vertex-Restricted Problem}
\label{subsec:discrete_safety_vf}

The restriction to control vertices is already made in continuous time in \cref{eq:stationary_hjb_vi_vert}, and we now discretize that problem in time. For a time step $\Delta t > 0$, we define the forward-Euler dynamics map
\begin{equation} \label{eq:euler_map}
    x_{k+1} = F(x_k, u_k) = x_k + \Delta t \, \phi(x_k, u_k).
\end{equation}

Applying this map to the vertex-restricted problem, we define the discrete-time safety value function $V_{\Delta t} : \mathcal{X} \to \mathbb{R}$ as
\begin{equation} \label{eq:safety_vf_d}
    V_{\Delta t}(x) = \sup_{\{u_i\}_{i=0}^{\infty} \in \mathcal V} \inf_{k\ge 0} c(x_k),
\end{equation}
where $x_0 = x$. This is the infinite-horizon safety value function of the Euler-discretized system with input sequences restricted to the control vertices. It satisfies the Bellman equation
\begin{equation} \label{eq:bellman_eq}
    V_{\Delta t}(x)
    =
    \min\left\{
        c(x),
        \;
        \max_{u\in\mathcal V}
        V_{\Delta t}\bigl(F(x, u)\bigr)
    \right\}.
\end{equation}

The order of these two steps is important. We do not restrict a full-control discrete-time Bellman operator to $\mathcal{V}$, which would not be exact, because $u \mapsto V_{\Delta t}(F(x,u))$ is nonlinear in $u$. We instead restrict the continuous-time Hamiltonian, where the maximization is linear in the control, and only then apply the forward-Euler map. Therefore, $V_{\Delta t}$ approximates the vertex-restricted continuous-time problem, not the full-control discrete-time value function. Wherever $V$ is differentiable, an optimal control can be chosen from $\mathcal{V}$, and such value functions are differentiable almost everywhere. \cref{eq:bellman_eq} is exact for Euler-discretized vertex-restricted system, but it is not the Bellman equation of the full-control discrete-time system and only approximates the original continuous-time problem.

The discrete-time solution is not guaranteed to be obtained from \cref{eq:bellman_eq} by standard value iteration, because the Bellman operator is not a contraction and the usual contraction-based convergence guarantees therefore do not apply \cite{fisac_bridging_2019, so_how_2024}. One possible remedy is to introduce a discounted value function, as in \cite{akametalu_minimum_2024, fisac_bridging_2019}, and gradually increase the discount factor to recover the optimal solution. Restricting the inner maximization to control vertices is an additional constraint and may yield a suboptimal solution for the discretized system. Our primary objective, however, is to approximate the safety values of the continuous-time system, and we reformulate \cref{eq:bellman_eq} as
\begin{equation} \label{eq:discrete_hjb_vi}
    \min\left\{
        c(x)-V_{\Delta t}(x),
        \;
        \frac{
        \max_{u \in\mathcal V}
        V_{\Delta t}\bigl(F(x, u)\bigr)
        -
        V_{\Delta t}(x)
        }{\Delta t}
    \right\}
    =
    0,
\end{equation}
where the second term is a finite-difference approximation of the Hamiltonian in \cref{eq:stationary_hjb_vi}. In \cref{app:first_order_consistency}, we show that the vertex-restricted Euler Bellman operator \cref{eq:bellman_eq} is a first-order consistent local approximation of the continuous-time HJB-VI.

\subsection{Finite-Horizon Search over Vertex-Restricted Controls}
\label{sec:finite_horizon_search} 

Obtaining the infinite-horizon vertex-restricted value $V_{\Delta t}$ may still be impractical for many systems of interest. Moreover, we compute each label independently over a sparsely sampled set of states, so that it does not depend on values at other states, unlike grid-based numerical methods. We therefore consider the finite-horizon approximation
\begin{equation} \label{eq:finite_horizon_value}
    V_{\Delta t}^{K}(x) =
    \max_{u_0,\ldots,u_{K-1} \in \mathcal V} \,
    \min_{0\le k\le K}
    c(x_k),
\end{equation}
where $x_0 = x$ and $K$ is the horizon length. Since inputs are restricted to control vertices, the search space is reduced to a typically small set of control vectors. Instead of gradient-based or sampling-based optimization, we can therefore perform a tree search over the finite-horizon trajectories. If $M = |\mathcal{V}|$ denotes the number of control vertices, the branching factor is $M$ and the tree depth is $K$, resulting in $M^K$ possible trajectories. For low-dimensional systems, a full tree search finds the optimal solution exactly. However, when $M$ or $K$ is large, we use approximate tree-search variants such as beam search, stochastic beam search, or branch-and-bound, summarized in \cref{app:overview_tree_search}.

A non-complete search introduces an additional approximation by evaluating only a subset of the full tree. Let $V_{\Delta t}^{K, B}(x)$ denote the value returned by a tree-search method with beam width $B$, as described in \cref{app:overview_tree_search}. Since beam search optimizes over a subset of all vertex-control sequences,
\begin{equation}
    V_{\Delta t}^{K, B}(x)
    \le
    V_{\Delta t}^{K}(x),
\end{equation}
with equality if the retained beam contains an optimal sequence. Width-limited labels are therefore lower bounds on the full vertex-restricted search. This is the conservative direction, because a lower label can only shrink the safe set. The influence of beam-search hyperparameters is analyzed in \cref{app:ablation_beam_search}. The total discrepancy between the generated label and the continuous infinite-horizon safety value can therefore be decomposed conceptually as
\begin{equation}
    \bigl|
    V_{\Delta t}^{K, B}(x)
    -
    V(x)
    \bigr|
    \le\;
    \underbrace{
    \bigl|
    V_{\Delta t}(x)-V(x)
    \bigr|
    }_{\substack{\text{discretization and} \\ \text{vertex-restriction error}}}
    +
    \underbrace{
    \bigl|
    V_{\Delta t}^{K}(x)-V_{\Delta t}(x)
    \bigr|
    }_{\text{horizon truncation error}}
    +
    \underbrace{
    \bigl|
    V_{\Delta t}^{K, B}(x)-V_{\Delta t}^{K}(x)
    \bigr|
    }_{\text{beam-pruning error}} .
\label{eq:error_decomposition}
\end{equation}

The first term collects the error of the forward-Euler discretization and the error of holding one vertex control over each interval of length $\Delta t$, and a smaller $\Delta t$ reduces both (\cref{app:ablation_dt}). The second term is caused by truncating the horizon after $K$ steps. It vanishes under the assumption that an optimal sequence of \cref{eq:finite_horizon_value} can be extended by further vertex controls that keep $c(x_k) \geq V_{\Delta t}^{K}(x)$ for all $k > K$, so that the safety level attained within the horizon is maintained forever. Such a continuation gives $V_{\Delta t}(x) \geq V_{\Delta t}^{K}(x)$, while the opposite inequality always holds. The assumption is plausible when the horizon is long enough to reach a region where the attained safety level can be maintained. It fails for a short horizon, where the label overestimates the true value. The third term vanishes under a complete tree search, and in practice we reduce it with stochastic beam search or branch-and-bound.

\paragraph{Scope of the vertex restriction.}
Restricting the search to vertices does not by itself improve the finite-horizon optimum, since \cref{eq:finite_horizon_value} is a constrained version of the same problem over the full control set. The benefit is effectiveness under a finite computational budget. An optimal continuous-time control can be chosen from $\mathcal{V}$ at almost every state, so the tree search operates on a small set that already contains these controls. A sampling-based method such as MPPI instead allocates much of its budget to interior controls that are rarely optimal. This relies on control-affine dynamics, a polytope control set, and a sufficiently small $\Delta t$. The approximation can become conservative or suboptimal for a large $\Delta t$, for a short horizon $K$, or for a small beam width $B$ when $M$ or $K$ is large. We therefore do not claim that vertex-restricted search always outperforms full-control optimization.

\subsection{Model Architecture and Training Procedure}
\label{subsec:model_training}

Following \cite{manda_learning_2025, derajic_orn-cbf_2025, derajic_cn-cbf_2026}, we do not approximate the value function directly, but instead approximate its non-negative residual $r_{\Theta}$ with respect to the constraint function:
\begin{equation} \label{eq:residual_param}
    V_{\Theta}(x) = c(x) - r_{\Theta}(x), \quad r_{\Theta}(x) \geq 0, \;\; \forall x \in \mathcal{X}.
\end{equation}
Since $V_{\Theta}(x) \leq c(x)$, the learned safe set $\hat{\mathcal{S}} = \{x : V_{\Theta}(x) \geq 0\}$ is always contained in the constraint set $\mathcal{C}$. Moreover, this parametrization makes trivial constant solutions less natural, because a constant $V_\Theta(x)=a$ would require $r_\Theta(x)=c(x)-a$ rather than a constant network output.

As the residual function, we use a multi-layer perceptron (MLP) with sinusoidal hidden activations \cite{sitzmann_implicit_2020} and a softplus output that enforces $r_\Theta \geq 0$ while remaining smooth. Non-periodic inputs are linearly rescaled to $[-1, 1]$ to keep the input features on a common scale. Each periodic state $\theta$ is replaced by the pair $\left(\cos(\theta), \sin(\theta)\right)$, so that periodicity is built in by construction.

\paragraph{Loss function.}
The model is trained by minimizing a convex combination of the physics-informed PDE loss and the data loss:
\begin{equation} \label{eq:total_loss}
    \mathcal{L}(\Theta) = \lambda \, \mathcal{L}_{\mathrm{PDE}}(\Theta) + (1-\lambda) \, \mathcal{L}_{\mathrm{data}}(\Theta), \quad \lambda \in [0, 1].
\end{equation}
The PDE term penalizes pointwise violations of the stationary HJB-VI \cref{eq:stationary_hjb_vi_vert} on collocation states $\{x_i\}_{i=1}^{N_\mathrm{PDE}}$ sampled uniformly from $\mathcal{X}$. Substituting \cref{eq:residual_param} into \cref{eq:stationary_hjb_vi_vert} simplifies the first term because $c(x)-V_\Theta(x) = r_\Theta(x)$, yielding
\begin{equation} \label{eq:pde_loss}
    \mathcal{L}_{\mathrm{PDE}}(\Theta) = \frac{1}{N_\mathrm{PDE}} \sum_{i=1}^{N_\mathrm{PDE}} \min \left\{ r_\Theta(x_i), \; H \left( x_i, \nabla V_\Theta(x_i) \right) \right\}^2.
\end{equation}
The state gradient $\nabla V_\Theta$ is obtained by automatic differentiation, and the Hamiltonian is evaluated exactly over the finite vertex set $\mathcal{V}$. The minimum in \cref{eq:pde_loss} is non-differentiable where its two arguments cross, near the boundary of the safe set. We also evaluated a smooth minimum, but observed no training instability and no consistent difference in the results, so all reported results use the exact minimum. The data term is a mean-squared error against labels generated by vertex-restricted tree search at supervision states $\{(x_j, y_j)\}_{j=1}^{N_\mathrm{data}}$, with $y_j = V_{\Delta t}^{K, B}(x_j)$:
\begin{equation} \label{eq:data_loss}
    \mathcal{L}_{\mathrm{data}}(\Theta) = \frac{1}{N_\mathrm{data}} \sum_{j=1}^{N_\mathrm{data}} \left( c(x_j) - r_\Theta(x_j) - y_j \right)^2.
\end{equation}
The two terms are complementary. The term $\mathcal{L}_{\mathrm{PDE}}$ enforces local self-consistency on densely sampled unlabeled states, while $\mathcal{L}_{\mathrm{data}}$ anchors the solution to non-trivial values and steers optimization away from spurious local minima of the PDE objective alone.
	
%===============================================================================

\section{Experimental Results}
\label{sec:results}

We evaluate the proposed method on 15 systems with varying dynamics, state and control dimensions, and safety constraints\footnote{All numerical experiments are conducted on a single NVIDIA RTX 3090 24GB GPU. The code is available at \url{https://github.com/bojan-derajic/vertexcbf}.}. This broad evaluation is important because learning-based CBF methods can be highly system-dependent. A method may succeed on one system yet fail on another of similar dimension with a more complex value function. We compare against two baselines. The first, \emph{PDE-only}, follows \cite{manda_learning_2025} and uses only the PDE-based loss, without data supervision. Since the methods differ only in supervision data, we denote it by \emph{ND} (no data) in the tables. The second is motivated by \cite{feng_bridging_2025} and uses full-control data (\emph{FCD}) generated by the MPPI technique. It reproduces only the data-generation step of \cite{feng_bridging_2025}, and not the pretraining, curriculum training and iterative refinement of the complete method. Therefore, the comparison isolates the effect of the supervision data. Our method, VertexCBF, uses vertex-restricted control data (\emph{VRCD}) generated via a tree search method. Details on the dynamics, data generation, training, and validation are given in \cref{app:details_num_exp}. All methods use the same model architecture, hyperparameters, training settings, and validation protocol, and differ only in supervision data. FCD uses the same horizon length and step size as our tree search, and a number of sampled trajectories equal to the beam width. For FCD, we run five MPPI iterations, using the best trajectory of each iteration as the nominal trajectory for the next \cite{feng_bridging_2025}. The influence of the PDE-loss weight on the result is studied in \cref{app:ablation_pde_weight}.

\paragraph{Validation procedure.}
Since there are no formal convergence or error-bound guarantees, we emphasize the importance of post-training validation. For each learned CBF $V_\Theta$, we draw $N_{\mathrm{valid}}$ initial states $\{x_i\}_{i=1}^{N_{\mathrm{valid}}}$ from the state-space bounding box $\mathcal{X} = [x_{\min}, x_{\max}]$ by stratified rejection sampling, so that roughly half lie in the predicted safe set $\hat{\mathcal{S}}$ and the other half lie in its complement. Balancing the samples reduces the variance of the validation metrics when the predicted safe set is much smaller than its complement. From each $x_i$, we execute the closed-loop dynamics forward in time for a period $T_{\mathrm{valid}}$ under the Hamiltonian-maximizing policy
\begin{equation}
  u^\star(x) = \arg\max_{u \in \mathcal{V}}\;
    \nabla V_\Theta(x)^\top \left( f(x) + g(x)u \right).
  \label{eq:hj_policy}
\end{equation}
An initial state $x_i$ is declared \emph{validated safe} if the entire rolled-out trajectory remains within $\mathcal{C}$. This estimates the forward-invariant subset of $\mathcal{C}$ induced by the learned CBF $V_\Theta$.

\paragraph{Validation metrics.}
We report the false-safe rate $\rho_{\mathrm{FS}}$, false-unsafe rate $\rho_{\mathrm{FU}}$, and effective safe volume $\eta_{\mathrm{eff}}$. The false-safe rate is the fraction of states in $\hat{\mathcal{S}}$ whose validated trajectories leave $\mathcal{C}$, and it measures the trustworthiness of the learned certificate. The false-unsafe rate is the fraction of states outside $\hat{\mathcal{S}}$ whose trajectories remain safe, and it measures conservatism. Finally,
\begin{equation}
  \eta_{\mathrm{eff}}
  =
  \frac{\operatorname{vol}(\hat{\mathcal{S}})}
       {\operatorname{vol}(\mathcal{X})}
  \bigl(1-\rho_{\mathrm{FS}}\bigr),
\end{equation}
measures the fraction of the state space that the neural CBF both predicts and validates as safe. The ratio $\operatorname{vol}(\hat{\mathcal{S}})/\operatorname{vol}(\mathcal{X})$ is estimated as the fraction of states drawn uniformly from $\mathcal{X}$, before the rejection step, for which $V_\Theta \geq 0$. Thus, $\eta_{\mathrm{eff}}$ captures the practical operating region of the safety filter, while $\rho_{\mathrm{FS}}$ indicates whether that region is reliable. A good method should have small $\rho_{\mathrm{FS}}$ and large $\eta_{\mathrm{eff}}$, while $\rho_{\mathrm{FU}}$ indicates how much safe behavior is unnecessarily rejected. These metrics rely only on rollouts, so they are available for every system. For the seven systems with at most four states, we additionally compare against the ground truth in \cref{app:iou}.

The results in \cref{tab:num_exp_results} show that PDE-only learning is unreliable across systems. It performs well on some simpler systems, such as double integrators, but fails completely on several others. FCD often reduces false-safe errors, but it tends to be conservative or ineffective because the full-control MPPI supervision data are less informative. In contrast, VRCD generally achieves low false-safe rates and competitive false-unsafe rates, while maintaining the highest or tied-highest effective safe volume. These results demonstrate the benefits of the proposed method and the importance of evaluating multiple complementary metrics. Each configuration is trained with five seeds. The false-safe rate is undefined when the predicted safe set is empty, so such seeds are excluded from its average and contribute $\eta_\mathrm{eff} = 0$. All three methods share the residual parametrization \cref{eq:residual_param}, whose contribution is isolated in \cref{app:ablation_res_arch}.

\begin{table*}[t]
  \centering
  \caption{CBF validation results, reported as mean $\pm$ standard deviation over five seeds. A dash marks systems whose predicted safe set is empty for every seed, and a missing standard deviation indicates that $\rho_\mathrm{FS}$ is defined for a single seed only.}
  \label{tab:num_exp_results}
  \scriptsize
  \providecommand{\stdv}[1]{\,\scalebox{0.75}{$\pm$#1}}
  \setlength{\tabcolsep}{0.6pt}
  \begin{tabularx}{\textwidth}{@{}l *{9}{Y}@{}}
    \toprule
    \multirow{2}{*}{System ($n_x$, $n_u$)} & \multicolumn{3}{c}{$\rho_\mathrm{FS}$ (\%) $\downarrow$} & \multicolumn{3}{c}{$\rho_\mathrm{FU}$ (\%) $\downarrow$} & \multicolumn{3}{c}{$\eta_\mathrm{eff}$ $\uparrow$} \\
    \cmidrule(lr){2-4}\cmidrule(lr){5-7}\cmidrule(lr){8-10}
     & ND & FCD & \shortstack{VRCD\\(ours)} & ND & FCD & \shortstack{VRCD\\(ours)} & ND & FCD & \shortstack{VRCD\\(ours)} \\
    \midrule
    Inverted Pendulum (2, 1) & 100.00 & \textbf{0.00}\stdv{0.00} & \textbf{0.00}\stdv{0.00} & \textbf{0.00}\stdv{0.00} & 15.96\stdv{0.53} & 1.11\stdv{0.30} & 0.00\stdv{0.00} & 0.08\stdv{0.00} & \textbf{0.22}\stdv{0.00} \\
    1D Double Integrator (2, 1) & \textbf{0.00}\stdv{0.00} & \textbf{0.00}\stdv{0.00} & \textbf{0.00}\stdv{0.00} & \textbf{0.95}\stdv{0.56} & 4.01\stdv{0.42} & 1.18\stdv{0.31} & \textbf{0.41}\stdv{0.00} & 0.39\stdv{0.00} & \textbf{0.41}\stdv{0.00} \\
    2D Vertical Drone (2, 1) & 36.80\stdv{5.39} & -- & \textbf{0.00}\stdv{0.00} & 4.73\stdv{6.81} & 4.72\stdv{6.81} & \textbf{1.88}\stdv{0.46} & 0.35\stdv{0.20} & 0.00\stdv{0.00} & \textbf{0.58}\stdv{0.01} \\
    Dubins Car (3, 1) & 0.06\stdv{0.04} & \textbf{0.01}\stdv{0.01} & 0.04\stdv{0.01} & 4.75\stdv{1.28} & 4.33\stdv{0.31} & \textbf{2.25}\stdv{0.28} & 0.92\stdv{0.00} & 0.92\stdv{0.00} & \textbf{0.93}\stdv{0.00} \\
    2D Double Integrator (4, 2) & 0.03\stdv{0.01} & \textbf{0.00}\stdv{0.00} & 0.03\stdv{0.01} & 4.68\stdv{0.25} & 6.84\stdv{0.30} & \textbf{3.42}\stdv{0.22} & \textbf{0.93}\stdv{0.00} & 0.92\stdv{0.00} & \textbf{0.93}\stdv{0.00} \\
    Kinematic Bicycle (4, 2) & 4.68\stdv{0.91} & \textbf{0.52}\stdv{0.38} & 1.16\stdv{0.51} & 15.49\stdv{1.07} & \textbf{3.16}\stdv{0.24} & 5.03\stdv{0.15} & 0.86\stdv{0.01} & \textbf{0.91}\stdv{0.00} & \textbf{0.91}\stdv{0.00} \\
    Cart-Pole (4, 1) & 88.14\stdv{8.15} & -- & \textbf{0.16}\stdv{0.19} & \textbf{0.65}\stdv{0.48} & 4.06\stdv{2.46} & 2.64\stdv{0.34} & 0.06\stdv{0.04} & 0.00\stdv{0.00} & \textbf{0.32}\stdv{0.00} \\
    Dynamic Unicycle (5, 2) & 4.17\stdv{0.89} & \textbf{0.09}\stdv{0.09} & 0.23\stdv{0.26} & \textbf{2.93}\stdv{0.38} & 5.50\stdv{0.42} & 6.49\stdv{0.44} & 0.90\stdv{0.01} & \textbf{0.93}\stdv{0.00} & \textbf{0.93}\stdv{0.00} \\
    Relative Unicycle (5, 2) & 6.06\stdv{0.10} & 0.26\stdv{0.51} & \textbf{0.16}\stdv{0.02} & 4.49\stdv{0.33} & 37.07\stdv{3.49} & \textbf{2.35}\stdv{0.13} & 0.87\stdv{0.00} & 0.87\stdv{0.01} & \textbf{0.93}\stdv{0.00} \\
    3D Double Integrator (6, 3) & 0.01\stdv{0.00} & \textbf{0.00}\stdv{0.00} & \textbf{0.00}\stdv{0.00} & \textbf{9.12}\stdv{0.37} & 11.20\stdv{0.43} & 10.60\stdv{0.27} & \textbf{0.97}\stdv{0.00} & \textbf{0.97}\stdv{0.00} & \textbf{0.97}\stdv{0.00} \\
    3-DOF Manipulator (6, 3) & 13.52\stdv{12.09} & 0.55\stdv{0.36} & \textbf{0.43}\stdv{0.49} & \textbf{3.69}\stdv{0.78} & 6.74\stdv{0.16} & 5.68\stdv{0.12} & 0.58\stdv{0.08} & 0.65\stdv{0.00} & \textbf{0.66}\stdv{0.00} \\
    Landing Rocket (7, 2) & 76.45\stdv{14.68} & 56.41\stdv{13.54} & \textbf{1.17}\stdv{0.78} & \textbf{0.13}\stdv{0.07} & 2.95\stdv{0.91} & 5.95\stdv{0.20} & 0.09\stdv{0.06} & 0.02\stdv{0.01} & \textbf{0.24}\stdv{0.00} \\
    Quadruped Trunk (9, 4) & 99.98\stdv{0.05} & -- & \textbf{0.07}\stdv{0.09} & \textbf{0.00}\stdv{0.00} & \textbf{0.00}\stdv{0.00} & 3.14\stdv{0.11} & 0.00\stdv{0.00} & 0.00\stdv{0.00} & \textbf{0.07}\stdv{0.00} \\
    6-DoF Underw. Veh. (12, 6) & 22.24\stdv{0.93} & \textbf{0.61}\stdv{0.07} & 1.00\stdv{0.25} & \textbf{0.80}\stdv{0.14} & 5.15\stdv{0.06} & 2.27\stdv{0.09} & 0.33\stdv{0.00} & 0.34\stdv{0.00} & \textbf{0.36}\stdv{0.00} \\
    Quadrotor (13, 4) & 5.75\stdv{0.20} & 1.21\stdv{0.08} & \textbf{1.07}\stdv{0.12} & 11.18\stdv{0.35} & \textbf{9.50}\stdv{0.42} & 10.10\stdv{0.42} & 0.88\stdv{0.00} & \textbf{0.91}\stdv{0.00} & \textbf{0.91}\stdv{0.00} \\
    \bottomrule
  \end{tabularx}
\end{table*}

\begin{figure}[ht]
  \centering
  \includegraphics[width=\textwidth]{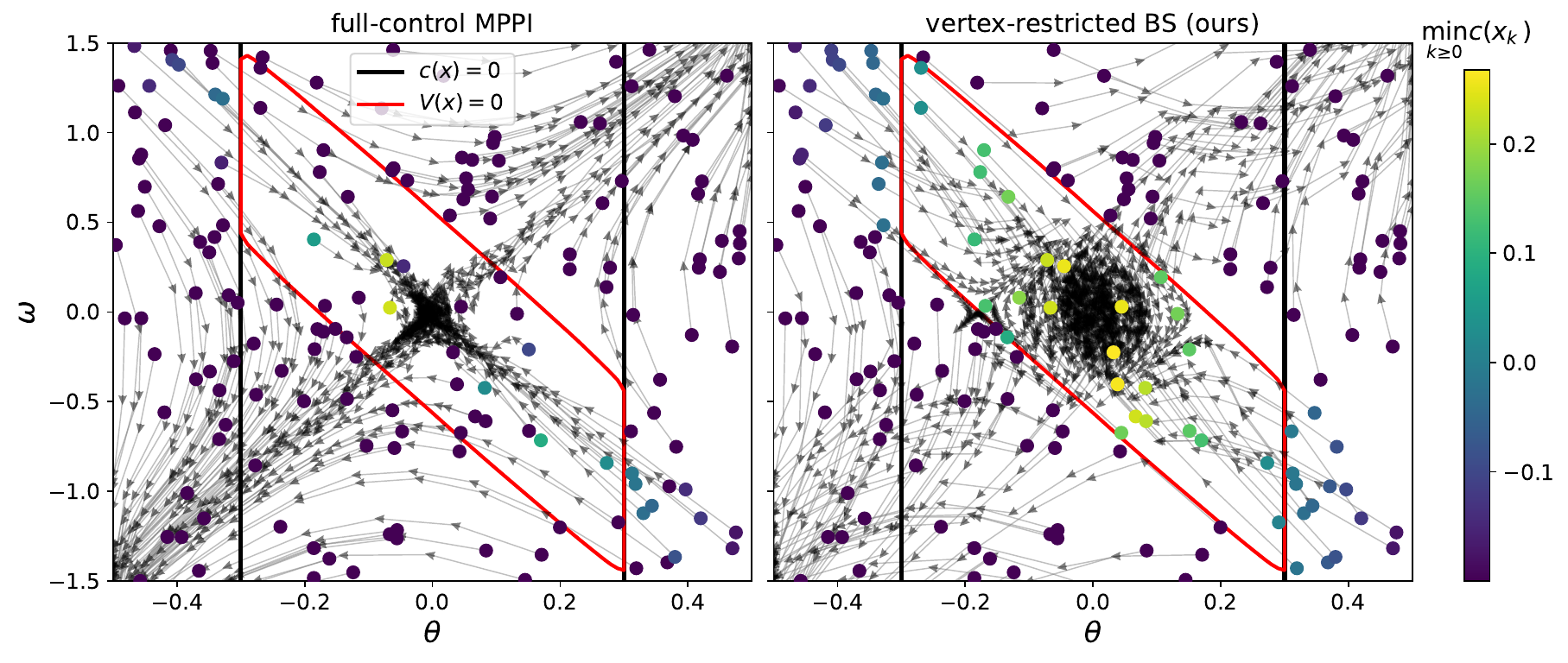}
  \caption{Finite-horizon trajectories for the inverted pendulum generated by full-control MPPI (left) and vertex-restricted beam search (right). Both methods start from the same $160$ random initial states, drawn as large circles and colored by the worst constraint value $\min_{k \geq 0} c(x_k)$ reached along the trajectory. The states that follow are drawn as small arrows that point in the direction of motion. The black line is the constraint boundary and the red line is the boundary of the true safe set. Vertex-restricted beam search keeps the system inside the constraint set for all sampled initial states of the true safe set, whereas full-control MPPI succeeds only for a few of them.}
  \label{fig:compare_traj_ip}
\end{figure}

In \cref{fig:compare_traj_ip}, we illustrate the difference between data-generation methods on the inverted pendulum system. Although this low-dimensional example can be handled with gradient- or grid-based solvers, our goal is to highlight the effects that arise in higher dimensions, where optimization runs in parallel across many initial states and sampling-based methods are standard. In this example, MPPI searches over the full control set but keeps the system within the constraint set for only a small number of initial states, whereas vertex-restricted beam search generates safe trajectories whenever possible.

\begin{figure}[ht]
  \centering
  \includegraphics[width=\textwidth]{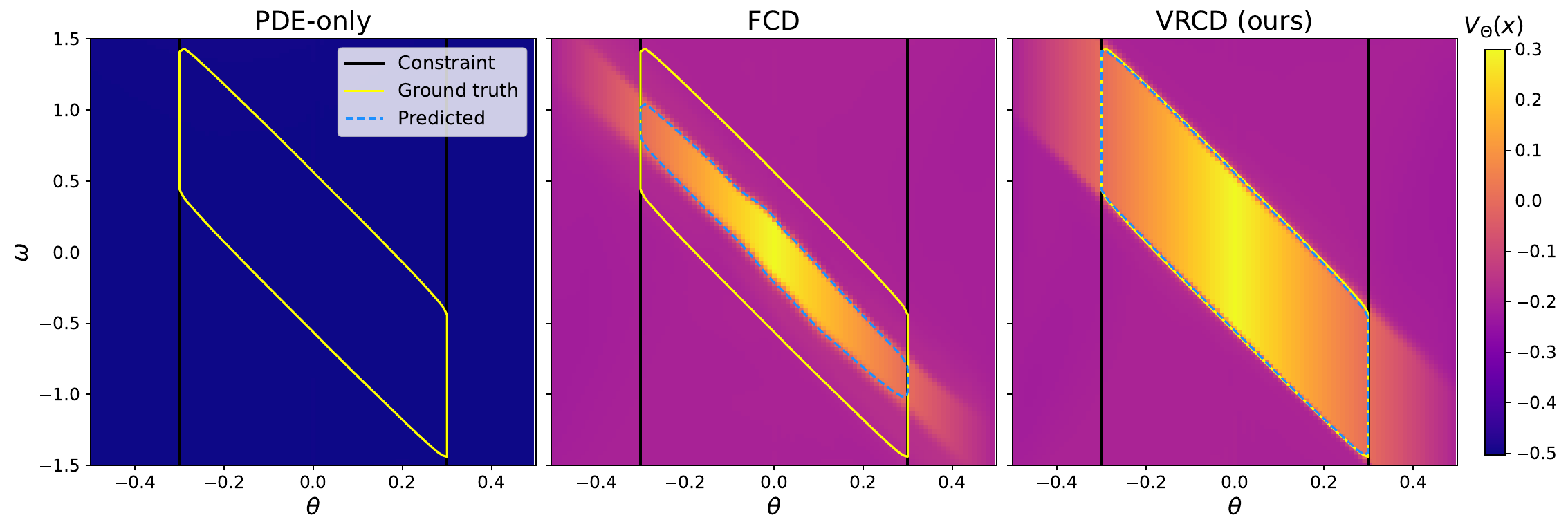}
  \caption{Learned neural CBFs for the inverted pendulum. PDE-only learning collapses to a trivial constant solution with an empty safe set, while FCD supervision results in an overly conservative solution. Only the proposed VRCD method approximately recovers the maximal safe set.}
  \label{fig:comparison_ip}
\end{figure}

\cref{fig:comparison_ip} shows that the PDE-only baseline can collapse to a trivial constant solution with an empty predicted safe set, even for a two-dimensional system, and the only seed that avoids this collapse yields a completely unsafe certificate (\cref{tab:num_exp_results}). Adding MPPI-based supervision improves the learned CBF, but the result is overly conservative. On the other hand, vertex-restricted beam search provides better supervision and enables the neural CBF to approximately recover the maximal safe set.

\subsection{Hardware Experiments}
\label{subsec:hardw_exp}

\begin{figure}[ht]
  \centering
  \includegraphics[width=\textwidth]{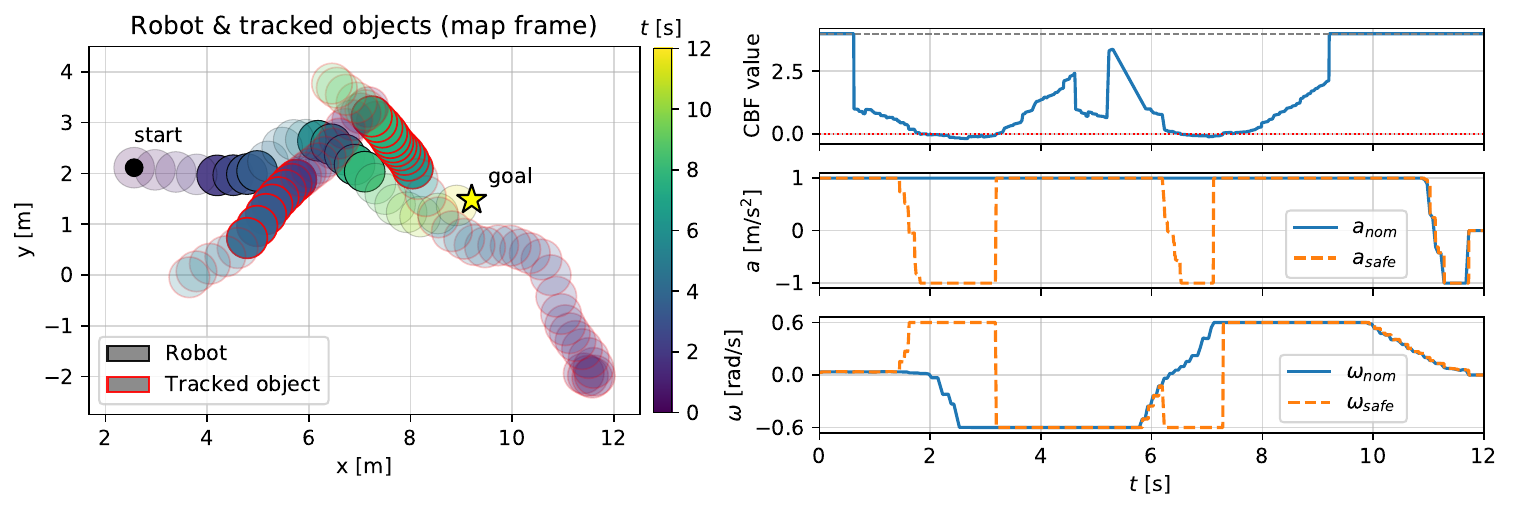}
  \caption{Results from the hardware experiment with the mobile robot --- visualized motion of the robot and tracked pedestrians (left), and CBF value and control signals (right).}
  \label{fig:hardw_exp}
\end{figure}

We conduct hardware experiments in which a mobile robot must avoid collisions with pedestrians using the neural CBF obtained with the proposed method. The nominal controller is an MPC planner without safety constraints that drives the robot from the start to the goal position. The CBF-QP filter receives pedestrian detections from the perception module and filters the nominal control to achieve safe motion. The CBF is designed for the relative dynamics between the robot and an obstacle, as described in \cite{derajic_cn-cbf_2026}. The learned CBF is the one reported for the relative unicycle in \cref{tab:num_exp_results}, and additional details are provided in \cref{app:details_hardw_exp} and \cref{app:rel_unicycle}. We conduct 10 trials in which one or two pedestrians intentionally intercept the robot's motion, and the robot avoids a collision in all of them. \cref{fig:hardw_exp} shows results from one representative scenario, in which the robot safely navigates among pedestrians using the learned CBF.

In this scenario, the CBF value stays close to zero and becomes slightly negative while the pedestrians pass the robot. The CBF constraint in the QP is relaxed by a slack variable, so the QP remains feasible at all times. The small violations are therefore not caused by an infeasible QP, but by the mismatch between the training model and the real system. The CBF is learned for an ideal kinematic unicycle, whereas the real skid-steer robot is affected by tire slip, unmodeled actuator dynamics, and noisy and delayed perception. To compensate, the robot and pedestrian radii used during training are a few centimeters larger than the physical dimensions. A small negative CBF value therefore does not cause a collision.

%===============================================================================

\section{Conclusion}
\label{sec:conclusion}

We presented VertexCBF, a framework for learning neural CBFs that approximate stationary HJ safety value functions. By exploiting control-affine dynamics and convex-polytope control constraints, it generates supervision through vertex-restricted finite-horizon search, avoiding grid-based PDE solvers and full-control trajectory optimization. Combined with a physics-informed loss and a residual architecture that upper-bounds the learned function by the constraint function, this provides a practical and systematic design approach for learning neural CBFs.

Experiments across diverse systems show that the proposed method recovers larger and more reliable validated safe sets than PDE-only learning and full-control MPPI supervision. Vertex-restricted control therefore provides an effective supervision signal that guides learning toward the optimal solution. The hardware experiment further demonstrates that the learned CBF can be deployed in real time as a CBF-QP safety filter for robots operating around people.

Future work includes establishing stronger convergence and error guarantees for the learned CBFs, better scalability for systems with many control vertices, and extensions to uncertain dynamics and perception-based constraints.

%===============================================================================

\section{Limitations}
\label{sec:limitations}

Our method does not yet provide formal convergence guarantees or deterministic error bounds on the learned CBF. The approximation error comes from time discretization, finite-horizon truncation, width-limited tree search, and neural function approximation. We therefore rely on post-training validation with large-scale closed-loop rollouts. Future work could strengthen this validation with sample-based certification methods such as scenario optimization or conformal prediction, which compute safety margins with finite-sample probabilistic guarantees \cite{lin_verification_2024, feng_bridging_2025}.

A second limitation is the scalability of data generation. Although restricting controls to polytope vertices reduces the search space, the tree search still grows rapidly with the horizon and the number of control vertices. The method scales well to the systems considered in this paper, including systems with up to 13 states and 6 controls. However, applying it directly to very high-dimensional whole-body dynamics, such as full humanoid models, may be computationally prohibitive. In such settings, the method may be more suitable for reduced-order models or safety-critical subsystems.

%=====================

\clearpage
% The acknowledgments are automatically included only in the final and preprint versions of the paper.
% \acknowledgments{This research was supported by the German Federal Ministry for Economic Affairs and Energy within the project "NXT GEN AI METHODS – Generative Methoden für Perzeption, Prädiktion und Planung".}

%===============================================================================

\bibliography{references}  % .bib

\begin{thebibliography}{29}
\providecommand{\natexlab}[1]{#1}
\providecommand{\url}[1]{\texttt{#1}}
\expandafter\ifx\csname urlstyle\endcsname\relax
  \providecommand{\doi}[1]{doi: #1}\else
  \providecommand{\doi}{doi: \begingroup \urlstyle{rm}\Url}\fi

\bibitem[Hsu et~al.(2024)Hsu, Hu, and Fisac]{hsu_safety_2024}
K.-C. Hsu, H.~Hu, and J.~F. Fisac.
\newblock The {Safety} {Filter}: {A} {Unified} {View} of {Safety}-{Critical} {Control} in {Autonomous} {Systems}.
\newblock \emph{Annual Review of Control, Robotics, and Autonomous Systems}, 7\penalty0 (Volume 7, 2024):\penalty0 47--72, 2024.
\newblock \doi{10.1146/annurev-control-071723-102940}.

\bibitem[Ames et~al.(2014)Ames, Grizzle, and Tabuada]{ames_control_2014}
A.~D. Ames, J.~W. Grizzle, and P.~Tabuada.
\newblock Control barrier function based quadratic programs with application to adaptive cruise control.
\newblock In \emph{{IEEE} {Conference} on {Decision} and {Control} ({CDC})}, pages 6271--6278, 2014.
\newblock \doi{10.1109/CDC.2014.7040372}.

\bibitem[Ames et~al.(2019)Ames, Coogan, Egerstedt, Notomista, Sreenath, and Tabuada]{ames_control_2019}
A.~D. Ames, S.~Coogan, M.~Egerstedt, G.~Notomista, K.~Sreenath, and P.~Tabuada.
\newblock Control {Barrier} {Functions}: {Theory} and {Applications}.
\newblock In \emph{European {Control} {Conference} ({ECC})}, pages 3420--3431, 2019.
\newblock \doi{10.23919/ECC.2019.8796030}.

\bibitem[Wabersich et~al.(2023)Wabersich, Taylor, Choi, Sreenath, Tomlin, Ames, and Zeilinger]{wabersich_data-driven_2023}
K.~P. Wabersich, A.~J. Taylor, J.~J. Choi, K.~Sreenath, C.~J. Tomlin, A.~D. Ames, and M.~N. Zeilinger.
\newblock Data-{Driven} {Safety} {Filters}: {Hamilton}-{Jacobi} {Reachability}, {Control} {Barrier} {Functions}, and {Predictive} {Methods} for {Uncertain} {Systems}.
\newblock \emph{IEEE Control Systems}, 43\penalty0 (5):\penalty0 137--177, 2023.
\newblock \doi{10.1109/MCS.2023.3291885}.

\bibitem[Dawson et~al.(2023)Dawson, Gao, and Fan]{dawson_safe_2023}
C.~Dawson, S.~Gao, and C.~Fan.
\newblock Safe {Control} {With} {Learned} {Certificates}: {A} {Survey} of {Neural} {Lyapunov}, {Barrier}, and {Contraction} {Methods} for {Robotics} and {Control}.
\newblock \emph{IEEE Transactions on Robotics}, 39\penalty0 (3):\penalty0 1749--1767, 2023.
\newblock \doi{10.1109/TRO.2022.3232542}.

\bibitem[Srinivasan et~al.(2020)Srinivasan, Dabholkar, Coogan, and Vela]{srinivasan_synthesis_2020}
M.~Srinivasan, A.~Dabholkar, S.~Coogan, and P.~A. Vela.
\newblock Synthesis of {Control} {Barrier} {Functions} {Using} a {Supervised} {Machine} {Learning} {Approach}.
\newblock In \emph{2020 {IEEE}/{RSJ} {International} {Conference} on {Intelligent} {Robots} and {Systems} ({IROS})}, pages 7139--7145. IEEE, 2020.
\newblock \doi{10.1109/IROS45743.2020.9341190}.

\bibitem[Long et~al.(2021)Long, Qian, Cortés, and Atanasov]{long_learning_2021}
K.~Long, C.~Qian, J.~Cortés, and N.~Atanasov.
\newblock Learning {Barrier} {Functions} {With} {Memory} for {Robust} {Safe} {Navigation}.
\newblock \emph{IEEE Robotics and Automation Letters}, 6\penalty0 (3):\penalty0 4931--4938, 2021.
\newblock \doi{10.1109/LRA.2021.3070250}.

\bibitem[Chen et~al.(2025)Chen, Swann, Yu, Shorinwa, Murai, Kennedy, and Schwager]{chen_control_2025}
T.~Chen, A.~Swann, J.~Yu, O.~Shorinwa, R.~Murai, M.~Kennedy, and M.~Schwager.
\newblock A {Control} {Barrier} {Function} for {Safe} {Navigation} with {Online} {Gaussian} {Splatting} {Maps}.
\newblock In \emph{{IEEE} {International} {Conference} on {Robotics} and {Automation} ({ICRA})}, pages 11758--11765, 2025.
\newblock \doi{10.1109/ICRA55743.2025.11128723}.

\bibitem[Liu et~al.(2022)Liu, Liu, and Dolan]{liu_safe_2022}
S.~Liu, C.~Liu, and J.~Dolan.
\newblock Safe {Control} {Under} {Input} {Limits} with {Neural} {Control} {Barrier} {Functions}.
\newblock In \emph{Conference on {Robot} {Learning} ({CoRL})}, 2022.

\bibitem[Xiao et~al.(2023)Xiao, Wang, Hasani, Chahine, Amini, Li, and Rus]{xiao_barriernet_2023}
W.~Xiao, T.-H. Wang, R.~Hasani, M.~Chahine, A.~Amini, X.~Li, and D.~Rus.
\newblock {BarrierNet}: {Differentiable} {Control} {Barrier} {Functions} for {Learning} of {Safe} {Robot} {Control}.
\newblock \emph{IEEE Transactions on Robotics}, 39\penalty0 (3):\penalty0 2289--2307, 2023.
\newblock \doi{10.1109/TRO.2023.3249564}.

\bibitem[Derajić et~al.(2025)Derajić, Bernhard, and Hönig]{derajic_orn-cbf_2025}
B.~Derajić, S.~Bernhard, and W.~Hönig.
\newblock {ORN}-{CBF}: {Learning} {Observation}-conditioned {Residual} {Neural} {Control} {Barrier} {Functions} via {Hypernetworks}, 2025.
\newblock arXiv:2509.16614 [cs].

\bibitem[Lin et~al.(2026)Lin, Peng, and Bansal]{lin_one_2026}
A.~Lin, S.~Peng, and S.~Bansal.
\newblock One {Filter} to {Deploy} {Them} {All}: {Robust} {Safety} for {Quadrupedal} {Navigation} in {Unknown} {Environments}.
\newblock \emph{IEEE Transactions on Robotics}, 42:\penalty0 545--560, 2026.
\newblock \doi{10.1109/TRO.2025.3644957}.

\bibitem[Derajić et~al.(2026)Derajić, Bernhard, and Hönig]{derajic_cn-cbf_2026}
B.~Derajić, S.~Bernhard, and W.~Hönig.
\newblock {CN}-{CBF}: {Composite} {Neural} {Control} {Barrier} {Function} for {Robot} {Navigation} in {Dynamic} {Environments}, 2026.
\newblock arXiv:2603.06921 [cs].

\bibitem[Derajic et~al.(2025)Derajic, Bouzidi, Bernhard, and Hönig]{derajic_residual_2025}
B.~Derajic, M.-K. Bouzidi, S.~Bernhard, and W.~Hönig.
\newblock Residual {Neural} {Terminal} {Constraint} for {MPC}-based {Collision} {Avoidance} in {Dynamic} {Environments}.
\newblock In \emph{Proceedings of {The} 9th {Conference} on {Robot} {Learning}}, pages 1452--1469. PMLR, 2025.

\bibitem[Bansal and Tomlin(2021)]{bansal_deepreach_2021}
S.~Bansal and C.~J. Tomlin.
\newblock {DeepReach}: {A} {Deep} {Learning} {Approach} to {High}-{Dimensional} {Reachability}.
\newblock In \emph{{IEEE} {International} {Conference} on {Robotics} and {Automation} ({ICRA})}, pages 1817--1824, 2021.
\newblock \doi{10.1109/ICRA48506.2021.9561949}.

\bibitem[Singh et~al.(2025)Singh, Feng, and Bansal]{singh_exact_2025}
A.~Singh, Z.~Feng, and S.~Bansal.
\newblock Exact {Imposition} of {Safety} {Boundary} {Conditions} in {Neural} {Reachable} {Tubes}.
\newblock In \emph{2025 {IEEE} {International} {Conference} on {Robotics} and {Automation} ({ICRA})}, pages 5489--5495, 2025.
\newblock \doi{10.1109/ICRA55743.2025.11127972}.

\bibitem[Lin and Bansal(2024)]{lin_verification_2024}
A.~Lin and S.~Bansal.
\newblock Verification of neural reachable tubes via scenario optimization and conformal prediction.
\newblock In \emph{Proceedings of the 6th {Annual} {Learning} for {Dynamics} \& {Control} {Conference}}, pages 719--731. PMLR, 2024.

\bibitem[Manda et~al.(2025)Manda, Chen, and Fazlyab]{manda_learning_2025}
L.~Manda, S.~Chen, and M.~Fazlyab.
\newblock Learning {Performance}-oriented {Control} {Barrier} {Functions} {Under} {Complex} {Safety} {Constraints} and {Limited} {Actuation}.
\newblock In \emph{Proceedings of {The} 8th {Conference} on {Robot} {Learning}}, pages 4793--4804. PMLR, 2025.

\bibitem[Feng et~al.(2025)Feng, Qiu, and Bansal]{feng_bridging_2025}
Z.~Feng, L.~Qiu, and S.~Bansal.
\newblock Bridging {Model} {Predictive} {Control} and {Deep} {Learning} for {Scalable} {Reachability} {Analysis}.
\newblock In \emph{Proceedings of {Robotics}: {Science} and {Systems}}, volume~21, 2025.
\newblock ISBN 979-8-9902848-1-4.

\bibitem[Bansal et~al.(2017)Bansal, Chen, Herbert, and Tomlin]{bansal_hamilton-jacobi_2017}
S.~Bansal, M.~Chen, S.~Herbert, and C.~J. Tomlin.
\newblock Hamilton-{Jacobi} reachability: {A} brief overview and recent advances.
\newblock In \emph{{IEEE} {Conference} on {Decision} and {Control} ({CDC})}, pages 2242--2253, 2017.
\newblock \doi{10.1109/CDC.2017.8263977}.

\bibitem[Hsu et~al.(2021)Hsu, Rubies-Royo, Tomlin, and Fisac]{hsu_safety_2021}
K.-C. Hsu, V.~Rubies-Royo, C.~Tomlin, and J.~Fisac.
\newblock Safety and {Liveness} {Guarantees} through {Reach}-{Avoid} {Reinforcement} {Learning}.
\newblock In \emph{Robotics: {Science} and {Systems} {XVII}}. Robotics: Science and Systems Foundation, 2021.
\newblock ISBN 978-0-9923747-7-8.
\newblock \doi{10.15607/RSS.2021.XVII.077}.

\bibitem[Nagumo(1942)]{nagumo_uber_1942}
M.~Nagumo.
\newblock Über die {Lage} der {Integralkurven} gewöhnlicher {Differentialgleichungen}.
\newblock \emph{Proceedings of the Physico-Mathematical Society of Japan. 3rd Series}, 24:\penalty0 551--559, 1942.
\newblock \doi{10.11429/ppmsj1919.24.0_551}.

\bibitem[Fisac et~al.(2019)Fisac, Lugovoy, Rubies-Royo, Ghosh, and Tomlin]{fisac_bridging_2019}
J.~F. Fisac, N.~F. Lugovoy, V.~Rubies-Royo, S.~Ghosh, and C.~J. Tomlin.
\newblock Bridging {Hamilton}-{Jacobi} {Safety} {Analysis} and {Reinforcement} {Learning}.
\newblock In \emph{2019 {International} {Conference} on {Robotics} and {Automation} ({ICRA})}, pages 8550--8556, 2019.
\newblock \doi{10.1109/ICRA.2019.8794107}.

\bibitem[So et~al.(2024)So, Serlin, Mann, Gonzales, Rutledge, Roy, and Fan]{so_how_2024}
O.~So, Z.~Serlin, M.~Mann, J.~Gonzales, K.~Rutledge, N.~Roy, and C.~Fan.
\newblock How to {Train} {Your} {Neural} {Control} {Barrier} {Function}: {Learning} {Safety} {Filters} for {Complex} {Input}-{Constrained} {Systems}.
\newblock In \emph{{IEEE} {International} {Conference} on {Robotics} and {Automation} ({ICRA})}, pages 11532--11539, 2024.
\newblock \doi{10.1109/ICRA57147.2024.10610418}.

\bibitem[Akametalu et~al.(2024)Akametalu, Ghosh, Fisac, Rubies-Royo, and Tomlin]{akametalu_minimum_2024}
A.~K. Akametalu, S.~Ghosh, J.~F. Fisac, V.~Rubies-Royo, and C.~J. Tomlin.
\newblock A {Minimum} {Discounted} {Reward} {Hamilton}–{Jacobi} {Formulation} for {Computing} {Reachable} {Sets}.
\newblock \emph{IEEE Transactions on Automatic Control}, 69\penalty0 (2):\penalty0 1097--1103, 2024.
\newblock \doi{10.1109/TAC.2023.3327159}.

\bibitem[Sitzmann et~al.(2020)Sitzmann, Martel, Bergman, Lindell, and Wetzstein]{sitzmann_implicit_2020}
V.~Sitzmann, J.~Martel, A.~Bergman, D.~Lindell, and G.~Wetzstein.
\newblock Implicit {Neural} {Representations} with {Periodic} {Activation} {Functions}.
\newblock In \emph{Advances in {Neural} {Information} {Processing} {Systems}}, volume~33, pages 7462--7473. Curran Associates, Inc., 2020.

\bibitem[Russell and Norvig(2016)]{russell_artificial_2016}
S.~J. Russell and P.~Norvig.
\newblock \emph{Artificial {Intelligence}: {A} {Modern} {Approach}}.
\newblock Prentice {Hall} {Series} in {Artificial} {Intelligence}. Pearson, 3rd edition, 2016.
\newblock ISBN 978-0-13-604259-4.

\bibitem[Kool et~al.(2019)Kool, Hoof, and Welling]{kool_stochastic_2019}
W.~Kool, H.~V. Hoof, and M.~Welling.
\newblock Stochastic {Beams} and {Where} {To} {Find} {Them}: {The} {Gumbel}-{Top}-k {Trick} for {Sampling} {Sequences} {Without} {Replacement}.
\newblock In \emph{Proceedings of the 36th {International} {Conference} on {Machine} {Learning} ({ICML})}, pages 3499--3508. PMLR, 2019.

\bibitem[Land and Doig(1960)]{land_automatic_1960}
A.~H. Land and A.~G. Doig.
\newblock An {Automatic} {Method} of {Solving} {Discrete} {Programming} {Problems}.
\newblock \emph{Econometrica}, 28\penalty0 (3):\penalty0 497, 1960.
\newblock \doi{10.2307/1910129}.

\end{thebibliography}

\newpage

\appendix
\crefalias{section}{appendix}
\crefalias{subsection}{appendix}

\section{First-order consistency of the vertex-restricted discretization}
\label{app:first_order_consistency}

In this appendix, we show that the vertex-restricted Bellman equation \cref{eq:bellman_eq} is a first-order consistent approximation of the continuous stationary HJB-VI \cref{eq:stationary_hjb_vi}. Since safety value functions are generally nonsmooth, consistency is checked using a smooth test function $\psi$ rather than the value function itself. Intuitively, $\psi$ acts as a smooth local proxy for the value function near a regular point.

Let $\mathcal Y \subseteq \mathcal X$ be a compact local region containing $x$ and $F(x,v)$ for all $v\in\mathcal V$ and all sufficiently small $\Delta t$. Let $\psi\in C^{1,1}(\mathcal Y)$. For any vertex $v\in\mathcal V$, Taylor's theorem gives
\begin{equation}
    \psi(F(x,v))
    =
    \psi(x)
    +
    \Delta t\,\nabla\psi(x)^\top \phi(x,v)
    +
    R_{\Delta t}(x,v),
    \label{eq:taylor_test_function}
\end{equation}
where
\begin{equation}
    |R_{\Delta t}(x,v)|
    \le
    \frac12 L_{\nabla\psi}
    \|\phi(x,v)\|^2
    \Delta t^2 .
    \label{eq:taylor_remainder}
\end{equation}
Here $L_{\nabla\psi}$ denotes a Lipschitz constant of $\nabla\psi$ on $\mathcal Y$. If $\psi\in C^2(\mathcal Y)$, one may take $L_{\nabla\psi}=\sup_{y\in\mathcal Y}\|\nabla^2\psi(y)\|$.

Taking the maximum over vertices in \cref{eq:taylor_test_function} yields
\begin{align}
    \max_{v\in\mathcal V}\psi(F(x,v))
    &=
    \psi(x)
    +
    \Delta t
    \max_{v\in\mathcal V}
    \nabla\psi(x)^\top \phi(x,v)
    +
    O(\Delta t^2) \\
    &=
    \psi(x)
    +
    \Delta t
    H(x,\nabla\psi(x))
    +
    O(\Delta t^2),
    \label{eq:consistency_expansion}
\end{align}
where the second equality uses the vertex representation of the Hamiltonian. Equivalently,
\begin{equation}
    \frac{
    \max_{v\in\mathcal V}\psi(F(x,v))
    -
    \psi(x)
    }{\Delta t}
    =
    H(x,\nabla\psi(x))
    +
    O(\Delta t).
    \label{eq:normalized_consistency}
\end{equation}
More explicitly, the local truncation error satisfies
\begin{equation}
\left|
    \frac{
    \max_{v\in\mathcal V}\psi(F(x,v))
    -
    \psi(x)
    }{\Delta t}
    -
    H(x,\nabla\psi(x))
\right|
\le
\frac12 L_{\nabla\psi}
\max_{v\in\mathcal V}
\|\phi(x,v)\|^2
\Delta t .
\label{eq:local_truncation_error}
\end{equation}
Thus, the one-step Bellman expansion has $O(\Delta t^2)$ error, while the time-normalized finite-difference approximation of the Hamiltonian has $O(\Delta t)$ error.

Substituting \cref{eq:normalized_consistency} into the discrete variational-inequality operator \cref{eq:discrete_hjb_vi} gives
\begin{equation}
\begin{split}
    \min&\left\{
        c(x)-\psi(x),
        \;
        \frac{
        \max_{v\in\mathcal V}\psi(F(x,v))
        -
        \psi(x)
        }{\Delta t}
    \right\} \\
    &=
    \min\left\{
        c(x)-\psi(x),
        \;
        H(x,\nabla\psi(x))
    \right\}
    +
    O(\Delta t).
\end{split}
    \label{eq:discrete_to_continuous_vi_consistency}
\end{equation}
Therefore, the vertex-restricted Euler Bellman operator is a first-order consistent discretization of the continuous stationary HJB-VI operator.

\paragraph{Scope of the consistency result.}
The calculation above establishes only local consistency of the vertex-restricted Bellman operator with the stationary HJB-VI. It is not a complete convergence proof for the infinite-horizon value function, which would also require well-posedness of the limiting viscosity problem, including a comparison principle, appropriate boundary or state-constraint conditions, and a selection argument identifying the dynamic-programming solution among possible stationary viscosity solutions. We use this result only to justify the local first-order behavior of the proposed vertex-restricted discretization.

\section{Overview of the Tree Search Methods}
\label{app:overview_tree_search}

This appendix details the three vertex-restricted tree search methods used to generate supervision labels in \cref{eq:finite_horizon_value}, namely beam search, stochastic beam search, and branch and bound \cite{russell_artificial_2016, kool_stochastic_2019, land_automatic_1960}. All three methods seek to approximate the same optimization problem
\begin{equation} \label{eq:bs_objective}
    V_{\Delta t}^{K}(x_0) \;=\; \max_{u_0,\ldots,u_{K-1} \in \mathcal{V}}
    \;\min_{0 \le k \le K} c(x_k),
    \qquad x_{k+1} = F(x_k, u_k),
\end{equation}
over the tree whose nodes at depth $k$ are reachable states and whose $M = |\mathcal{V}|$ children of each node correspond to the control vertices. The full tree has $M^K$ leaves, so exact enumeration is intractable for large $M$ or $K$ and we use width-bounded heuristic searches with a budget $B$. All three methods are implemented in a batched GPU-parallel form. At every depth, all $N$ initial states, all current beam entries, and all $M$ vertices are expanded and scored in a single forward pass. The three methods differ only in the rule for selecting which candidates to retain.

\paragraph{Score function.}
For a partial trajectory committed up to depth $k$, the running-minimum score
\begin{equation}
    s_k \;=\; \min_{0 \le j \le k} c(x_j)
\end{equation}
is monotonically non-increasing in $k$, so $s_k$ is an upper bound on $\min_{0 \le j \le K} c(x_j)$ for any extension to depth $K$. The three methods differ only in how they use $s_k$ to prune the search.

\subsection{Beam Search (BS)}
\label{app:bs}

Beam search keeps, after each expansion, the $B$ partial trajectories with the largest score $s_k$. At depth $k$ with current beam of size $W \le B$, all $W \cdot M$ children are generated by appending every $u \in \mathcal{V}$ and propagating via $F$. The running-minimum score is then updated, and the top-$B$ children by $s_{k+1}$ are retained. After $K$ steps, the returned label is $\max_{i=1,\ldots,B} s_K^{(i)}$, where $s_K^{(i)}$ is the score of the $i$-th surviving trajectory in the final beam.

\subsection{Stochastic Beam Search (SBS)}
\label{app:sbs}

Deterministic beam search may converge to a single locally optimal branch early in the search. SBS replaces the deterministic top-$B$ selection with a score-proportional draw of $B$ children \emph{without replacement}, preserving diversity. We implement four sampling strategies:
\begin{itemize}
    \item \textbf{Softmax}: multinomial sampling from $\operatorname{softmax}(s/T)$ with temperature $T > 0$.
    \item \textbf{Gumbel top-$k$}: add i.i.d.~$\operatorname{Gumbel}(0,1)$ noise to $s/T$ and take the top-$B$ — an unbiased, fully parallel sampler without replacement.
    \item \textbf{Rank}: weight proportional to $(C - \operatorname{rank}+1)$, independent of score scale and requiring no temperature.
    \item \textbf{$\varepsilon$-greedy}: with probability $\varepsilon$, replace the deterministic top-$B$ by $B$ uniformly random candidates.
\end{itemize}

\subsection{Branch and Bound (BnB)}
\label{app:bnb}

Because $s_k$ is an upper bound on the score of any completion, a child node with $s_{k+1} \le s_{\mathrm{LB}}$ cannot improve the global lower bound $s_{\mathrm{LB}}$ on \cref{eq:bs_objective} and may be safely discarded. Our BnB performs $n_{\mathrm{restarts}}$ passes over the same tree:
\begin{itemize}
    \item \textbf{Pass 1}: $s_{\mathrm{LB}} = -\infty$, no pruning, so the pass behaves like BS and yields an initial $s_{\mathrm{LB}}$ from the best leaf $s_K$.
    \item \textbf{Passes 2 onwards}: $s_{\mathrm{LB}}$ is the best leaf score found over all previous passes. At every depth, children with $s_{k+1} \le s_{\mathrm{LB}}$ are masked out before the top-$B$ selection. Small Gaussian tie-breaking noise on the selection scores diversifies which $B$ survive across passes, so that pruning allows previously unexplored subtrees to enter the beam.
\end{itemize}
Pruning preserves optimality with respect to the searched subtree and BnB therefore returns a value at least as large as BS for the same $B$, at the cost of running $n_{\mathrm{restarts}}$ passes.

\paragraph{Choice of the search variant.}
All three methods solve the same problem \cref{eq:bs_objective} and differ only in how they use the budget $B$. The choice therefore depends on the width of the beam relative to the size of the tree. When $M$ is small, the retained beam already covers a large portion of the complete control tree at every depth, the discarded subtrees are unlikely to contain the optimum, and BS is preferred for its simplicity and its deterministic labels. As $M$ grows, the beam becomes narrow relative to the $M^k$ nodes at depth $k$, and a deterministic top-$B$ selection is more likely to commit early to a locally optimal branch and to discard the optimal one. In this regime it is beneficial to use part of the budget for exploration, either by randomizing the selection as in SBS, or by running several passes in which the best value found in the previous passes is used to prune the tree, so that previously unexplored subtrees can enter the beam, as in BnB. Both increase the chance of obtaining a better solution at the same beam width, at the cost of stochastic labels for SBS and of $n_{\mathrm{restarts}}$ times the runtime for BnB. The variant used for each system is listed in the configuration tables of \cref{app:details_num_exp}.

\section{Influence of Beam-Search Hyperparameters}
\label{app:ablation_beam_search}
% -----------------------------------------------------------------------------

The supervision data for VertexCBF are generated by one of the beam-search-based methods, and the data quality is mainly determined by three parameters, namely the beam width $B$, the horizon length $K$, and the step size $\Delta t$. In this section, we consider the inverted pendulum because its numerically computed value function is available, although the conclusions are applicable to other systems.

\begin{figure}[ht]
  \centering
  \includegraphics[width=\textwidth]{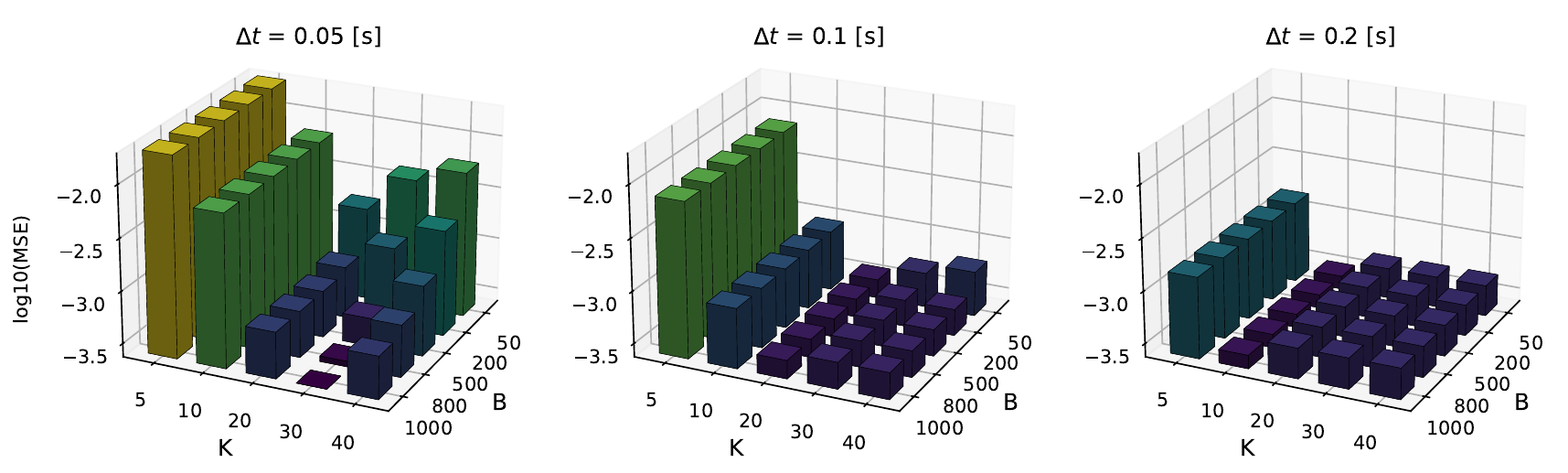}
  \caption{Beam-search MSE against the ground-truth value on the inverted pendulum system, as a function of $B$ and $K$ for different $\Delta t$.}
  \label{fig:ablation_mse_ip}
\end{figure}

We evaluate all combinations of $B \in \{50, 200, 500, 800, 1000\}$, $K \in \{5, 10, 20, 30, 40\}$, and $\Delta t \in \{0.05, 0.1, 0.2\}$ for the settings detailed in \cref{app:ip}. As the main metric, we report the mean-squared error (MSE) of the resulting $V_{\Delta t}^{K, B}(x)$ with respect to the ground-truth value $V(x)$ over a uniform state-space lattice. The results are presented in \cref{fig:ablation_mse_ip}, and several trends are evident. First, if the effective prediction horizon $K \Delta t$ is too short to detect future violations, the error is high because the value is systematically overestimated. Second, as the number of prediction steps $K$ increases, the beam width $B$ must also increase to retain near-optimal trajectories and maintain low approximation error. Finally, for larger $\Delta t$, fewer steps are needed to obtain an accurate approximation. However, a further increase of the step size makes the discretization inaccurate.

\begin{figure}[ht]
  \centering
  \includegraphics[width=0.4\textwidth]{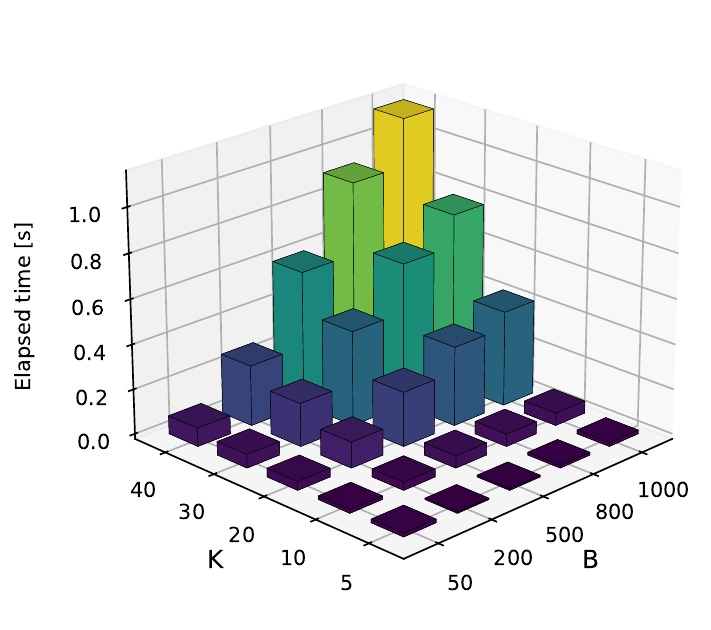}
  \caption{Execution time of the beam-search algorithm for the inverted pendulum as a function of beam width $B$ and tree depth $K$. Runtime scales approximately proportionally to $B \cdot K$ and does not depend on $\Delta t$.}
  \label{fig:ablation_runtime_ip}
\end{figure}

In addition to MSE, \cref{fig:ablation_runtime_ip} reports the beam-search execution time for the inverted pendulum under the same settings. Since the number of search operations does not depend on the step size $\Delta t$, we show runtime only as a function of $B$ and $K$. As expected, runtime grows approximately linearly with both parameters. At each depth, beam search evaluates up to $BM$ successors and retains the best $B$ partial trajectories. Therefore, for $N$ initial states and branching factor $M = |\mathcal{V}|$, the overall complexity is $\mathcal{O}(NKBM)$, up to the cost of simulating transitions and evaluating the objective.

\section{Influence of the Search Step Size}
\label{app:ablation_dt}

The step size $\Delta t$ enters the first term of \cref{eq:error_decomposition}, which collects the error of the forward-Euler discretization and the error of holding one vertex control over an interval of length $\Delta t$. A smaller $\Delta t$ reduces both, but for a fixed horizon $K$ it also shortens the look-ahead time $K \Delta t$, so the labels cover a shorter part of the future. We study this trade-off on the 6-DoF underwater vehicle.

We regenerate the supervision data for $\Delta t \in \{0.02, 0.05, 0.1, 0.2, 0.4\}$\,s and train a model from scratch for each of them with three different seeds. Every other setting is the one of \cref{tab:exp_auv_6dof}, including the horizon $K = 40$ and the beam width $B = 800$, so the look-ahead time ranges from $0.8$\,s to $16$\,s. The cost of the search does not depend on $\Delta t$, because the number of expanded nodes remains the same. The five settings corresponding to different $\Delta t$ values therefore differ in accuracy and not in runtime.

\begin{figure*}[ht]
  \centering
  \includegraphics[width=\textwidth]{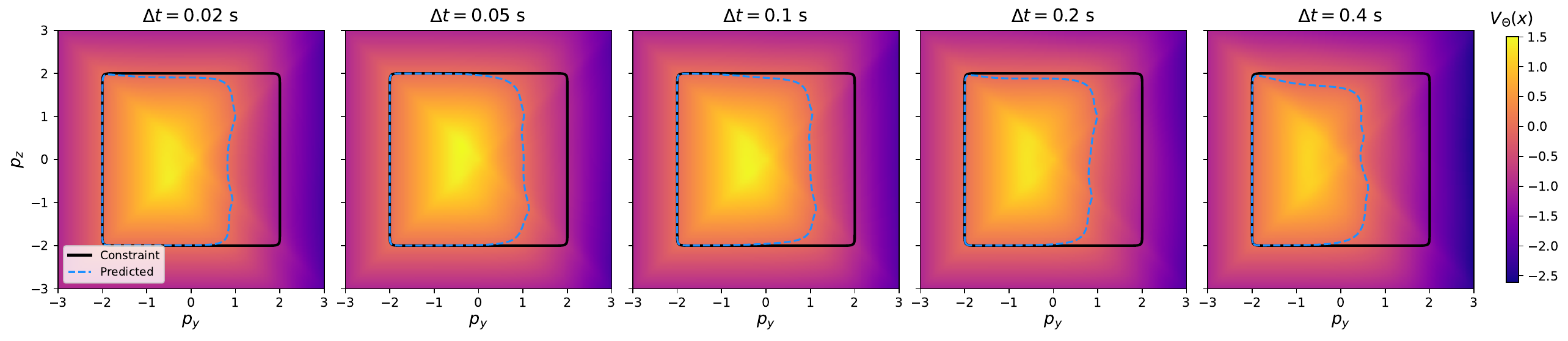}
  \caption{Learned neural CBF slices for the 6-DoF underwater vehicle for each search step size. The black line is the constraint boundary and the dashed line is the predicted safe-set boundary.}
  \label{fig:dt_slices_auv}
\end{figure*}

The validation metrics are reported in \cref{tab:ablation_dt} and the corresponding value function slices in \cref{fig:dt_slices_auv}. The smallest and the largest step sizes underperform for different reasons. At $\Delta t = 0.02$\,s the look-ahead time is only $0.8$\,s, which is shorter than the braking time of the vehicle. The labels then fail to detect later violations, and the false-safe rate rises to $3.94\%$. At $\Delta t = 0.4$\,s the discretization itself becomes inaccurate. The predicted safe set shrinks visibly in \cref{fig:dt_slices_auv}, the false-unsafe rate rises to $7.02\%$, and the effective safe volume drops to $0.317$.

Between these two extremes, the method is stable. For $\Delta t$ between $0.05$ and $0.2$\,s the false-safe rate remains at or below $1.18\%$ and the effective safe volume varies only between $0.348$ and $0.356$. The best result is obtained at $\Delta t = 0.1$\,s, which is the value used in the main experiments, and the spread over the three seeds is small everywhere. The resulting guideline is to choose $\Delta t$ small enough for the Euler step to remain accurate, and then to choose $K$ such that $K \Delta t$ covers the time the system needs to stop.

\begin{table}[ht]
  \centering
  \caption{Influence of the discretization step size $\Delta t$ on the 6-DoF underwater vehicle. All other settings are those of \cref{tab:exp_auv_6dof}. Entries are mean $\pm$ standard deviation over three training seeds.}
  \vspace{6pt}
  \label{tab:ablation_dt}
  \small
  \providecommand{\stdv}[1]{\,\scalebox{0.75}{$\pm$#1}}
  \begin{tabular}{@{}lccc@{}}
    \toprule
    $\Delta t$ [s] & $\rho_\mathrm{FS}$ (\%) $\downarrow$ & $\rho_\mathrm{FU}$ (\%) $\downarrow$ & $\eta_\mathrm{eff}$ $\uparrow$ \\
    \midrule
    $0.02$ & 3.94\stdv{0.24} & 2.88\stdv{0.03} & 0.340\stdv{0.001} \\
    $0.05$ & 0.63\stdv{0.03} & 3.12\stdv{0.02} & 0.354\stdv{0.000} \\
    $0.1$ & 1.15\stdv{0.06} & 2.16\stdv{0.02} & 0.356\stdv{0.000} \\
    $0.2$ & 1.18\stdv{0.13} & 3.22\stdv{0.06} & 0.348\stdv{0.000} \\
    $0.4$ & 0.51\stdv{0.13} & 7.02\stdv{0.14} & 0.317\stdv{0.000} \\
    \bottomrule
  \end{tabular}
\end{table}

\section{Ablation on the Residual Architecture}
\label{app:ablation_res_arch}

All methods in \cref{tab:num_exp_results} use the residual parametrization \cref{eq:residual_param}. To isolate its contribution, we repeat every experiment with a network that approximates the value function directly, replacing the softplus output of the residual $r_\Theta$ by a linear output activation. Architecture, hyperparameters, supervision data, and validation protocol are otherwise unchanged, and each configuration is again trained with five seeds. The results are reported in \cref{tab:cbf_results_abl_res}.

The residual parametrization proves essential for PDE-only learning. Without it, training converges to a trivial solution with an empty predicted safe set for 14 of the 15 systems, so both $\rho_\mathrm{FS}$ and $\eta_\mathrm{eff}$ degenerate and the large false-unsafe rates merely reflect that every state is predicted unsafe. Only the 3-DOF Manipulator retains a non-empty safe set, and it does so with a high and strongly seed-dependent false-safe rate. This behavior is expected, because a sufficiently negative constant value function makes the PDE loss \cref{eq:pde_loss} vanish everywhere.

The supervised methods are considerably less sensitive because the data loss already anchors the solution to non-trivial values. Removing the residual parametrization nevertheless degrades most entries. For VRCD, the false-safe rate increases on 13 of the 15 systems and the effective safe volume decreases on 11, with the largest losses on the Cart-Pole and the Quadrotor. The two components are thus complementary. The residual parametrization significantly decreases the risk of a trivial solution and guarantees $\hat{\mathcal{S}} \subseteq \mathcal{C}$ by construction, while the supervision data determine how closely $\hat{\mathcal{S}}$ approaches the maximal safe set.

\begin{table*}[b]
  \centering
  \caption{CBF validation results \emph{without} the residual parametrization, where $V_\Theta$ is approximated directly by a network with linear output activation. All other settings and reporting conventions follow \cref{tab:num_exp_results}.}
  \label{tab:cbf_results_abl_res}
  \scriptsize
  \providecommand{\stdv}[1]{\,\scalebox{0.75}{$\pm$#1}}
  \setlength{\tabcolsep}{0.6pt}
  \begin{tabularx}{\textwidth}{@{}l *{9}{Y}@{}}
    \toprule
    \multirow{2}{*}{System ($n_x$, $n_u$)} & \multicolumn{3}{c}{$\rho_\mathrm{FS}$ (\%) $\downarrow$} & \multicolumn{3}{c}{$\rho_\mathrm{FU}$ (\%) $\downarrow$} & \multicolumn{3}{c}{$\eta_\mathrm{eff}$ $\uparrow$} \\
    \cmidrule(lr){2-4}\cmidrule(lr){5-7}\cmidrule(lr){8-10}
     & ND & FCD & \shortstack{VRCD\\(ours)} & ND & FCD & \shortstack{VRCD\\(ours)} & ND & FCD & \shortstack{VRCD\\(ours)} \\
    \midrule
    Inverted Pendulum (2, 1) & -- & \textbf{1.06}\stdv{0.19} & 6.49\stdv{3.23} & \textbf{0.00}\stdv{0.01} & 14.91\stdv{0.76} & 0.29\stdv{0.17} & 0.00\stdv{0.00} & 0.08\stdv{0.00} & \textbf{0.21}\stdv{0.01} \\
    1D Double Integrator (2, 1) & -- & 0.16\stdv{0.07} & \textbf{0.15}\stdv{0.07} & 26.15\stdv{15.13} & 4.35\stdv{0.28} & \textbf{1.20}\stdv{0.10} & 0.00\stdv{0.00} & 0.39\stdv{0.00} & \textbf{0.41}\stdv{0.00} \\
    2D Vertical Drone (2, 1) & -- & -- & \textbf{0.14}\stdv{0.06} & 3.95\stdv{6.12} & 4.06\stdv{7.96} & \textbf{1.80}\stdv{0.14} & 0.00\stdv{0.00} & 0.00\stdv{0.00} & \textbf{0.58}\stdv{0.00} \\
    Dubins Car (3, 1) & -- & \textbf{0.08}\stdv{0.04} & 0.18\stdv{0.07} & 81.06\stdv{4.98} & 4.78\stdv{1.16} & \textbf{2.51}\stdv{0.49} & 0.00\stdv{0.00} & \textbf{0.92}\stdv{0.00} & \textbf{0.92}\stdv{0.00} \\
    2D Double Integrator (4, 2) & -- & 1.60\stdv{0.55} & \textbf{1.58}\stdv{0.83} & 67.67\stdv{10.10} & \textbf{3.69}\stdv{0.39} & 3.87\stdv{0.43} & 0.00\stdv{0.00} & \textbf{0.91}\stdv{0.01} & \textbf{0.91}\stdv{0.01} \\
    Kinematic Bicycle (4, 2) & -- & \textbf{0.19}\stdv{0.12} & 0.73\stdv{0.81} & 71.98\stdv{5.65} & \textbf{4.91}\stdv{0.25} & 5.77\stdv{0.45} & 0.00\stdv{0.00} & \textbf{0.91}\stdv{0.00} & \textbf{0.91}\stdv{0.01} \\
    Cart-Pole (4, 1) & -- & -- & \textbf{20.39}\stdv{20.37} & \textbf{1.66}\stdv{2.98} & 6.00\stdv{10.21} & 1.96\stdv{0.62} & 0.00\stdv{0.00} & 0.00\stdv{0.00} & \textbf{0.26}\stdv{0.07} \\
    Dynamic Unicycle (5, 2) & -- & \textbf{0.15}\stdv{0.05} & 0.26\stdv{0.05} & 79.18\stdv{1.55} & \textbf{8.41}\stdv{1.10} & 10.61\stdv{0.69} & 0.00\stdv{0.00} & \textbf{0.93}\stdv{0.00} & 0.92\stdv{0.00} \\
    Relative Unicycle (5, 2) & -- & \textbf{0.17}\stdv{0.11} & 0.36\stdv{0.07} & 83.25\stdv{2.48} & 39.64\stdv{0.67} & \textbf{4.67}\stdv{1.05} & 0.00\stdv{0.00} & 0.88\stdv{0.00} & \textbf{0.93}\stdv{0.00} \\
    3D Double Integrator (6, 3) & -- & 0.79\stdv{0.71} & \textbf{0.75}\stdv{0.87} & 84.35\stdv{3.09} & \textbf{4.62}\stdv{0.77} & 4.97\stdv{1.04} & 0.00\stdv{0.00} & \textbf{0.96}\stdv{0.01} & \textbf{0.96}\stdv{0.01} \\
    3-DOF Manipulator (6, 3) & 40.01\stdv{32.95} & 0.48\stdv{0.10} & \textbf{0.44}\stdv{0.05} & \textbf{0.56}\stdv{0.29} & 7.31\stdv{0.18} & 5.88\stdv{0.05} & 0.43\stdv{0.24} & 0.65\stdv{0.00} & \textbf{0.66}\stdv{0.00} \\
    Landing Rocket (7, 2) & -- & 80.40\stdv{9.66} & \textbf{1.10}\stdv{0.41} & \textbf{0.00}\stdv{0.00} & 2.22\stdv{1.40} & 6.01\stdv{0.14} & 0.00\stdv{0.00} & 0.01\stdv{0.01} & \textbf{0.24}\stdv{0.00} \\
    Quadruped Trunk (9, 4) & -- & -- & \textbf{15.62}\stdv{14.32} & \textbf{0.00}\stdv{0.00} & 1.96\stdv{1.79} & 2.47\stdv{0.28} & 0.00\stdv{0.00} & 0.00\stdv{0.00} & \textbf{0.06}\stdv{0.01} \\
    6-DoF Underw. Veh. (12, 6) & -- & \textbf{1.92}\stdv{0.15} & 3.24\stdv{0.16} & \textbf{1.65}\stdv{0.32} & 5.38\stdv{0.03} & 2.49\stdv{0.04} & 0.00\stdv{0.00} & 0.33\stdv{0.00} & \textbf{0.35}\stdv{0.00} \\
    Quadrotor (13, 4) & -- & 6.09\stdv{2.48} & \textbf{6.04}\stdv{4.17} & 82.36\stdv{0.62} & \textbf{6.93}\stdv{0.50} & 7.78\stdv{0.57} & 0.00\stdv{0.00} & \textbf{0.87}\stdv{0.02} & \textbf{0.87}\stdv{0.04} \\
    \bottomrule
  \end{tabularx}
\end{table*}

\section{Sensitivity to the PDE-Loss Weight}
\label{app:ablation_pde_weight}

The weight $\lambda$ in \cref{eq:total_loss} balances the PDE loss and the data loss. We vary it on the inverted pendulum, keeping every other setting of \cref{tab:exp_ip} and training three seeds per value. The two ends of the range are the limiting cases, because $\lambda = 0$ trains on the supervision data alone and $\lambda = 1$ reduces the method to PDE-only learning.

\begin{figure}[ht]
  \centering
  \includegraphics[width=\textwidth]{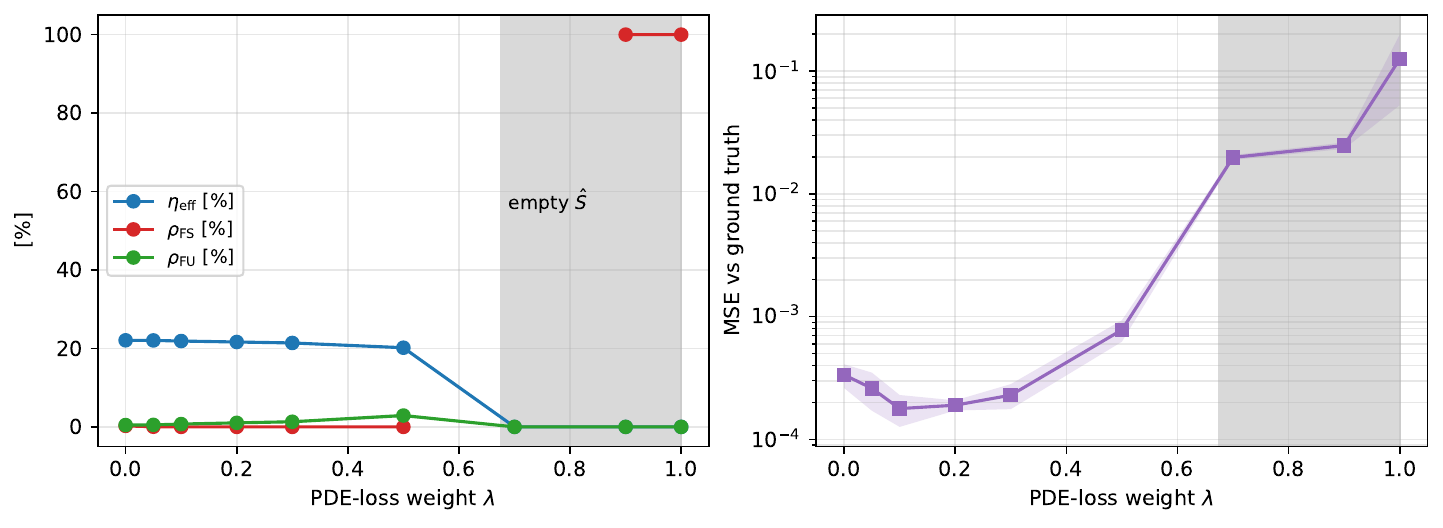}
  \caption{Sensitivity to the PDE-loss weight $\lambda$ on the inverted pendulum. The results show validation metrics (left) and mean squared error against the ground-truth value function (right). Mean values are obtained over three seeds and the bands are one standard deviation. In the shaded region $\eta_\mathrm{eff}$ is zero and $\rho_\mathrm{FS}$ is undefined for $\lambda = 0.7$ where the predicted safe set is empty for every seed.}
  \label{fig:pde_weight_ip}
\end{figure}

The results are shown in \cref{fig:pde_weight_ip}. The method is not significantly sensitive to $\lambda$ over the entire range from $0$ to $0.5$. The effective safe volume stays between $20.2\%$ and $22.1\%$, the false-safe rate stays below $0.3\%$, and the error against the ground truth varies between $1.8 \cdot 10^{-4}$ and $7.8 \cdot 10^{-4}$. The smallest error is reached near $\lambda = 0.1$, and the value $\lambda = 0.2$ used for this system lies inside the flat region. The behavior changes abruptly above $\lambda = 0.5$. At $\lambda = 0.7$ the predicted safe set is empty for every seed, and at $\lambda = 0.9$ and $\lambda = 1$ it is non-empty but every predicted-safe state is false-safe. The error against the ground truth grows by more than two orders of magnitude over the same range. The data term is therefore what keeps the optimization away from the trivial solutions of the PDE loss, while a moderate weight on the PDE term is enough to enforce local consistency without dominating the objective. We selected $\lambda$ per system in the same way, by choosing a value inside the range where the validation metrics are flat.

\section{Comparison with the Ground Truth}
\label{app:iou}

The metrics of \cref{tab:num_exp_results} are obtained from closed-loop rollouts, so they can be computed for every system. For seven of the fifteen systems, a reference value function computed on a state-space grid is also available\footnote{The reference value functions are computed with the \texttt{hj\_reachability} toolbox available at \url{https://github.com/StanfordASL/hj_reachability}.}.
These are systems with at most four states, because the memory and the runtime of a grid-based solver grow exponentially with the state dimension. For the remaining systems, a grid solution is not feasible on our hardware.

For these seven systems, we report the intersection over union metric defined as
\begin{equation*}
  \mathrm{IoU}
  =
  \frac{\lvert \hat{\mathcal{S}} \cap \mathcal{S} \rvert}
       {\lvert \hat{\mathcal{S}} \cup \mathcal{S} \rvert},
\end{equation*}
where $\hat{\mathcal{S}} = \{x : V_\Theta(x) \geq 0\}$ is the predicted safe set, $\mathcal{S} = \{x : V(x) \geq 0\}$ is the ground-truth safe set, and both are evaluated on the grid points of the reference solution. The IoU penalizes a predicted safe set that is too large and one that is too small, so it complements $\rho_\mathrm{FS}$ and $\eta_\mathrm{eff}$, which separate the two effects.

The results are given in \cref{tab:iou}. VRCD reaches the highest IoU on five of the seven systems and stays within $0.3$ percentage points of the best value on the other two. The three methods are almost identical on the systems that all of them solve well, such as the Dubins car and the 2D double integrator. The differences are large on the inverted pendulum, the vertical drone and the cart-pole, where FCD predicts an empty safe set (vertical drone, cart-pole) or only a small subset of the true one (inverted pendulum), and PDE-only recovers only a part of it.

\begin{table}[ht]
  \centering
  \caption{Intersection over union (\%) between the predicted safe set $\hat{\mathcal{S}}$ and the ground-truth safe set $\mathcal{S}$, evaluated on the ground-truth grid of every system for which it is available. Entries are mean $\pm$ standard deviation over five training seeds.}
  \vspace{6pt}
  \label{tab:iou}
  \small
  \providecommand{\stdv}[1]{\,\scalebox{0.75}{$\pm$#1}}
  \begin{tabular}{@{}lccc@{}}
    \toprule
    System & ND & FCD & VRCD (ours) \\
    \midrule
    Inverted Pendulum & 1.54\stdv{3.07} & 33.53\stdv{0.61} & \textbf{96.39}\stdv{1.16} \\
    1D Double Integrator & \textbf{98.37}\stdv{0.72} & 94.02\stdv{0.43} & 98.10\stdv{0.35} \\
    2D Vertical Drone & 64.09\stdv{32.05} & 0.00\stdv{0.00} & \textbf{98.48}\stdv{0.20} \\
    Dubins Car & 99.56\stdv{0.15} & 99.66\stdv{0.04} & \textbf{99.80}\stdv{0.03} \\
    2D Double Integrator & 99.52\stdv{0.03} & 99.39\stdv{0.02} & \textbf{99.64}\stdv{0.02} \\
    Kinematic Bicycle & 97.71\stdv{0.24} & \textbf{99.65}\stdv{0.00} & 99.59\stdv{0.00} \\
    Cart-Pole & 65.07\stdv{0.22} & 0.00\stdv{0.00} & \textbf{93.38}\stdv{0.35} \\
    \bottomrule
  \end{tabular}
\end{table}

\section{Details on Hardware Experiments}
\label{app:details_hardw_exp}

The hardware experiments are conducted on a mobile delivery robot equipped with a 3D LiDAR and stereo cameras. All computation and data processing, including object recognition and tracking, robot localization, and QP optimization, are performed on the onboard computer\footnote{An Intel NUC11PHi7 with an 11th generation Intel Core i7-1165G7 processor at $2.80$\,GHz, $16$\,GB of RAM, and an NVIDIA GeForce RTX 2060 graphics card.}.

\begin{figure}[ht]
  \centering
  \includegraphics[width=\textwidth]{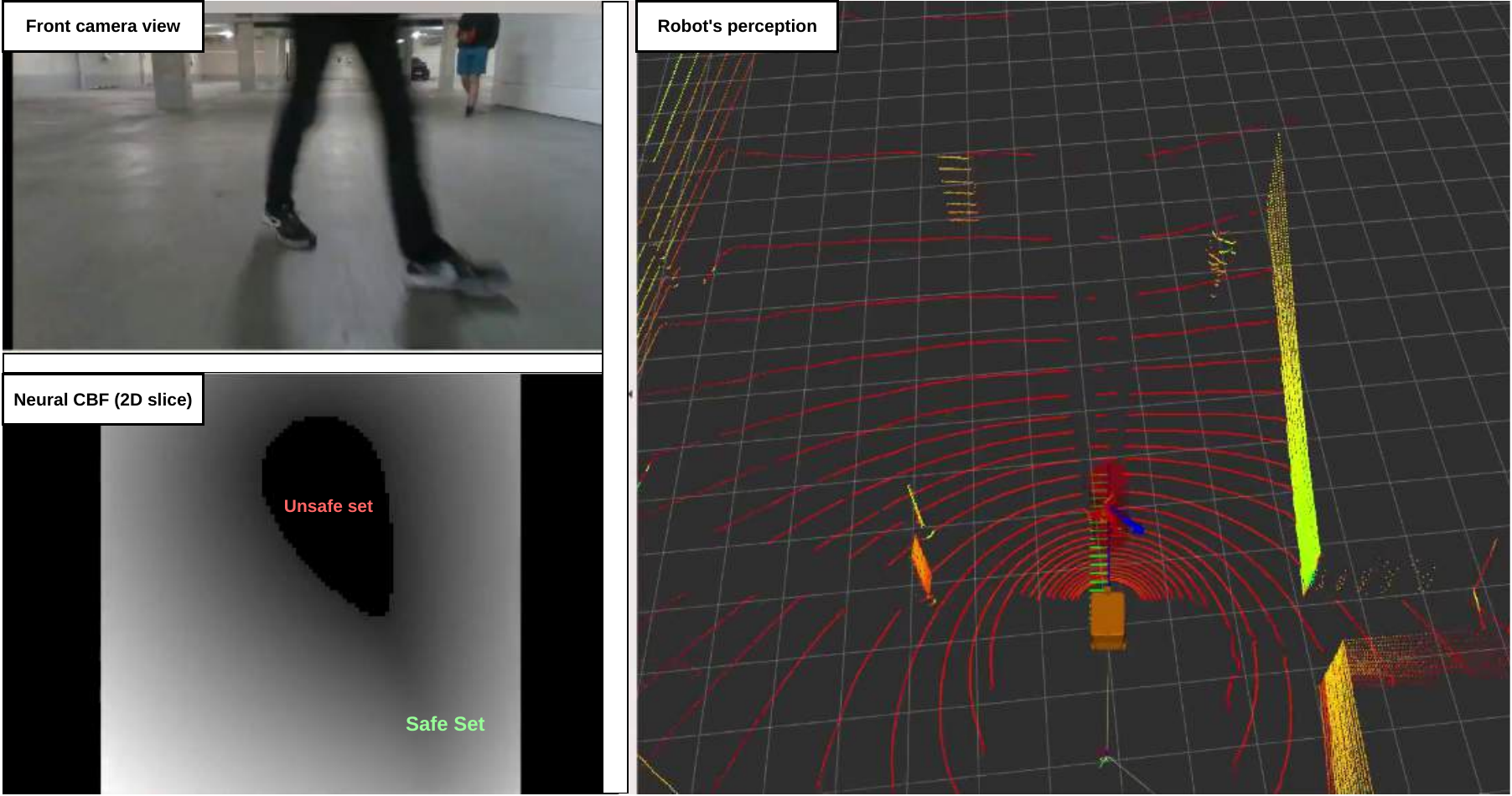}
  \caption{Screenshot from the onboard computer of the mobile robot used in the hardware experiments. The screenshot shows the front camera view, a 2D slice of the neural CBF in positional coordinates, and a visualization from the perception module.}
  \label{fig:hardw_exp_screen}
\end{figure}

During the hardware experiments, we conduct 10 trials in which the robot navigates from the start position to the goal position while avoiding pedestrians. In all trials, the robot successfully avoids collisions with pedestrians using the neural CBF trained with the approach proposed in this paper and deployed as in \cite{derajic_cn-cbf_2026}. As the extended class-$\mathcal{K}$ function, we use $\alpha(h(x)) = k h(x)$ with $k = 2.8$. Details on the relative system dynamics, constraint function, state limits, and control limits are provided in \cref{app:rel_unicycle}. In \cref{fig:hardw_exp_screen}, we show a screenshot from the onboard computer of the mobile robot used in the experiments.

The CBF is evaluated for the relative state between the robot and each tracked pedestrian, and then aggregated into a single \emph{composite} CBF constraint added to the QP. As it is standard in practice, the constraint is relaxed by a slack variable, which is penalized in the QP cost. The relaxation keeps the QP feasible even when the perception module reports a new pedestrian at a state where the CBF constraint cannot be satisfied by any admissible control. When a slack variable becomes active, the corresponding constraint is violated and forward invariance of the safe set is no longer guaranteed.

Two properties of the deployment explain the small negative CBF values in \cref{fig:hardw_exp}. First, the learned CBF assumes the ideal relative unicycle model of \cref{app:rel_unicycle}, while the real platform is a skid-steer robot whose wheels slip during turns and whose velocity controller has its own dynamics. The pedestrian states are estimated from LiDAR and camera measurements, so they are noisy and slightly delayed, and the constant-velocity assumption of the relative model only holds approximately. The QP also runs at a finite rate, so the control is held constant between two updates. Second, the radii $r_{\text r}$ and $r_{\text o}$ used during training are inflated by $5$\,cm with respect to the physical dimensions of the robot and of a pedestrian.
The learned safe set is therefore slightly smaller than the collision-free set of the real system, which absorbs the effects listed above. This is why the robot completes all trials without a collision even though the CBF value becomes negative for short periods.

\section{Details on Numerical Experiments}
\label{app:details_num_exp}

% -----------------------------------------------------------------------------
\subsection{Inverted Pendulum}
\label{app:ip}
% -----------------------------------------------------------------------------

\begin{figure}[ht]
  \centering
  \includegraphics[width=0.55\textwidth]{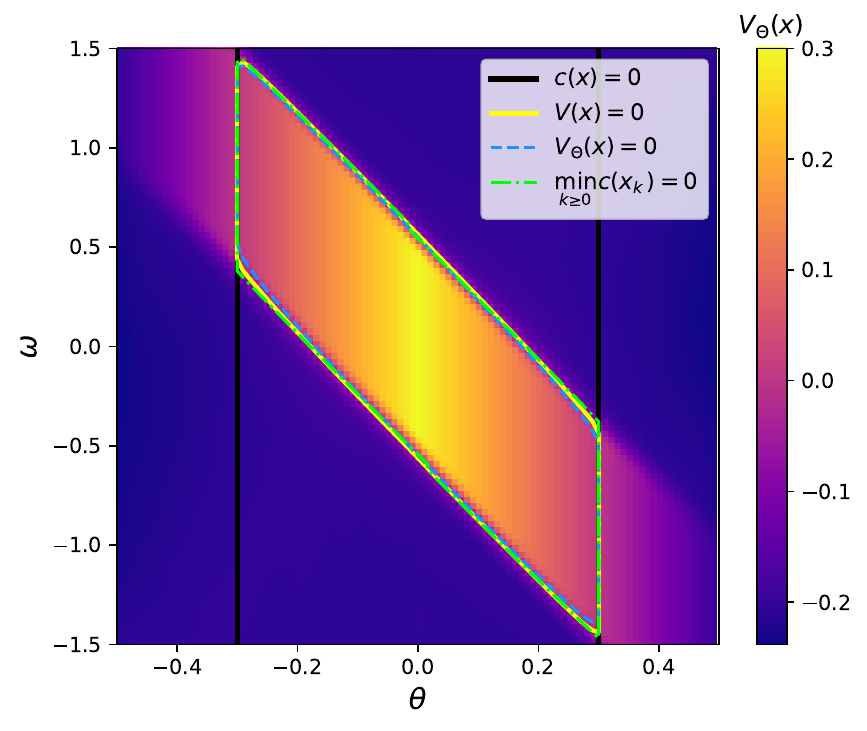}
  \caption{2D slice of the learned neural CBF for the inverted pendulum in the $(\theta,\omega)$ plane.}
  \label{fig:slice_ip}
\end{figure}

The state and control are
\begin{equation*}
  x = \begin{bmatrix} \theta & \omega \end{bmatrix}^\top,
  \qquad u = \tau,
\end{equation*}
where $\theta$ is the pole angle from the upright equilibrium, $\omega$ the angular velocity, and $\tau$ the applied pivot torque.  The dynamics are
\begin{equation*}
  f(x) = \begin{bmatrix} \omega \\[2pt] \dfrac{g}{l}\sin\theta \end{bmatrix},
  \qquad
  g(x) = \begin{bmatrix} 0 \\[2pt] \dfrac{1}{m\,l^{2}} \end{bmatrix}.
\end{equation*}
The constraint function bounds the pole angle:
\begin{equation*}
  c(x) = d - |\theta|.
\end{equation*}
See \cref{tab:exp_ip} for the configuration and \cref{fig:slice_ip} for a 2D CBF slice.

\begin{table}[ht]
  \centering
  \caption{Configuration --- Inverted Pendulum.}
  \label{tab:exp_ip}
  \footnotesize
  \begin{tabular}{@{}llr@{}}
    \toprule
    \multirow{5}{*}{Dynamics}
      & $x_{\min}$       & $[-0.5,\;-1.5]$ \\
      & $x_{\max}$       & $[\phantom{-}0.5,\;\phantom{-}1.5]$ \\
      & $u_{\min}$       & $-3.5$ \\
      & $u_{\max}$       & $\phantom{-}3.5$ \\
      & Model params     & $m{=}2.0$,\; $l{=}1.0$,\; $g{=}9.81$ \\
    \midrule
    \multirow{1}{*}{Constraint}
      & $d$              & $0.3$ \\
    \midrule
    \multirow{7}{*}{Data}
      & Method           & Beam search \\
      & \# samples       & $60\times 60$ (grid) \\
      & $K$              & $20$ \\
      & $B$              & $500$ \\
      & $\Delta t$       & $0.1$\,s \\
      & extra params     & --- \\
      & $T_{\text{data}}$ & $1$\,s \\
    \midrule
    \multirow{2}{*}{Model}
      & Hidden layers    & $5\times 32$, $\sin$ \\
      & Output           & softplus ($\beta=10$) \\
    \midrule
    \multirow{6}{*}{Training}
      & Epochs           & $10{,}000$ \\
      & Learning rate    & $10^{-3}$ \\
      & LR schedule      & $\times 0.1$ at epoch $7{,}000$ \\
      & PDE \# samples   & $10{,}000$ \\
      & PDE weight       & $0.2$ (fixed) \\
      & $T_{\text{train}}$ & $1$\,m\,$0$\,s \\
    \midrule
    \multirow{4}{*}{Validation}
      & Horizon $T$      & $5.0$\,s \\
      & $\Delta t$       & $0.01$\,s \\
      & \# samples       & $20{,}000$ \\
      & $T_{\text{valid}}$ & $1$\,s \\
    \bottomrule
  \end{tabular}
\end{table}

% -----------------------------------------------------------------------------
\subsection{1D Double Integrator}
\label{app:di1d}
% -----------------------------------------------------------------------------

\begin{figure}[ht]
  \centering
  \includegraphics[width=0.55\textwidth]{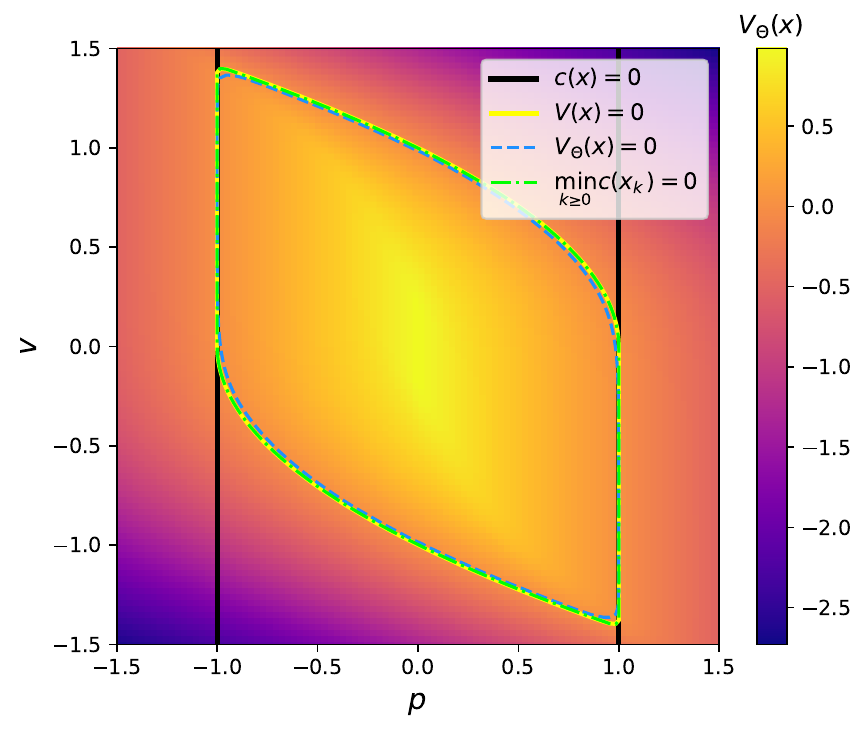}
  \caption{2D slice of the learned neural CBF for the 1D double integrator in the $(p, v)$ plane.}
  \label{fig:slice_di1d}
\end{figure}

The state and control are
\begin{equation*}
  x = \begin{bmatrix} p & v \end{bmatrix}^\top,
  \qquad u = a,
\end{equation*}
where $p$ is the position, $v$ the velocity, and $a$ the commanded acceleration.  The dynamics are
\begin{equation*}
  f(x) = \begin{bmatrix} v \\ 0 \end{bmatrix},
  \qquad
  g(x) = \begin{bmatrix} 0 \\ 1 \end{bmatrix}.
\end{equation*}
The constraint function bounds the position around an interval center $p_{\text c}$:
\begin{equation*}
  c(x) = d - |p - p_{\text c}|.
\end{equation*}
See \cref{tab:exp_di1d} for the configuration and \cref{fig:slice_di1d} for a 2D CBF slice.

\begin{table}[ht]
  \centering
  \caption{Configuration --- 1D Double Integrator.}
  \label{tab:exp_di1d}
  \footnotesize
  \begin{tabular}{@{}llr@{}}
    \toprule
    \multirow{5}{*}{Dynamics}
      & $x_{\min}$       & $[-1.5,\;-1.5]$ \\
      & $x_{\max}$       & $[\phantom{-}1.5,\;\phantom{-}1.5]$ \\
      & $u_{\min}$       & $-0.5$ \\
      & $u_{\max}$       & $\phantom{-}0.5$ \\
      & Model params     & --- \\
    \midrule
    \multirow{2}{*}{Constraint}
      & $p_{\text c}$    & $0.0$ \\
      & $d$              & $1.0$ \\
    \midrule
    \multirow{7}{*}{Data}
      & Method           & Stochastic beam search \\
      & \# samples       & $50\times 50$ (grid) \\
      & $K$              & $40$ \\
      & $B$              & $1{,}500$ \\
      & $\Delta t$       & $0.1$\,s \\
      & extra params     & $T{=}0.05$ \\
      & $T_{\text{data}}$ & $1$\,s \\
    \midrule
    \multirow{2}{*}{Model}
      & Hidden layers    & $4\times 32$, $\sin$ \\
      & Output           & softplus ($\beta=1$) \\
    \midrule
    \multirow{6}{*}{Training}
      & Epochs           & $10{,}000$ \\
      & Learning rate    & $10^{-3}$ \\
      & LR schedule      & $\times 0.1$ at epoch $8{,}000$ \\
      & PDE \# samples   & $10{,}000$ \\
      & PDE weight       & $0.9$ (fixed) \\
      & $T_{\text{train}}$ & $51$\,s \\
    \midrule
    \multirow{4}{*}{Validation}
      & Horizon $T$      & $4.0$\,s \\
      & $\Delta t$       & $0.01$\,s \\
      & \# samples       & $20{,}000$ \\
      & $T_{\text{valid}}$ & $1$\,s \\
    \bottomrule
  \end{tabular}
\end{table}

% -----------------------------------------------------------------------------
\subsection{2D Vertical Drone}
\label{app:vd2d}
% -----------------------------------------------------------------------------

\begin{figure}[ht]
  \centering
  \includegraphics[width=0.55\textwidth]{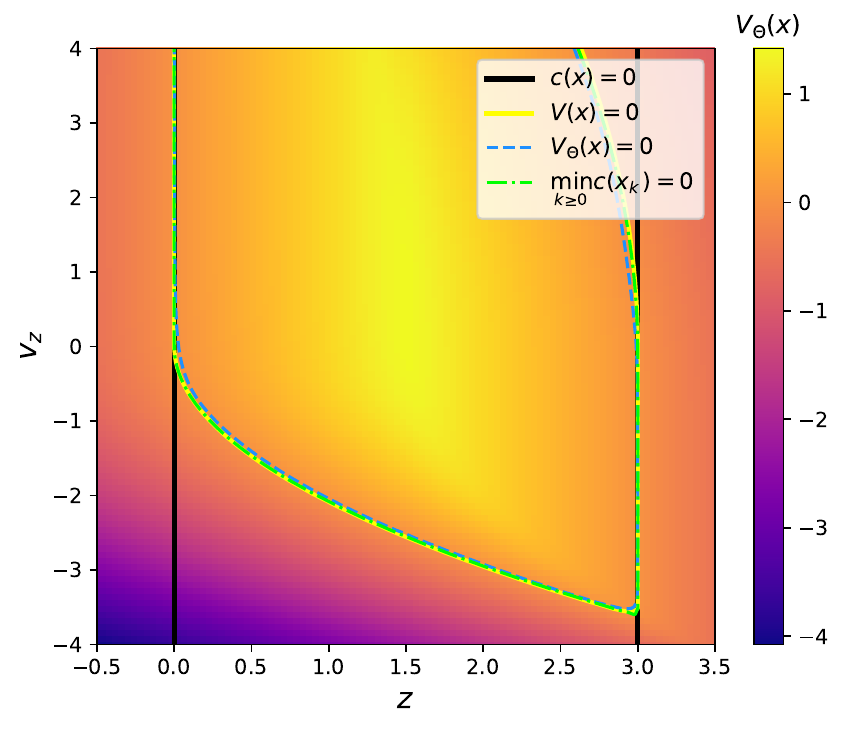}
  \caption{2D slice of the learned neural CBF for the 2D vertical drone in the $(z, v_z)$ plane.}
  \label{fig:slice_vd2d}
\end{figure}

The state and control are
\begin{equation*}
  x = \begin{bmatrix} z & v_z \end{bmatrix}^\top,
  \qquad u = a,
\end{equation*}
where $z$ is the altitude, $v_z$ the vertical velocity, and $a$ the commanded acceleration (scaled by a thrust gain $K$). The dynamics are
\begin{equation*}
  f(x) = \begin{bmatrix} v_z \\ -g \end{bmatrix},
  \qquad
  g(x) = \begin{bmatrix} 0 \\ K \end{bmatrix}.
\end{equation*}
The constraint function enforces an altitude band $z\in[z_{\text c} - d,\, z_{\text c} + d]$:
\begin{equation*}
  c(x) = d - |z - z_{\text c}|.
\end{equation*}
See \cref{tab:exp_vd2d} for the configuration and \cref{fig:slice_vd2d} for a 2D CBF slice.

\begin{table}[ht]
  \centering
  \caption{Configuration --- 2D Vertical Drone.}
  \label{tab:exp_vd2d}
  \footnotesize
  \begin{tabular}{@{}llr@{}}
    \toprule
    \multirow{5}{*}{Dynamics}
      & $x_{\min}$       & $[-0.5,\;-4.0]$ \\
      & $x_{\max}$       & $[\phantom{-}3.5,\;\phantom{-}4.0]$ \\
      & $u_{\min}$       & $-1.0$ \\
      & $u_{\max}$       & $\phantom{-}1.0$ \\
      & Model params     & $K{=}12.0$,\; $g{=}9.81$ \\
    \midrule
    \multirow{2}{*}{Constraint}
      & $z_{\text c}$    & $1.5$ \\
      & $d$              & $1.5$ \\
    \midrule
    \multirow{7}{*}{Data}
      & Method           & Branch-and-bound \\
      & \# samples       & $80\times 80$ (grid) \\
      & $K$              & $60$ \\
      & $B$              & $2{,}000$ \\
      & $\Delta t$       & $0.05$\,s \\
      & extra params     & $n_{\mathrm{restarts}}=2$ \\
      & $T_{\text{data}}$ & $5$\,s \\
    \midrule
    \multirow{2}{*}{Model}
      & Hidden layers    & $4\times 32$, $\sin$ \\
      & Output           & softplus ($\beta=10$) \\
    \midrule
    \multirow{6}{*}{Training}
      & Epochs           & $10{,}000$ \\
      & Learning rate    & $10^{-3}$ \\
      & LR schedule      & $\times 0.1$ at epoch $7{,}000$ \\
      & PDE \# samples   & $10{,}000$ \\
      & PDE weight       & $0.55$ (fixed) \\
      & $T_{\text{train}}$ & $52$\,s \\
    \midrule
    \multirow{4}{*}{Validation}
      & Horizon $T$      & $5.0$\,s \\
      & $\Delta t$       & $0.01$\,s \\
      & \# samples       & $20{,}000$ \\
      & $T_{\text{valid}}$ & $1$\,s \\
    \bottomrule
  \end{tabular}
\end{table}

% -----------------------------------------------------------------------------
\subsection{Dubins Car}
\label{app:dubins}
% -----------------------------------------------------------------------------

\begin{figure}[ht]
  \centering
  \includegraphics[width=0.55\textwidth]{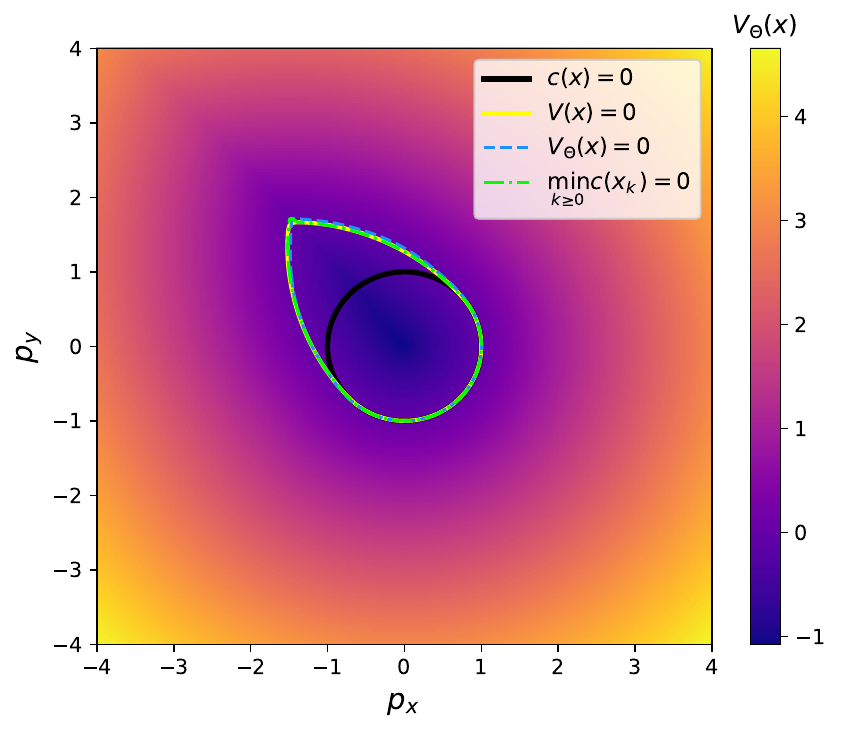}
  \caption{2D slice of the learned neural CBF for the Dubins car in the $(p_x, p_y)$ plane with the heading fixed at $\theta = -0.8377$\,rad.}
  \label{fig:slice_dubins}
\end{figure}

The state and control are
\begin{equation*}
  x = \begin{bmatrix} p_x & p_y & \theta \end{bmatrix}^\top,
  \qquad u = \omega,
\end{equation*}
where $(p_x, p_y)$ is the position, $\theta$ the heading angle, and $\omega$ the turning rate.  The forward speed $v$ is a fixed parameter.  The dynamics are
\begin{equation*}
  f(x) = \begin{bmatrix} v\cos\theta \\ v\sin\theta \\ 0 \end{bmatrix},
  \qquad
  g(x) = \begin{bmatrix} 0 \\ 0 \\ 1 \end{bmatrix}.
\end{equation*}
The constraint function is a circular obstacle of radius $r$ centered at the origin:
\begin{equation*}
  c(x) = \sqrt{p_x^{2} + p_y^{2}} - r.
\end{equation*}
See \cref{tab:exp_dubins} for the configuration and \cref{fig:slice_dubins} for a 2D CBF slice.

\begin{table}[ht]
  \centering
  \caption{Configuration --- Dubins Car.}
  \label{tab:exp_dubins}
  \footnotesize
  \begin{tabular}{@{}llr@{}}
    \toprule
    \multirow{5}{*}{Dynamics}
      & $x_{\min}$       & $[-4.0,\;-4.0,\;-3.14]$ \\
      & $x_{\max}$       & $[\phantom{-}4.0,\;\phantom{-}4.0,\;\phantom{-}3.14]$ \\
      & $u_{\min}$       & $-0.5$ \\
      & $u_{\max}$       & $\phantom{-}0.5$ \\
      & Model params     & $v{=}1.0$ \\
    \midrule
    \multirow{1}{*}{Constraint}
      & $r$              & $1.0$ \\
    \midrule
    \multirow{7}{*}{Data}
      & Method           & Beam search \\
      & \# samples       & $50{,}000$ \\
      & $K$              & $150$ \\
      & $B$              & $1{,}000$ \\
      & $\Delta t$       & $0.05$\,s \\
      & extra params     & --- \\
      & $T_{\text{data}}$ & $27$\,s \\
    \midrule
    \multirow{2}{*}{Model}
      & Hidden layers    & $4\times 32$, $\sin$ \\
      & Output           & softplus ($\beta=10$) \\
    \midrule
    \multirow{6}{*}{Training}
      & Epochs           & $10{,}000$ \\
      & Learning rate    & $10^{-3}$ \\
      & LR schedule      & $\times 0.1$ at epoch $8{,}000$ \\
      & PDE \# samples   & $200{,}000$ \\
      & PDE weight       & $0.8$ (fixed) \\
      & $T_{\text{train}}$ & $2$\,m\,$15$\,s \\
    \midrule
    \multirow{4}{*}{Validation}
      & Horizon $T$      & $5.0$\,s \\
      & $\Delta t$       & $0.01$\,s \\
      & \# samples       & $200{,}000$ \\
      & $T_{\text{valid}}$ & $3$\,s \\
    \bottomrule
  \end{tabular}
\end{table}

% -----------------------------------------------------------------------------
\subsection{2D Double Integrator}
\label{app:di2d}
% -----------------------------------------------------------------------------

\begin{figure}[ht]
  \centering
  \includegraphics[width=0.55\textwidth]{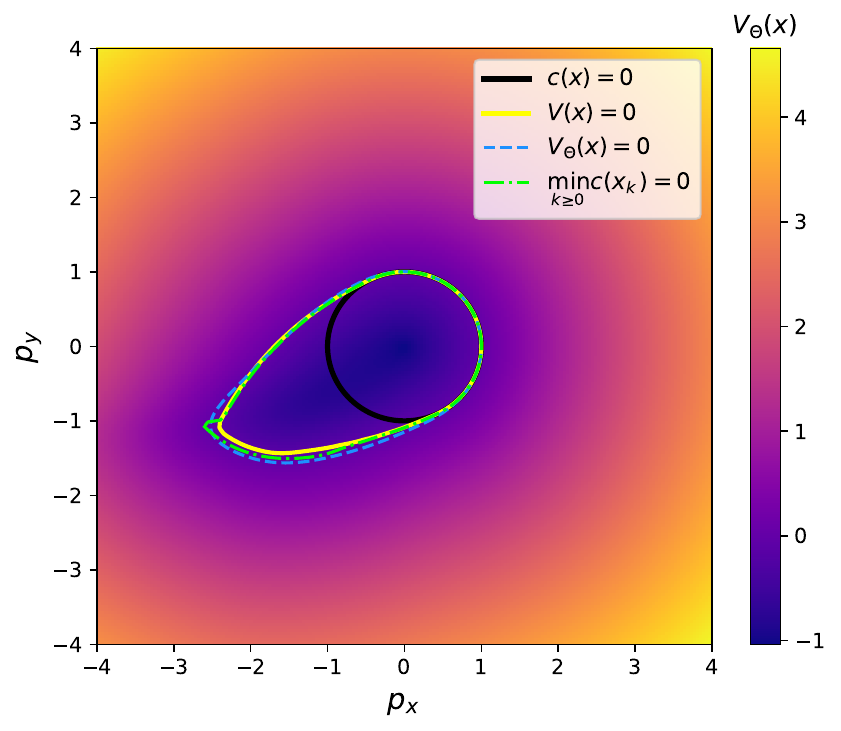}
  \caption{2D slice of the learned neural CBF for the 2D double integrator in the $(p_x, p_y)$ plane with velocity fixed at $(v_x, v_y) = (2.0, 1.0)$.}
  \label{fig:slice_di2d}
\end{figure}

The state and control are
\begin{equation*}
  x = \begin{bmatrix} p_x & p_y & v_x & v_y \end{bmatrix}^\top,
  \qquad
  u = \begin{bmatrix} a_x & a_y \end{bmatrix}^\top,
\end{equation*}
with $(p_x, p_y)$ and $(v_x, v_y)$ the planar position and velocity and $(a_x, a_y)$ the commanded accelerations.  The dynamics are
\begin{equation*}
  f(x) = \begin{bmatrix} v_x \\ v_y \\ 0 \\ 0 \end{bmatrix},
  \qquad
  g(x) = \begin{bmatrix} 0 & 0 \\ 0 & 0 \\ 1 & 0 \\ 0 & 1 \end{bmatrix}.
\end{equation*}
The constraint function is a circular obstacle of radius $r$ centered at the origin:
\begin{equation*}
  c(x) = \sqrt{p_x^{2} + p_y^{2}} - r.
\end{equation*}
See \cref{tab:exp_di2d} for the configuration and \cref{fig:slice_di2d} for a 2D CBF slice.

\begin{table}[ht]
  \centering
  \caption{Configuration --- 2D Double Integrator.}
  \label{tab:exp_di2d}
  \footnotesize
  \begin{tabular}{@{}llr@{}}
    \toprule
    \multirow{5}{*}{Dynamics}
      & $x_{\min}$       & $[-4.0,\;-4.0,\;-2.0,\;-2.0]$ \\
      & $x_{\max}$       & $[\phantom{-}4.0,\;\phantom{-}4.0,\;\phantom{-}2.0,\;\phantom{-}2.0]$ \\
      & $u_{\min}$       & $[-1.0,\;-1.0]$ \\
      & $u_{\max}$       & $[\phantom{-}1.0,\;\phantom{-}1.0]$ \\
      & Model params     & --- \\
    \midrule
    \multirow{1}{*}{Constraint}
      & $r$              & $1.0$ \\
    \midrule
    \multirow{7}{*}{Data}
      & Method           & Branch-and-bound \\
      & \# samples       & $300{,}000$ \\
      & $K$              & $100$ \\
      & $B$              & $1{,}000$ \\
      & $\Delta t$       & $0.05$\,s \\
      & extra params     & $n_{\mathrm{restarts}}=3$ \\
      & $T_{\text{data}}$ & $8$\,m\,$2$\,s \\
    \midrule
    \multirow{2}{*}{Model}
      & Hidden layers    & $5\times 32$, $\sin$ \\
      & Output           & softplus ($\beta=10$) \\
    \midrule
    \multirow{6}{*}{Training}
      & Epochs           & $10{,}000$ \\
      & Learning rate    & $10^{-3}$ \\
      & LR schedule      & $\times 0.1$ at epoch $7{,}000$ \\
      & PDE \# samples   & $600{,}000$ \\
      & PDE weight       & normalized \\
      & $T_{\text{train}}$ & $6$\,m\,$46$\,s \\
    \midrule
    \multirow{4}{*}{Validation}
      & Horizon $T$      & $5.0$\,s \\
      & $\Delta t$       & $0.01$\,s \\
      & \# samples       & $1{,}000{,}000$ \\
      & $T_{\text{valid}}$ & $12$\,s \\
    \bottomrule
  \end{tabular}
\end{table}

% -----------------------------------------------------------------------------
\subsection{Kinematic Bicycle}
\label{app:bicycle}
% -----------------------------------------------------------------------------

\begin{figure}[ht]
  \centering
  \includegraphics[width=0.55\textwidth]{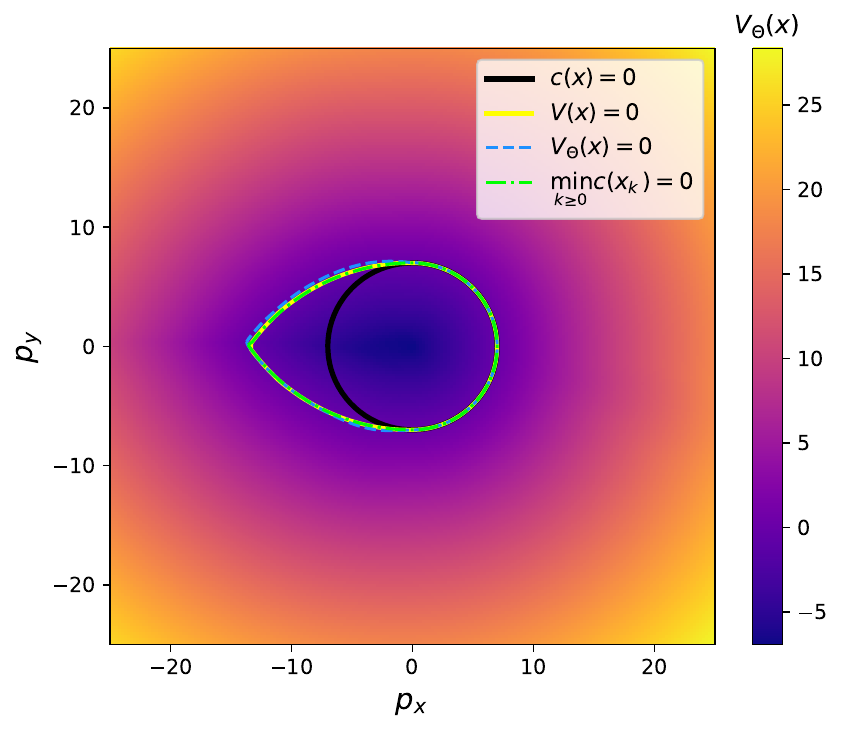}
  \caption{2D slice of the learned neural CBF for the kinematic bicycle in the $(p_x, p_y)$ plane with heading $\psi = 0.0$\,rad and speed $v = 20.0$\,m/s.}
  \label{fig:slice_bicycle}
\end{figure}

The state and control are
\begin{equation*}
  x = \begin{bmatrix} p_x & p_y & \psi & v \end{bmatrix}^\top,
  \qquad
  u = \begin{bmatrix} \tan\delta & a \end{bmatrix}^\top,
\end{equation*}
where $(p_x, p_y)$ is the rear-axle position, $\psi$ the heading, $v$ the longitudinal speed, $\delta$ the front-wheel steering angle, and $a$ the longitudinal acceleration.  We use $\tan\delta$ in place of $\delta$ to keep the dynamics control-affine.  With wheelbase $L$,
\begin{equation*}
  f(x) = \begin{bmatrix} v\cos\psi \\ v\sin\psi \\ 0 \\ 0 \end{bmatrix},
  \qquad
  g(x) = \begin{bmatrix} 0 & 0 \\ 0 & 0 \\ v/L & 0 \\ 0 & 1 \end{bmatrix}.
\end{equation*}
The constraint function is a circular obstacle of radius $r$ centered at the origin:
\begin{equation*}
  c(x) = \sqrt{p_x^{2} + p_y^{2}} - r.
\end{equation*}
See \cref{tab:exp_bicycle} for the configuration and \cref{fig:slice_bicycle} for a 2D CBF slice.

\begin{table}[ht]
  \centering
  \caption{Configuration --- Kinematic Bicycle.}
  \label{tab:exp_bicycle}
  \footnotesize
  \begin{tabular}{@{}llr@{}}
    \toprule
    \multirow{5}{*}{Dynamics}
      & $x_{\min}$       & $[-25.0,\;-25.0,\;-3.14,\;\phantom{-}0.0]$ \\
      & $x_{\max}$       & $[\phantom{-}25.0,\;\phantom{-}25.0,\;\phantom{-}3.14,\;20.0]$ \\
      & $u_{\min}$       & $[-0.3,\;-3.0]$ \\
      & $u_{\max}$       & $[\phantom{-}0.3,\;\phantom{-}3.0]$ \\
      & Model params     & $L{=}2.8$ \\
    \midrule
    \multirow{1}{*}{Constraint}
      & $r$              & $7.0$ \\
    \midrule
    \multirow{7}{*}{Data}
      & Method           & Beam search \\
      & \# samples       & $200{,}000$ \\
      & $K$              & $80$ \\
      & $B$              & $1{,}000$ \\
      & $\Delta t$       & $0.1$\,s \\
      & extra params     & --- \\
      & $T_{\text{data}}$ & $2$\,m\,$24$\,s \\
    \midrule
    \multirow{2}{*}{Model}
      & Hidden layers    & $32$-$64$-$64$-$32$, $\sin$ \\
      & Output           & softplus ($\beta=10$) \\
    \midrule
    \multirow{6}{*}{Training}
      & Epochs           & $10{,}000$ \\
      & Learning rate    & $10^{-3}$ \\
      & LR schedule      & $\times 0.1$ at epoch $7{,}000$ \\
      & PDE \# samples   & $500{,}000$ \\
      & PDE weight       & $0.15$ (fixed) \\
      & $T_{\text{train}}$ & $6$\,m\,$3$\,s \\
    \midrule
    \multirow{4}{*}{Validation}
      & Horizon $T$      & $5.0$\,s \\
      & $\Delta t$       & $0.01$\,s \\
      & \# samples       & $500{,}000$ \\
      & $T_{\text{valid}}$ & $6$\,s \\
    \bottomrule
  \end{tabular}
\end{table}

% -----------------------------------------------------------------------------
\subsection{Cart-Pole}
\label{app:cart_pole}
% -----------------------------------------------------------------------------

\begin{figure}[ht]
  \centering
  \includegraphics[width=0.55\textwidth]{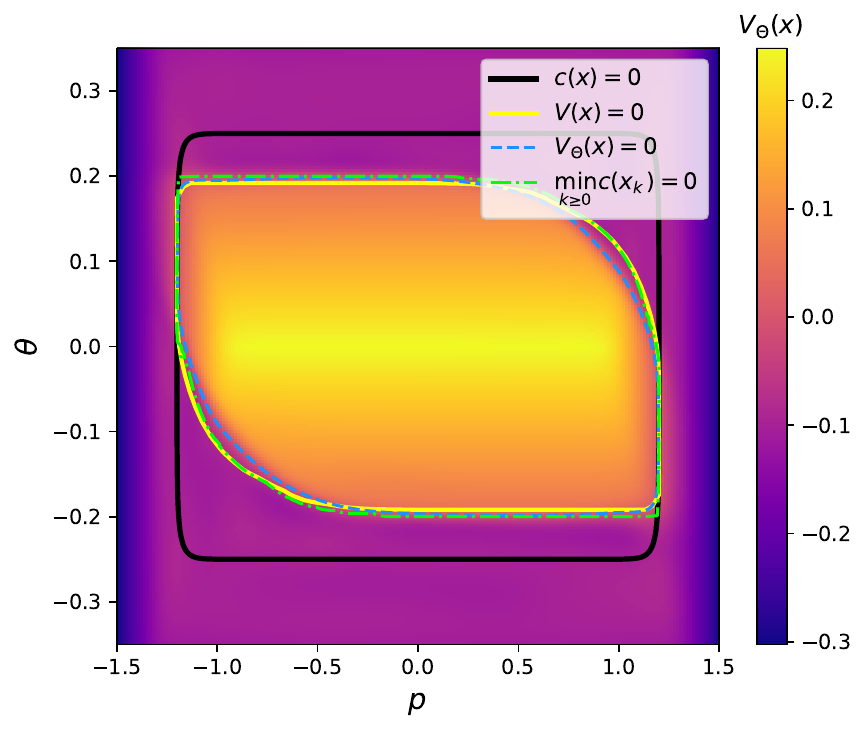}
  \caption{2D slice of the learned neural CBF for the cart-pole in the $(p, \theta)$ plane with velocities fixed at $\dot p = 0.0$ and $\dot\theta = 0.0$.}
  \label{fig:slice_cart_pole}
\end{figure}

The state and control vectors are
\begin{equation*}
  x = \begin{bmatrix} p & \theta & \dot{p} & \dot{\theta} \end{bmatrix}^\top,
  \qquad
  u = F,
\end{equation*}
where $p$ is the cart position, $\theta$ the pole angle from the upright equilibrium, and $F$ the horizontal force on the cart. With $D(\theta)=m_c + m_p\sin^2\!\theta$ we have:
\begin{equation*}
  f(x) =
  \begin{bmatrix}
    \dot{p}\\[2pt]
    \dot{\theta}\\[2pt]
    \dfrac{m_p\sin\theta\!\left(l\dot{\theta}^2 - g\cos\theta\right)}{D(\theta)}\\[8pt]
    \dfrac{(m_c+m_p)\,g\sin\theta - m_p\,l\,\dot{\theta}^2\cos\theta\sin\theta}{l\,D(\theta)}
  \end{bmatrix},
  \qquad
  g(x) =
  \begin{bmatrix}
    0\\[2pt]
    0\\[2pt]
    \dfrac{1}{D(\theta)}\\[8pt]
    \dfrac{-\cos\theta}{l\,D(\theta)}
  \end{bmatrix}.
\end{equation*}
The constraint function is a smooth box constraint on $p$ and $\theta$:
\begin{equation*}
  c(x) = -\frac{1}{\alpha}\log\!\Bigl(
      e^{-\alpha(p_{\max}-|p|)}
    + e^{-\alpha(\theta_{\max}-|\theta|)}
    \Bigr).
\end{equation*}
See \cref{tab:exp_cart_pole} for the configuration and \cref{fig:slice_cart_pole} for a 2D CBF slice.

\begin{table}[ht]
  \centering
  \caption{Configuration --- Cart-Pole.}
  \label{tab:exp_cart_pole}
  \footnotesize
  \begin{tabular}{@{}llr@{}}
    \toprule
    \multirow{5}{*}{Dynamics}
      & $x_{\min}$       & $[-1.5,\;-0.35,\;-1.0,\;-1.0]$ \\
      & $x_{\max}$       & $[\phantom{-}1.5,\;\phantom{-}0.35,\;\phantom{-}1.0,\;\phantom{-}1.0]$ \\
      & $u_{\min}$       & $-5.0$ \\
      & $u_{\max}$       & $\phantom{-}5.0$ \\
      & Model params     & $m_c{=}2.0$,\; $m_p{=}0.5$,\; $l{=}0.5$,\; $g{=}9.81$ \\
    \midrule
    \multirow{3}{*}{Constraint}
      & $p_{\max}$       & $1.2$ \\
      & $\theta_{\max}$  & $0.25$ \\
      & $\alpha$         & $30.0$ \\
    \midrule
    \multirow{7}{*}{Data}
      & Method           & Branch-and-bound \\
      & \# samples       & $400{,}000$ \\
      & $K$              & $50$ \\
      & $B$              & $500$ \\
      & $\Delta t$       & $0.05$\,s \\
      & extra params     & $n_{\mathrm{restarts}}=2$ \\
      & $T_{\text{data}}$ & $1$\,m\,$40$\,s \\
    \midrule
    \multirow{2}{*}{Model}
      & Hidden layers    & $32$-$64$-$64$-$64$-$32$, $\sin$ \\
      & Output           & softplus ($\beta=10$) \\
    \midrule
    \multirow{6}{*}{Training}
      & Epochs           & $10{,}000$ \\
      & Learning rate    & $10^{-3}$ \\
      & LR schedule      & $\times 0.1$ at epoch $8{,}000$ \\
      & PDE \# samples   & $800{,}000$ \\
      & PDE weight       & $0.0$ (fixed) \\
      & $T_{\text{train}}$ & $2$\,m\,$14$\,s \\
    \midrule
    \multirow{4}{*}{Validation}
      & Horizon $T$      & $3.0$\,s \\
      & $\Delta t$       & $0.002$\,s \\
      & \# samples       & $1{,}000{,}000$ \\
      & $T_{\text{valid}}$ & $51$\,s \\
    \bottomrule
  \end{tabular}
\end{table}

% -----------------------------------------------------------------------------
\subsection{Dynamic Unicycle}
\label{app:unicycle}
% -----------------------------------------------------------------------------

\begin{figure}[ht]
  \centering
  \includegraphics[width=0.55\textwidth]{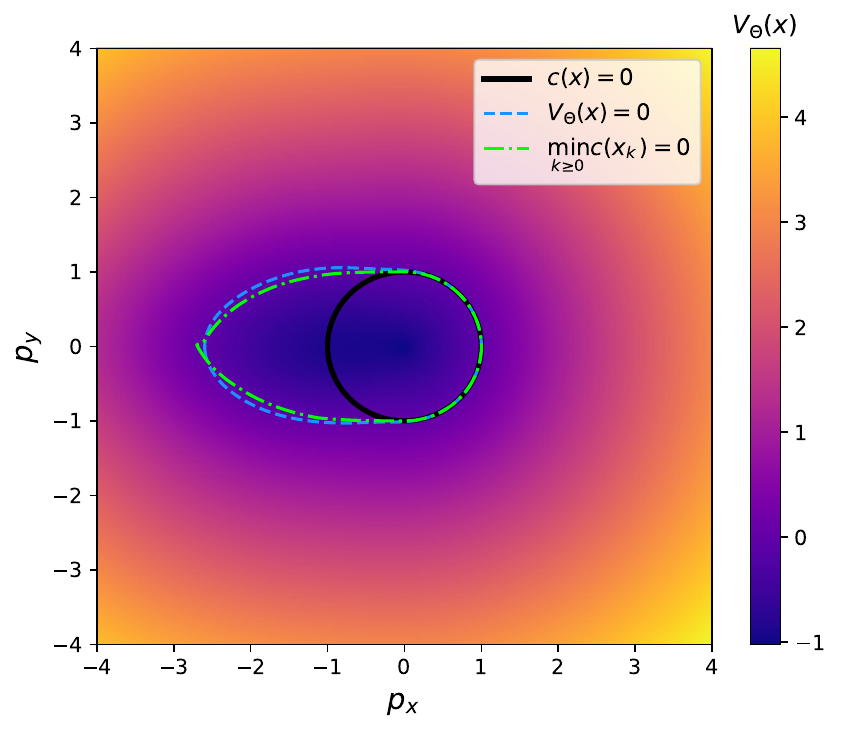}
  \caption{2D slice of the learned neural CBF for the dynamic unicycle in the $(p_x, p_y)$ plane with heading $\theta = 0.0$\,rad, linear speed $v = 2.0$, and angular velocity $\omega = 0.0$.}
  \label{fig:slice_unicycle}
\end{figure}

The state and control are
\begin{equation*}
  x = \begin{bmatrix} p_x & p_y & \theta & v & \omega \end{bmatrix}^\top,
  \qquad
  u = \begin{bmatrix} a & \alpha \end{bmatrix}^\top,
\end{equation*}
where $(p_x, p_y)$ is position, $\theta$ heading, $v$ longitudinal speed, $\omega$ angular velocity, and $(a, \alpha)$ are the linear and angular accelerations. The dynamics are
\begin{equation*}
  f(x) = \begin{bmatrix} v\cos\theta \\ v\sin\theta \\ \omega \\ 0 \\ 0 \end{bmatrix},
  \qquad
  g(x) = \begin{bmatrix} 0 & 0 \\ 0 & 0 \\ 0 & 0 \\ 1 & 0 \\ 0 & 1 \end{bmatrix}.
\end{equation*}
The constraint function is a circular obstacle of radius $r$ centered at the origin:
\begin{equation*}
  c(x) = \sqrt{p_x^{2} + p_y^{2}} - r.
\end{equation*}
See \cref{tab:exp_unicycle} for the configuration and \cref{fig:slice_unicycle} for a 2D CBF slice.

\begin{table}[ht]
  \centering
  \caption{Configuration --- Dynamic Unicycle.}
  \label{tab:exp_unicycle}
  \footnotesize
  \begin{tabular}{@{}llr@{}}
    \toprule
    \multirow{5}{*}{Dynamics}
      & $x_{\min}$       & $[-4.0,\;-4.0,\;-3.14,\;-2.0,\;-1.0]$ \\
      & $x_{\max}$       & $[\phantom{-}4.0,\;\phantom{-}4.0,\;\phantom{-}3.14,\;\phantom{-}2.0,\;\phantom{-}1.0]$ \\
      & $u_{\min}$       & $[-1.0,\;-1.0]$ \\
      & $u_{\max}$       & $[\phantom{-}1.0,\;\phantom{-}1.0]$ \\
      & Model params     & --- \\
    \midrule
    \multirow{1}{*}{Constraint}
      & $r$              & $1.0$ \\
    \midrule
    \multirow{7}{*}{Data}
      & Method           & Beam search \\
      & \# samples       & $600{,}000$ \\
      & $K$              & $50$ \\
      & $B$              & $1{,}000$ \\
      & $\Delta t$       & $0.1$\,s \\
      & extra params     & --- \\
      & $T_{\text{data}}$ & $3$\,s \\
    \midrule
    \multirow{2}{*}{Model}
      & Hidden layers    & $4\times 64$, $\sin$ \\
      & Output           & softplus ($\beta=10$) \\
    \midrule
    \multirow{6}{*}{Training}
      & Epochs           & $10{,}000$ \\
      & Learning rate    & $10^{-3}$ \\
      & LR schedule      & $\times 0.1$ at epoch $7{,}000$ \\
      & PDE \# samples   & $1{,}000{,}000$ \\
      & PDE weight       & $0.6$ (fixed) \\
      & $T_{\text{train}}$ & $15$\,m\,$44$\,s \\
    \midrule
    \multirow{4}{*}{Validation}
      & Horizon $T$      & $5.0$\,s \\
      & $\Delta t$       & $0.01$\,s \\
      & \# samples       & $1{,}000{,}000$ \\
      & $T_{\text{valid}}$ & $17$\,s \\
    \bottomrule
  \end{tabular}
\end{table}

% -----------------------------------------------------------------------------
\subsection{Relative Unicycle}
\label{app:rel_unicycle}
% -----------------------------------------------------------------------------

\begin{figure}[ht]
  \centering
  \includegraphics[width=0.55\textwidth]{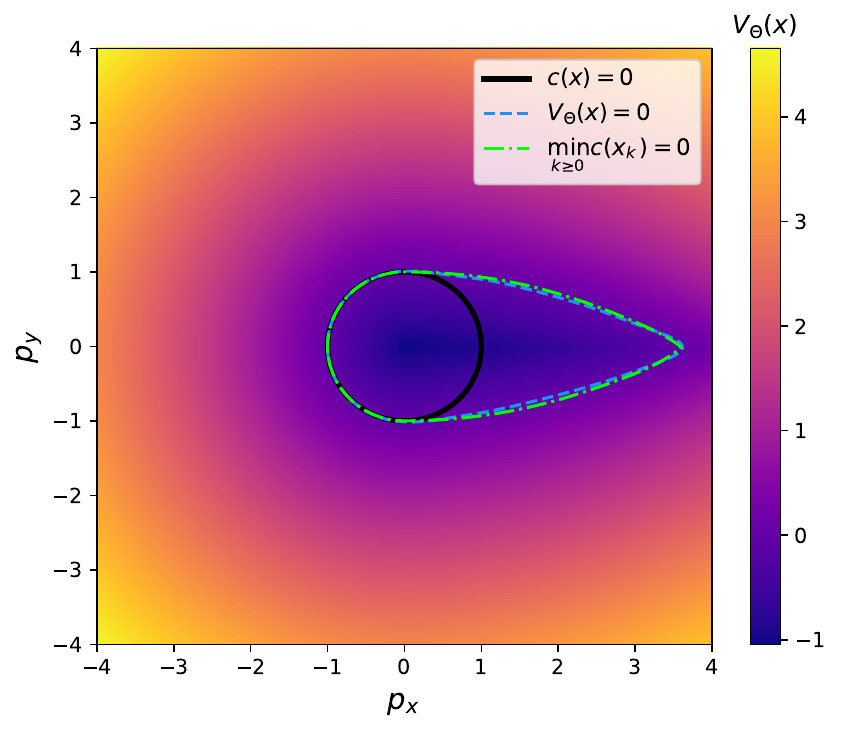}
  \caption{2D slice of the learned neural CBF for the relative unicycle in the $(p_x, p_y)$ plane with relative heading $\psi = \pi$\,rad, robot speed $v_r = 1.0$, and obstacle speed $v_p = 1.0$.}
  \label{fig:slice_rel_unicycle}
\end{figure}

This system models the relative dynamics between a dynamic-unicycle robot and a constant-velocity obstacle (e.g.~a pedestrian), expressed in the robot's body frame.  The obstacle speed $v_p$ is unknown and lifted into the state.  The state and control are
\begin{equation*}
  x = \begin{bmatrix} p_x & p_y & \psi & v_r & v_p \end{bmatrix}^\top,
  \qquad
  u = \begin{bmatrix} a & \omega \end{bmatrix}^\top,
\end{equation*}
where $(p_x, p_y)$ is the obstacle's position in the robot body frame, $\psi$ its heading relative to the robot, $v_r$ the robot's linear speed, $v_p$ the obstacle's linear speed, and $(a, \omega)$ the robot's linear acceleration and angular velocity.  The rotating body frame induces a Coriolis-like coupling that makes $g(x)$ state-dependent:
\begin{equation*}
  f(x) = \begin{bmatrix} v_p\cos\psi - v_r \\ v_p\sin\psi \\ 0 \\ 0 \\ 0 \end{bmatrix},
  \qquad
  g(x) = \begin{bmatrix} 0 & \phantom{-}p_y \\ 0 & -p_x \\ 0 & -1 \\ 1 & \phantom{-}0 \\ 0 & \phantom{-}0 \end{bmatrix}.
\end{equation*}
The constraint function is a two-disk collision check in the body frame between the robot disk (radius $r_{\text r}$) at the origin and the obstacle disk (radius $r_{\text o}$) at $(p_x, p_y)$:
\begin{equation*}
  c(x) = \sqrt{p_x^{2} + p_y^{2}} - (r_{\text r} + r_{\text o}).
\end{equation*}
See \cref{tab:exp_rel_unicycle} for the configuration and \cref{fig:slice_rel_unicycle} for a 2D CBF slice.

\begin{table}[ht]
  \centering
  \caption{Configuration --- Relative Unicycle.}
  \label{tab:exp_rel_unicycle}
  \footnotesize
  \begin{tabular}{@{}llr@{}}
    \toprule
    \multirow{5}{*}{Dynamics}
      & $x_{\min}$       & $[-4.0,\;-4.0,\;-3.14,\;\phantom{-}0.0,\;\phantom{-}0.0]$ \\
      & $x_{\max}$       & $[\phantom{-}4.0,\;\phantom{-}4.0,\;\phantom{-}3.14,\;\phantom{-}1.5,\;\phantom{-}1.5]$ \\
      & $u_{\min}$       & $[-1.0,\;-0.6]$ \\
      & $u_{\max}$       & $[\phantom{-}1.0,\;\phantom{-}0.6]$ \\
      & Model params     & --- \\
    \midrule
    \multirow{2}{*}{Constraint}
      & $r_{\text r}$    & $0.5$ \\
      & $r_{\text o}$    & $0.5$ \\
    \midrule
    \multirow{7}{*}{Data}
      & Method           & Beam search \\
      & \# samples       & $1{,}000{,}000$ \\
      & $K$              & $50$ \\
      & $B$              & $1{,}000$ \\
      & $\Delta t$       & $0.1$\,s \\
      & extra params     & --- \\
      & $T_{\text{data}}$ & $6$\,m\,$26$\,s \\
    \midrule
    \multirow{2}{*}{Model}
      & Hidden layers    & $5\times 32$, $\sin$ \\
      & Output           & softplus ($\beta=10$) \\
    \midrule
    \multirow{6}{*}{Training}
      & Epochs           & $10{,}000$ \\
      & Learning rate    & $10^{-3}$ \\
      & LR schedule      & $\times 0.1$ at epoch $8{,}000$ \\
      & PDE \# samples   & $2{,}000{,}000$ \\
      & PDE weight       & $0.3$ (fixed) \\
      & $T_{\text{train}}$ & $22$\,m\,$2$\,s \\
    \midrule
    \multirow{4}{*}{Validation}
      & Horizon $T$      & $5.0$\,s \\
      & $\Delta t$       & $0.005$\,s \\
      & \# samples       & $1{,}000{,}000$ \\
      & $T_{\text{valid}}$ & $28$\,s \\
    \bottomrule
  \end{tabular}
\end{table}

% -----------------------------------------------------------------------------
\subsection{3D Double Integrator}
\label{app:di3d}
% -----------------------------------------------------------------------------

\begin{figure}[ht]
  \centering
  \includegraphics[width=0.55\textwidth]{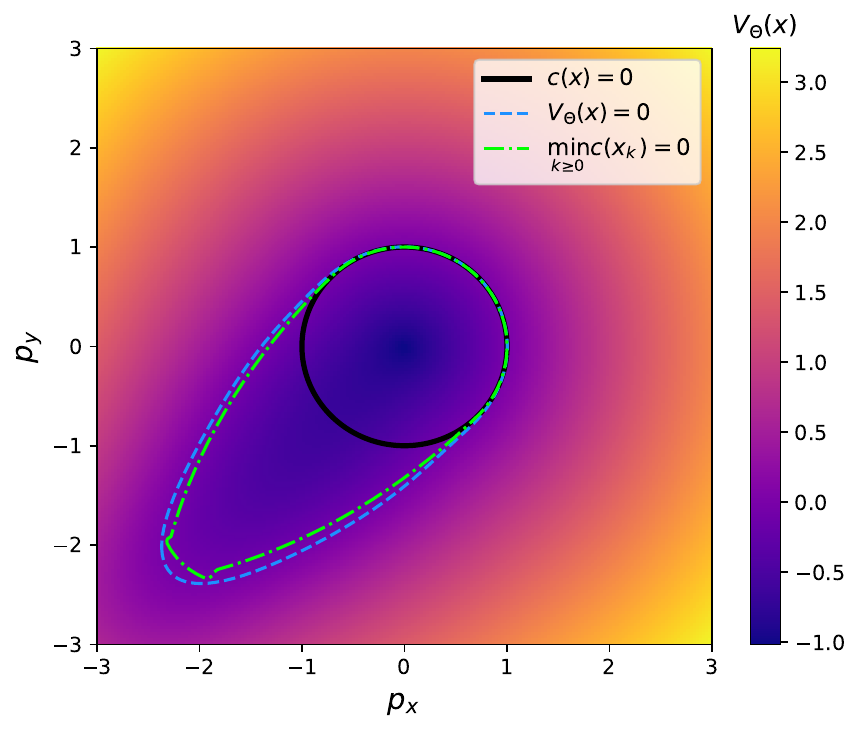}
  \caption{2D slice of the learned neural CBF for the 3D double integrator in the $(p_x, p_y)$ plane with $p_z = 0.0$ and velocity $(v_x, v_y, v_z) = (2.0, 2.0, 0.0)$.}
  \label{fig:slice_di3d}
\end{figure}

The state and control are
\begin{equation*}
  x = \begin{bmatrix} p_x & p_y & p_z & v_x & v_y & v_z \end{bmatrix}^\top,
  \qquad
  u = \begin{bmatrix} a_x & a_y & a_z \end{bmatrix}^\top,
\end{equation*}
where $(p_x, p_y, p_z)$ and $(v_x, v_y, v_z)$ are position and velocity in $\mathbb{R}^{3}$ and the controls are componentwise accelerations.  With $I_{3}$ and $0_{3}$ the $3\times 3$ identity and zero matrices,
\begin{equation*}
  f(x) = \begin{bmatrix} v_x \\ v_y \\ v_z \\ 0 \\ 0 \\ 0 \end{bmatrix},
  \qquad
  g(x) = \begin{bmatrix} 0_{3} \\ I_{3} \end{bmatrix}.
\end{equation*}
The constraint function is a spherical obstacle of radius $r$ centered at the origin:
\begin{equation*}
  c(x) = \sqrt{p_x^{2} + p_y^{2} + p_z^{2}} - r.
\end{equation*}
See \cref{tab:exp_di3d} for the configuration and \cref{fig:slice_di3d} for a 2D CBF slice.

\begin{table}[ht]
  \centering
  \caption{Configuration --- 3D Double Integrator.}
  \label{tab:exp_di3d}
  \footnotesize
  \begin{tabular}{@{}llr@{}}
    \toprule
    \multirow{5}{*}{Dynamics}
      & $x_{\min}$       & $[-3.0,\;-3.0,\;-3.0,\;-2.0,\;-2.0,\;-2.0]$ \\
      & $x_{\max}$       & $[\phantom{-}3.0,\;\phantom{-}3.0,\;\phantom{-}3.0,\;\phantom{-}2.0,\;\phantom{-}2.0,\;\phantom{-}2.0]$ \\
      & $u_{\min}$       & $[-1.0,\;-1.0,\;-1.0]$ \\
      & $u_{\max}$       & $[\phantom{-}1.0,\;\phantom{-}1.0,\;\phantom{-}1.0]$ \\
      & Model params     & --- \\
    \midrule
    \multirow{1}{*}{Constraint}
      & $r$              & $1.0$ \\
    \midrule
    \multirow{7}{*}{Data}
      & Method           & Stochastic beam search \\
      & \# samples       & $400{,}000$ \\
      & $K$              & $80$ \\
      & $B$              & $600$ \\
      & $\Delta t$       & $0.05$\,s \\
      & extra params     & Gumbel top-$k$, $T{=}0.05$ \\
      & $T_{\text{data}}$ & $4$\,m\,$12$\,s \\
    \midrule
    \multirow{2}{*}{Model}
      & Hidden layers    & $4\times 64$, $\sin$ \\
      & Output           & softplus ($\beta=10$) \\
    \midrule
    \multirow{6}{*}{Training}
      & Epochs           & $10{,}000$ \\
      & Learning rate    & $10^{-3}$ \\
      & LR schedule      & $\times 0.1$ at epoch $8{,}000$ \\
      & PDE \# samples   & $600{,}000$ \\
      & PDE weight       & normalized \\
      & $T_{\text{train}}$ & $9$\,m\,$40$\,s \\
    \midrule
    \multirow{4}{*}{Validation}
      & Horizon $T$      & $5.0$\,s \\
      & $\Delta t$       & $0.005$\,s \\
      & \# samples       & $1{,}000{,}000$ \\
      & $T_{\text{valid}}$ & $33$\,s \\
    \bottomrule
  \end{tabular}
\end{table}

% -----------------------------------------------------------------------------
\subsection{3-DOF Manipulator}
\label{app:manipulator_3dof}
% -----------------------------------------------------------------------------

\begin{figure}[ht]
  \centering
  \includegraphics[width=0.55\textwidth]{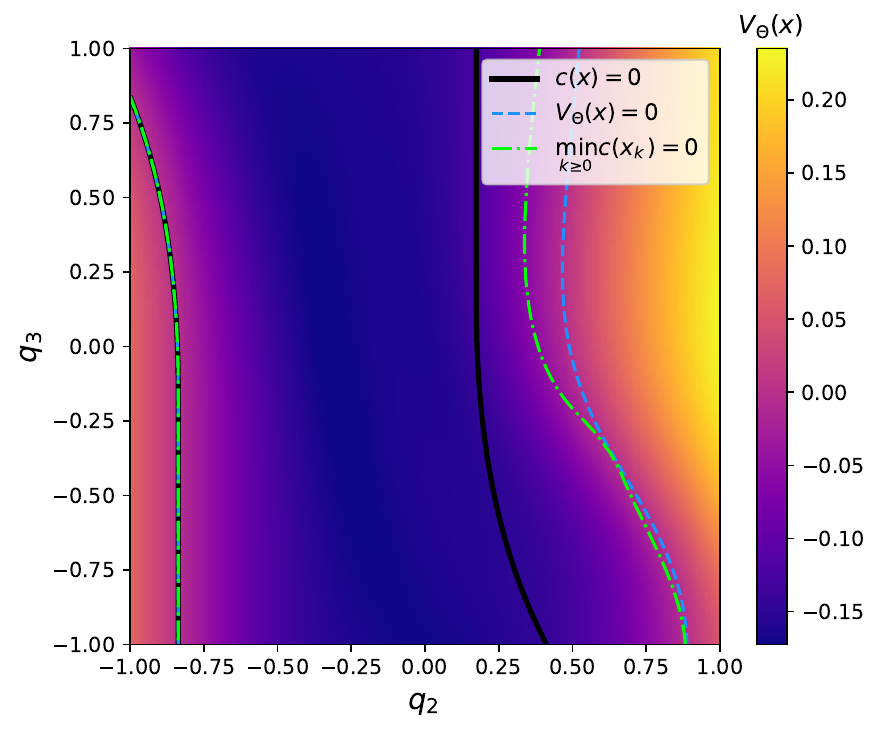}
  \caption{2D slice of the learned neural CBF for the 3-DOF manipulator in the $(q_2, q_3)$ plane with $q_1 = 0.25$\,rad and joint velocities $(\dot q_1, \dot q_2, \dot q_3) = (0.0, -5.0, 0.0)$\,rad/s.}
  \label{fig:slice_manipulator_3dof}
\end{figure}

This is a three-link spatial RRR arm with a shoulder yaw joint about world $+z$, a shoulder pitch joint about the local $+y$ axis, and an elbow pitch joint. At $q_1=q_2=q_3=0$ the arm is fully extended along world $+x$.  The state and control are
\begin{equation*}
  x = \begin{bmatrix} q_1 & q_2 & q_3 & \dot q_1 & \dot q_2 & \dot q_3 \end{bmatrix}^\top,
  \qquad
  u = \begin{bmatrix} \tau_1 & \tau_2 & \tau_3 \end{bmatrix}^\top.
\end{equation*}
The links are modeled as point masses $m_1$, $m_2$ at the centers of mass $l_{c1}$, $l_{c2}$ along the upper-arm ($l_1$) and forearm ($l_2$). Each joint carries a diagonal inertia $I_1, I_2, I_3$.  The manipulator equation $M(q)\ddot q + C(q,\dot q)\,\dot q + G(q) = \tau$ then has a non-zero pattern in which the shoulder-yaw row is dynamically decoupled from the two pitch rows, so
\begin{equation*}
  M(q) = \begin{bmatrix} M_{11}(q_2,q_3) & 0 & 0 \\ 0 & M_{22}(q_3) & M_{23}(q_3) \\ 0 & M_{23}(q_3) & M_{33} \end{bmatrix},
\end{equation*}
with (writing $r_c = l_1\cos q_2 + l_{c2}\cos(q_2{+}q_3)$ and $h_c = l_1\sin q_2 + l_{c2}\sin(q_2{+}q_3)$)
\begin{align*}
  M_{11} &= m_1 l_{c1}^{2}\cos^{2}q_2 + m_2\,r_c^{2} + I_1, \\
  M_{22} &= m_1 l_{c1}^{2} + m_2\!\left(l_1^{2} + l_{c2}^{2} + 2\,l_1 l_{c2}\cos q_3\right) + I_2, \\
  M_{33} &= m_2 l_{c2}^{2} + I_3, \quad
  M_{23}  = m_2\!\left(l_{c2}^{2} + l_1 l_{c2}\cos q_3\right).
\end{align*}
The Coriolis and gravity terms are
\begin{align*}
  (C\dot q)_1 &= -(m_1 l_{c1}^{2}\sin 2q_2 + 2 m_2 r_c h_c)\,\dot q_1\dot q_2
                 - 2 m_2 r_c l_{c2}\sin(q_2{+}q_3)\,\dot q_1\dot q_3, \\
  (C\dot q)_2 &= \tfrac{1}{2}m_1 l_{c1}^{2}\sin 2q_2\,\dot q_1^{2}
                 + m_2 r_c h_c\,\dot q_1^{2}
                 - 2 m_2 l_1 l_{c2}\sin q_3\,\dot q_2\dot q_3
                 - m_2 l_1 l_{c2}\sin q_3\,\dot q_3^{2}, \\
  (C\dot q)_3 &= m_2 r_c l_{c2}\sin(q_2{+}q_3)\,\dot q_1^{2}
                 + m_2 l_1 l_{c2}\sin q_3\,\dot q_2^{2}, \\
  G_1 &= 0, \quad
  G_2  = g\!\left((m_1 l_{c1} + m_2 l_1)\cos q_2 + m_2 l_{c2}\cos(q_2{+}q_3)\right), \\
  G_3 &= m_2 g\,l_{c2}\cos(q_2{+}q_3).
\end{align*}
The control-affine form is
\begin{equation*}
  f(x) = \begin{bmatrix} \dot q \\ -M^{-1}(C\dot q + G) \end{bmatrix},
  \qquad
  g(x) = \begin{bmatrix} 0_{3} \\ M^{-1}(q) \end{bmatrix},
\end{equation*}
with $M^{-1}$ available in closed form because of the block-diagonal structure. The constraint is a whole-arm collision check against a spherical obstacle at $c\in\mathbb{R}^{3}$ of radius $r$. Each link is modeled as a capsule of thickness $r_{\ell}$, and the safe-set SDF is the smooth minimum of the per-link capsule-vs-sphere distances,
\begin{equation*}
  c(x) = -\frac{1}{\alpha}\log\!\Bigl(
      \sum_{i=1}^{2} e^{-\alpha\,d_i(x)}
    \Bigr),
  \qquad
  d_i(x) = \operatorname{dist}\bigl(c,\,\text{link}_i(q)\bigr) - r - r_{\ell},
\end{equation*}
where $\operatorname{dist}(c, \text{link}_i)$ is the Euclidean distance from the obstacle center to the $i$-th link segment in world coordinates. See \cref{tab:exp_manipulator_3dof} for the configuration and \cref{fig:slice_manipulator_3dof} for a 2D CBF slice.

\begin{table}[ht]
  \centering
  \caption{Configuration --- 3-DOF Manipulator.}
  \label{tab:exp_manipulator_3dof}
  \footnotesize
  \begin{tabular}{@{}llr@{}}
    \toprule
    \multirow{5}{*}{Dynamics}
      & $x_{\min}$       & $[-1.0,\;-1.0,\;-1.0,\;-3.0,\;-6.0,\;-6.0]$ \\
      & $x_{\max}$       & $[\phantom{-}1.0,\;\phantom{-}1.0,\;\phantom{-}1.0,\;\phantom{-}3.0,\;\phantom{-}6.0,\;\phantom{-}6.0]$ \\
      & $u_{\min}$       & $[-10.0,\;-20.0,\;-10.0]$ \\
      & $u_{\max}$       & $[\phantom{-}10.0,\;\phantom{-}20.0,\;\phantom{-}10.0]$ \\
      & Model params     & \begin{tabular}[t]{@{}r@{}}
          $m_1{=}m_2{=}2.0$ \\
          $l_1{=}l_2{=}0.5$ \\
          $l_{c1}{=}l_{c2}{=}0.25$ \\
          $I_1{=}I_2{=}I_3{=}0.0417$ \\
          $g{=}9.81$
        \end{tabular} \\
    \midrule
    \multirow{4}{*}{Constraint}
      & Obstacle center  & $(0.6,\,0.0,\,-0.2)$ \\
      & Obstacle radius $r$ & $0.2$ \\
      & Link radius $r_{\ell}$ & $0.1$ \\
      & $\alpha$         & $20.0$ \\
    \midrule
    \multirow{7}{*}{Data}
      & Method           & Beam search \\
      & \# samples       & $600{,}000$ \\
      & $K$              & $60$ \\
      & $B$              & $1{,}000$ \\
      & $\Delta t$       & $0.05$\,s \\
      & extra params     & --- \\
      & $T_{\text{data}}$ & $28$\,m\,$49$\,s \\
    \midrule
    \multirow{2}{*}{Model}
      & Hidden layers    & $4\times 64$, $\sin$ \\
      & Output           & softplus ($\beta=10$) \\
    \midrule
    \multirow{6}{*}{Training}
      & Epochs           & $10{,}000$ \\
      & Learning rate    & $10^{-3}$ \\
      & LR schedule      & $\times 0.1$ at epoch $8{,}000$ \\
      & PDE \# samples   & $600{,}000$ \\
      & PDE weight       & $0.0$ (fixed) \\
      & $T_{\text{train}}$ & $3$\,m\,$17$\,s \\
    \midrule
    \multirow{4}{*}{Validation}
      & Horizon $T$      & $4.0$\,s \\
      & $\Delta t$       & $0.005$\,s \\
      & \# samples       & $1{,}000{,}000$ \\
      & $T_{\text{valid}}$ & $35$\,s \\
    \bottomrule
  \end{tabular}
\end{table}

% -----------------------------------------------------------------------------
\subsection{Landing Rocket}
\label{app:rocket}
% -----------------------------------------------------------------------------

\begin{figure}[ht]
  \centering
  \includegraphics[width=0.55\textwidth]{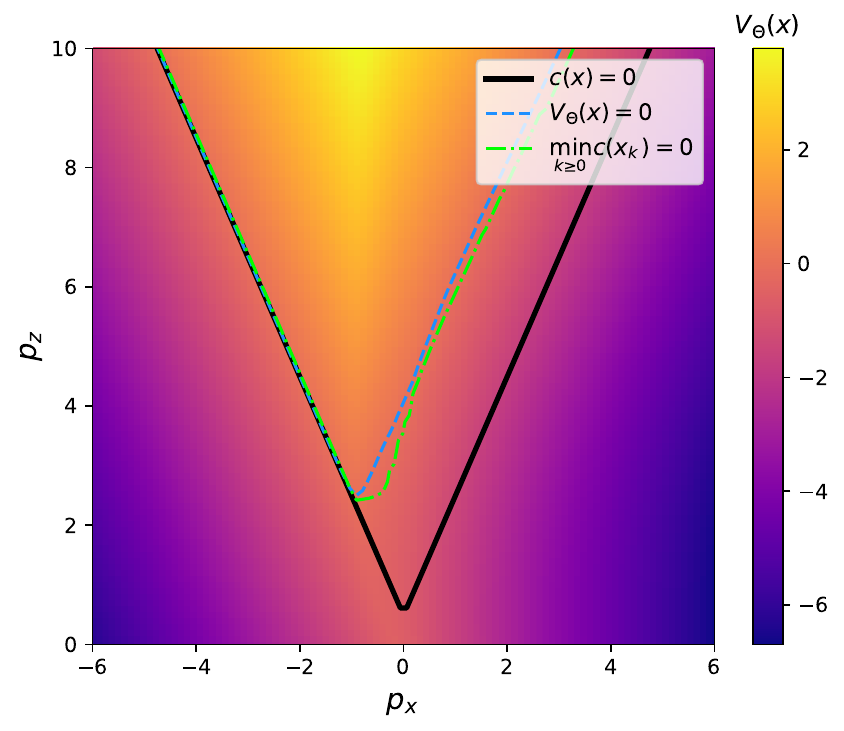}
  \caption{2D slice of the learned neural CBF for the landing rocket in the $(p_x, p_z)$ plane with velocity $(v_x, v_z) = (1.0, -6.0)$, pitch $\theta = -0.2$\,rad, angular rate $\omega = 0.0$, and mass $m = 0.9$.}
  \label{fig:slice_rocket}
\end{figure}

The state and control vectors are
\begin{equation*}
  x = \begin{bmatrix} p_x & p_z & v_x & v_z & \theta & \omega & m \end{bmatrix}^\top,
  \qquad
  u = \begin{bmatrix} T & \tau \end{bmatrix}^\top,
\end{equation*}
where $(p_x, p_z)$ is the planar position (the ground is at $p_z = 0$), $(v_x, v_z)$ the linear velocities, $\theta$ the pitch angle, $\omega$ the angular rate, $m$ the (time-varying) mass, $T \ge 0$ the thrust magnitude, and $\tau$ the body torque.  Denoting the rotational inertia by $I$ and the mass-flow coefficient by $\alpha$,
\begin{equation*}
  f(x) =
  \begin{bmatrix}
    v_x \\ v_z \\ 0 \\ -g \\ \omega \\ 0 \\ 0
  \end{bmatrix},
  \qquad
  g(x) =
  \begin{bmatrix}
    0          & 0 \\
    0          & 0 \\
   -\sin\theta/m & 0 \\
    \cos\theta/m & 0 \\
    0          & 0 \\
    0          & 1/I \\
   -\alpha     & 0
  \end{bmatrix}.
\end{equation*}
The constraint function is a smooth landing funnel that couples the lateral position to altitude and penalizes high translational velocity and large tilt.  With pad half-width $\rho$, funnel slope $s$, velocity weight $w_v$, tilt weight $w_\theta$, and smoothing $\varepsilon$,
\begin{equation*}
  c(x) = (\rho + s\,p_z) - \sqrt{p_x^{2} + \varepsilon}
         - w_v(v_x^{2} + v_z^{2}) - w_\theta\,\theta^{2}.
\end{equation*}
See \cref{tab:exp_rocket} for the configuration and \cref{fig:slice_rocket} for a 2D CBF slice.

\begin{table}[ht]
  \centering
  \caption{Configuration --- Landing Rocket.}
  \label{tab:exp_rocket}
  \footnotesize
  \begin{tabular}{@{}llr@{}}
    \toprule
    \multirow{5}{*}{Dynamics}
      & $x_{\min}$       & $[-6.0,\;\phantom{-}0.0,\;-3.0,\;-6.0,\;-1.0,\;-3.0,\;\phantom{-}0.6]$ \\
      & $x_{\max}$       & $[\phantom{-}6.0,\;10.0,\;\phantom{-}3.0,\;\phantom{-}3.0,\;\phantom{-}1.0,\;\phantom{-}3.0,\;\phantom{-}1.0]$ \\
      & $u_{\min}$       & $[\phantom{-}0.0,\;-1.0]$ \\
      & $u_{\max}$       & $[15.0,\;\phantom{-}1.0]$ \\
      & Model params     & $I{=}0.25$,\; $\alpha{=}0.01$,\; $g{=}9.81$ \\
    \midrule
    \multirow{5}{*}{Constraint}
      & $\rho$               & $0.5$ \\
      & $s$                  & $0.5$ \\
      & $w_v$                & $0.02$ \\
      & $w_\theta$           & $0.1$ \\
      & $\varepsilon$        & $10^{-4}$ \\
    \midrule
    \multirow{7}{*}{Data}
      & Method           & Stochastic beam search \\
      & \# samples       & $600{,}000$ \\
      & $K$              & $40$ \\
      & $B$              & $1{,}500$ \\
      & $\Delta t$       & $0.1$\,s \\
      & extra params     & Gumbel top-$k$, $T{=}0.2$ \\
      & $T_{\text{data}}$ & $6$\,m\,$55$\,s \\
    \midrule
    \multirow{2}{*}{Model}
      & Hidden layers    & $4\times 84$, $\sin$ \\
      & Output           & softplus ($\beta=10$) \\
    \midrule
    \multirow{6}{*}{Training}
      & Epochs           & $10{,}000$ \\
      & Learning rate    & $10^{-3}$ \\
      & LR schedule      & $\times 0.1$ at epoch $8{,}000$ \\
      & PDE \# samples   & $800{,}000$ \\
      & PDE weight       & $0.2$ (fixed) \\
      & $T_{\text{train}}$ & $18$\,m\,$2$\,s \\
    \midrule
    \multirow{4}{*}{Validation}
      & Horizon $T$      & $5.0$\,s \\
      & $\Delta t$       & $0.005$\,s \\
      & \# samples       & $1{,}000{,}000$ \\
      & $T_{\text{valid}}$ & $47$\,s \\
    \bottomrule
  \end{tabular}
\end{table}

% -----------------------------------------------------------------------------
\subsection{Quadruped Trunk}
\label{app:quadruped_trunk}
% -----------------------------------------------------------------------------

\begin{figure}[ht]
  \centering
  \includegraphics[width=0.55\textwidth]{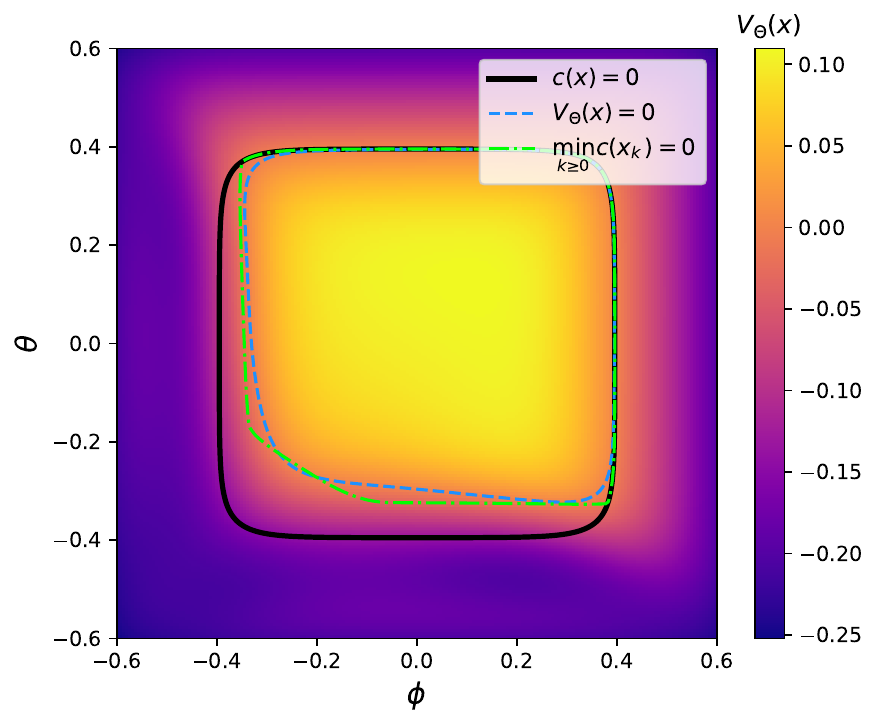}
  \caption{2D slice of the learned neural CBF for the quadruped trunk in the $(\phi, \theta)$ plane with trunk height $p_z = 0.3$, linear velocity $(v_x, v_y, v_z) = (0.0, 0.0, 0.0)$, and body-frame angular velocity $(p, q, r) = (-4.0, -4.0, 0.0)$.}
  \label{fig:slice_quadruped_trunk}
\end{figure}

This is a centroidal trunk model of a quadruped in quasi-static all-four-feet stance.  The uncontrolled world planar position and yaw are dropped, and contact forces enter only along body $+z$ at four nominal foot positions $r_i = (\pm a,\, \pm b,\, *)$ in (FL, FR, RL, RR) order. The state and control are
\begin{equation*}
  x = \begin{bmatrix} p_z & \phi & \theta & v_x & v_y & v_z & p & q & r \end{bmatrix}^\top,
  \qquad
  u = \begin{bmatrix} F_1 & F_2 & F_3 & F_4 \end{bmatrix}^\top,
\end{equation*}
where $p_z$ is the trunk height, $(\phi,\theta)$ are roll and pitch, $(v_x, v_y, v_z)$ the world-frame linear velocity, $(p,q,r)$ the body-frame angular velocity, and $F_i \ge 0$ the body-frame normal contact force at foot $i$.  Writing $F_\Sigma = F_1+F_2+F_3+F_4$, the dynamics are
\begin{equation*}
  f(x) =
  \begin{bmatrix}
    v_z \\
    p + \sin\phi\tan\theta\,q + \cos\phi\tan\theta\,r \\
    \cos\phi\,q - \sin\phi\,r \\
    0 \\ 0 \\ -g \\[2pt]
    -(I_z - I_y)/I_x\;q r \\
    -(I_x - I_z)/I_y\;p r \\
    -(I_y - I_x)/I_z\;p q
  \end{bmatrix},
\end{equation*}
\begin{equation*}
  g(x) =
  \begin{bmatrix}
    0          & 0          & 0          & 0 \\
    0          & 0          & 0          & 0 \\
    0          & 0          & 0          & 0 \\
    s_\theta c_\phi/m & s_\theta c_\phi/m & s_\theta c_\phi/m & s_\theta c_\phi/m \\
   -s_\phi/m   & -s_\phi/m  & -s_\phi/m  & -s_\phi/m \\
    c_\theta c_\phi/m & c_\theta c_\phi/m & c_\theta c_\phi/m & c_\theta c_\phi/m \\
    \phantom{-}b/I_x  & -b/I_x          & \phantom{-}b/I_x  & -b/I_x \\
   -a/I_y      & -a/I_y     & \phantom{-}a/I_y & \phantom{-}a/I_y \\
    0          & 0          & 0          & 0
  \end{bmatrix},
\end{equation*}
where $(s_\phi, c_\phi) = (\sin\phi, \cos\phi)$ and similarly for $\theta$.  The yaw-rate row is uncontrolled in this contact mode but retained so its gyroscopic coupling into roll/pitch is preserved.  The constraint is a smooth box constraint on trunk height, roll, and pitch that prevents the trunk from falling:
\begin{equation*}
  c(x) = -\frac{1}{\alpha}\log\!\Bigl(
      e^{-\alpha\,d_{p_z}(x)}
    + e^{-\alpha\,d_{\phi}(x)}
    + e^{-\alpha\,d_{\theta}(x)}
    \Bigr),
\end{equation*}
with $d_{p_z}(x) = \min(p_z - p_z^{\min},\; p_z^{\max} - p_z)$, and analogously $d_{\phi}(x) = |\phi_{\max}| - |\phi|$, $d_{\theta}(x) = |\theta_{\max}| - |\theta|$. See \cref{tab:exp_quadruped_trunk} for the configuration and \cref{fig:slice_quadruped_trunk} for a 2D CBF slice.

\begin{table}[ht]
  \centering
  \caption{Configuration --- Quadruped Trunk.}
  \label{tab:exp_quadruped_trunk}
  \footnotesize
  \begin{tabular}{@{}llr@{}}
    \toprule
    \multirow{5}{*}{Dynamics}
      & $x_{\min}$       & $[\phantom{-}0.05,\;-0.6,\;-0.6,\;-1.5,\;-1.5,\;-2.0,\;-5.0,\;-5.0,\;-2.0]$ \\
      & $x_{\max}$       & $[\phantom{-}0.55,\;\phantom{-}0.6,\;\phantom{-}0.6,\;\phantom{-}1.5,\;\phantom{-}1.5,\;\phantom{-}2.0,\;\phantom{-}5.0,\;\phantom{-}5.0,\;\phantom{-}2.0]$ \\
      & $u_{\min}$       & $[\phantom{-}0.0,\;\phantom{-}0.0,\;\phantom{-}0.0,\;\phantom{-}0.0]$ \\
      & $u_{\max}$       & $[60.0,\;60.0,\;60.0,\;60.0]$ \\
      & Model params     & \begin{tabular}[t]{@{}r@{}}
          $m{=}12.0$ \\
          $I_x{=}0.10,\;I_y{=}0.20,\;I_z{=}0.25$ \\
          $a{=}0.20,\;b{=}0.15$ \\
          $g{=}9.81$
        \end{tabular} \\
    \midrule
    \multirow{4}{*}{Constraint}
      & $[p_z^{\min}, p_z^{\max}]$ & $[0.18,\,0.42]$ \\
      & $\phi_{\max}$    & $0.4$ \\
      & $\theta_{\max}$  & $0.4$ \\
      & $\alpha$         & $20.0$ \\
    \midrule
    \multirow{7}{*}{Data}
      & Method           & Stochastic beam search \\
      & \# samples       & $600{,}000$ \\
      & $K$              & $100$ \\
      & $B$              & $1{,}000$ \\
      & $\Delta t$       & $0.02$\,s \\
      & extra params     & Gumbel top-$k$, $T{=}0.05$ \\
      & $T_{\text{data}}$ & $1$\,h\,$41$\,m\,$32$\,s \\
    \midrule
    \multirow{2}{*}{Model}
      & Hidden layers    & $5\times 84$, $\sin$ \\
      & Output           & softplus ($\beta=10$) \\
    \midrule
    \multirow{6}{*}{Training}
      & Epochs           & $10{,}000$ \\
      & Learning rate    & $10^{-3}$ \\
      & LR schedule      & $\times 0.1$ at epoch $8{,}000$ \\
      & PDE \# samples   & $600{,}000$ \\
      & PDE weight       & $0.0$ (fixed) \\
      & $T_{\text{train}}$ & $5$\,m\,$28$\,s \\
    \midrule
    \multirow{4}{*}{Validation}
      & Horizon $T$      & $3.0$\,s \\
      & $\Delta t$       & $0.002$\,s \\
      & \# samples       & $1{,}000{,}000$ \\
      & $T_{\text{valid}}$ & $1$\,m\,$38$\,s \\
    \bottomrule
  \end{tabular}
\end{table}

% -----------------------------------------------------------------------------
\subsection{6-DoF Underwater Vehicle}
\label{app:auv_6dof}
% -----------------------------------------------------------------------------

\begin{figure}[ht]
  \centering
  \includegraphics[width=0.55\textwidth]{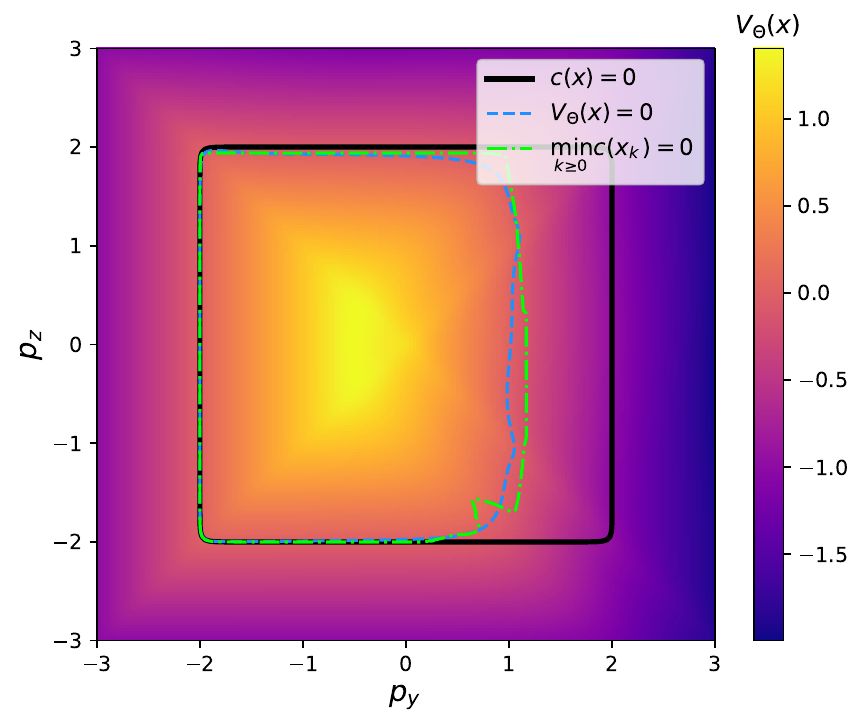}
  \caption{2D slice of the learned neural CBF for the 6-DoF underwater vehicle in the $(p_y, p_z)$ plane with world position $p_x = 0.0$, orientation $(\phi, \theta, \psi) = (0.0, -1.0, 1.0)$\,rad, body-frame linear velocity $(u_b, v_b, w_b) = (1.8, 0.0, -1.8)$, and body-frame angular velocity $(p, q, r) = (0.0, 0.0, 0.0)$.}
  \label{fig:slice_auv_6dof}
\end{figure}

We model a neutrally-buoyant 6-DoF AUV in a rigid-body Fossen-style formulation. Added mass is absorbed into a single isotropic translational mass $m_t$, the inertia is diagonal in the principal axes, and the damping is linear+quadratic.  The world frame is ENU and the body frame is $x$ forward, $y$ left, $z$ up.  The state is partitioned into the world-frame pose $\eta$ and the body-frame twist $\nu$,
\begin{equation*}
  x = \begin{bmatrix} \eta^\top & \nu^\top \end{bmatrix}^\top,
  \quad
  \eta = \begin{bmatrix} p_x & p_y & p_z & \phi & \theta & \psi \end{bmatrix}^\top,
  \quad
  \nu = \begin{bmatrix} u_b & v_b & w_b & p & q & r \end{bmatrix}^\top,
\end{equation*}
\begin{equation*}
  u = \begin{bmatrix} F_x & F_y & F_z & \tau_x & \tau_y & \tau_z \end{bmatrix}^\top,
\end{equation*}
where $(p_x, p_y, p_z)$ is world position, $(\phi,\theta,\psi)$ are ZYX Euler angles (roll, pitch, yaw), $(u_b, v_b, w_b)$ the body-frame linear velocity, $(p, q, r)$ the body-frame angular velocity, and $(F_i, \tau_i)$ the body-frame force/moment after thruster allocation.  The kinematics use the standard transformation $\dot\eta = J(\eta)\,\nu$:
\begin{equation*}
  \begin{bmatrix} \dot p_x \\ \dot p_y \\ \dot p_z \end{bmatrix} = R(\phi,\theta,\psi)\begin{bmatrix} u_b \\ v_b \\ w_b \end{bmatrix},
  \quad
  \begin{bmatrix} \dot\phi \\ \dot\theta \\ \dot\psi \end{bmatrix} =
  \begin{bmatrix}
    1 & \sin\phi\tan\theta & \cos\phi\tan\theta \\
    0 & \cos\phi          & -\sin\phi \\
    0 & \sin\phi/\cos\theta & \cos\phi/\cos\theta
  \end{bmatrix}
  \begin{bmatrix} p \\ q \\ r \end{bmatrix},
\end{equation*}
where $R$ is the body-to-world rotation.  The body-frame dynamics combine Coriolis terms, hydrodynamic damping $F_{\text{drag},i}(\nu_i) = -(d^{\ell}_i + d^{q}_i|\nu_i|)\nu_i$, and a self-righting restoring moment from the center-of-buoyancy offset $h_{\text{cb}}$ above the center of gravity:
\begin{equation*}
  \begin{aligned}
    m_t\dot u_b &= F_x - m_t(q w_b - r v_b) - (d^{\ell}_u + d^{q}_u|u_b|)\,u_b, \\
    m_t\dot v_b &= F_y - m_t(r u_b - p w_b) - (d^{\ell}_v + d^{q}_v|v_b|)\,v_b, \\
    m_t\dot w_b &= F_z - m_t(p v_b - q u_b) - (d^{\ell}_w + d^{q}_w|w_b|)\,w_b, \\
    I_x\dot p &= \tau_x - (I_z - I_y)\,q r - h_{\text{cb}} B \cos\theta\sin\phi - (d^{\ell}_p + d^{q}_p|p|)\,p, \\
    I_y\dot q &= \tau_y - (I_x - I_z)\,p r - h_{\text{cb}} B \sin\theta \;\;\;\;\;\;\;\;\;\,- (d^{\ell}_q + d^{q}_q|q|)\,q, \\
    I_z\dot r &= \tau_z - (I_y - I_x)\,p q \;\;\;\;\;\;\;\;\;\;\;\;\;\;\;\;\;\;\;\;\;\;\;\,- (d^{\ell}_r + d^{q}_r|r|)\,r,
  \end{aligned}
\end{equation*}
with $B = m_t g$ the buoyancy magnitude (neutrally buoyant).  Gravity and buoyancy cancel in the translational equations because the centers of gravity and buoyancy share the same body-$z$ line.  Collecting these into control-affine form gives a drift $f(x)$ assembling the kinematics and the $u{=}0$ body-frame dynamics above, and a constant $g(x) = [0_{6\times 6};\; \operatorname{diag}(1/m_t, 1/m_t, 1/m_t, 1/I_x, 1/I_y, 1/I_z)]$ input matrix.  The constraint confines the vehicle to a tunnel of half-width $r_t$ in the world $(p_y, p_z)$ plane, implemented as a smooth state-limits SDF
\begin{equation*}
  c(x) = -\frac{1}{\alpha}\log\!\Bigl(
      e^{-\alpha (p_y + r_t)} + e^{-\alpha (r_t - p_y)}
    + e^{-\alpha (p_z + r_t)} + e^{-\alpha (r_t - p_z)}
    \Bigr).
\end{equation*}
See \cref{tab:exp_auv_6dof} for the configuration and \cref{fig:slice_auv_6dof} for a 2D CBF slice.

\begin{table}[ht]
  \centering
  \caption{Configuration --- 6-DoF Underwater Vehicle.}
  \label{tab:exp_auv_6dof}
  \footnotesize
  \begin{tabular}{@{}llr@{}}
    \toprule
    \multirow{5}{*}{Dynamics}
      & $x_{\min}$       & $[-3,\,-3,\,-3,\,-\pi,\,-1,\,-\pi,\,-2,\,-2,\,-2,\,-1.5,\,-1.5,\,-1.5]$ \\
      & $x_{\max}$       & $[\phantom{-}3,\,\phantom{-}3,\,\phantom{-}3,\,\phantom{-}\pi,\,\phantom{-}1,\,\phantom{-}\pi,\,\phantom{-}2,\,\phantom{-}2,\,\phantom{-}2,\,\phantom{-}1.5,\,\phantom{-}1.5,\,\phantom{-}1.5]$ \\
      & $u_{\min}$       & $[-80,\,-10,\,-10,\,-20,\,-20,\,-20]$ \\
      & $u_{\max}$       & $[\phantom{-}80,\,\phantom{-}10,\,\phantom{-}10,\,\phantom{-}20,\,\phantom{-}20,\,\phantom{-}20]$ \\
      & Model params     & \begin{tabular}[t]{@{}r@{}}
          $m_t{=}30.0$ \\
          $I_x{=}0.5,\;I_y{=}2.0,\;I_z{=}2.0$ \\
          $d^{\ell}{=}(3,3,3,0.3,0.5,0.5)$ \\
          $d^{q}{=}(5,5,5,0.3,1,1)$ \\
          $h_{\text{cb}}{=}0.02,\;g{=}9.81$
        \end{tabular} \\
    \midrule
    \multirow{2}{*}{Constraint}
      & $r_t$            & $2.0$ \\
      & $\alpha$         & $20.0$ \\
    \midrule
    \multirow{7}{*}{Data}
      & Method           & Stochastic beam search \\
      & \# samples       & $800{,}000$ \\
      & $K$              & $40$ \\
      & $B$              & $800$ \\
      & $\Delta t$       & $0.1$\,s \\
      & extra params     & $T{=}0.08$ \\
      & $T_{\text{data}}$ & $3$\,h\,$7$\,m\,$6$\,s \\
    \midrule
    \multirow{2}{*}{Model}
      & Hidden layers    & $3\times 128$, $\sin$ \\
      & Output           & softplus ($\beta=10$) \\
    \midrule
    \multirow{6}{*}{Training}
      & Epochs           & $10{,}000$ \\
      & Learning rate    & $10^{-3}$ \\
      & LR schedule      & $\times 0.1$ at epoch $8{,}000$ \\
      & PDE \# samples   & $800{,}000$ \\
      & PDE weight       & $0.0$ (fixed) \\
      & $T_{\text{train}}$ & $4$\,m\,$58$\,s \\
    \midrule
    \multirow{4}{*}{Validation}
      & Horizon $T$      & $4.0$\,s \\
      & $\Delta t$       & $0.005$\,s \\
      & \# samples       & $1{,}000{,}000$ \\
      & $T_{\text{valid}}$ & $52$\,s \\
    \bottomrule
  \end{tabular}
\end{table}

% -----------------------------------------------------------------------------
\subsection{Quadrotor}
\label{app:quadrotor}
% -----------------------------------------------------------------------------

\begin{figure}[ht]
  \centering
  \includegraphics[width=0.55\textwidth]{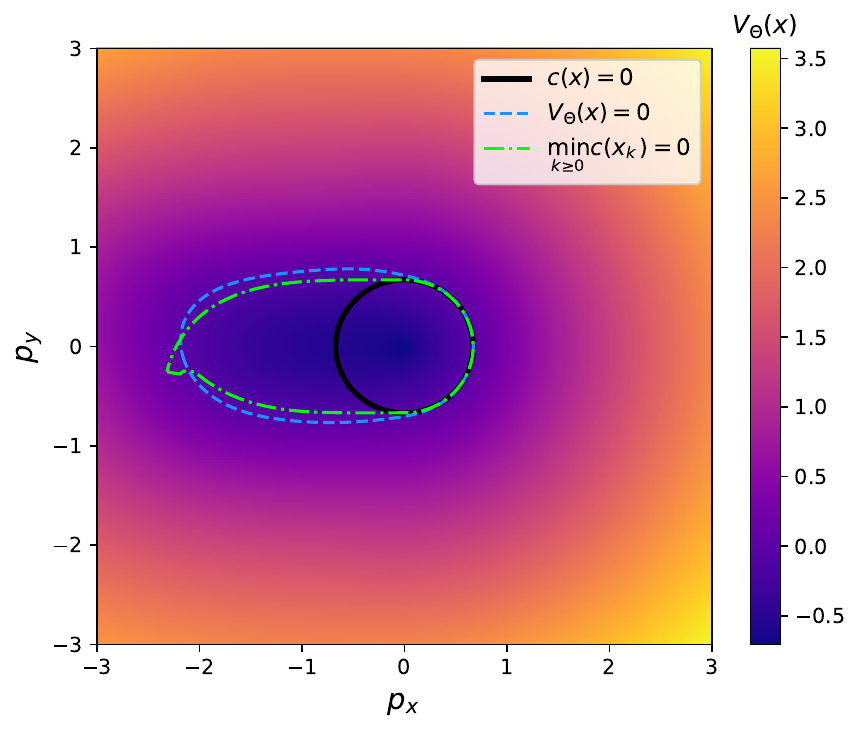}
  \caption{2D slice of the learned neural CBF for the quadrotor in the $(p_x, p_y)$ plane with altitude $p_z = 0.0$, orientation quaternion $q = (0, 0, 0, 1)$, linear velocity $(v_x, v_y, v_z) = (4.0, 0.0, 0.0)$, and body-frame angular velocity $(\omega_x, \omega_y, \omega_z) = (0.0, 0.0, 0.0)$.}
  \label{fig:slice_quadrotor}
\end{figure}

The 13-dimensional state and 4-dimensional control are
\begin{equation*}
  x = \begin{bmatrix} p^\top & q^\top & v^\top & \omega^\top \end{bmatrix}^\top,
  \qquad
  u = \begin{bmatrix} F & \alpha_x & \alpha_y & \alpha_z \end{bmatrix}^\top,
\end{equation*}
where $p\in\mathbb{R}^{3}$ is the world-frame position, $q = [q_w, q_x, q_y, q_z]^\top$ is the body-to-world unit quaternion, $v\in\mathbb{R}^{3}$ the world-frame linear velocity, and $\omega\in\mathbb{R}^{3}$ the body-frame angular velocity.  The control comprises the total thrust $F \ge 0$ and the three body-frame angular accelerations.  Let $R_{xz}(q) = 2(q_w q_y + q_x q_z)$, $R_{yz}(q) = 2(-q_w q_x + q_y q_z)$, and $R_{zz}(q) = 1 - 2(q_x^{2} + q_y^{2})$ denote the third column of the body-to-world rotation matrix, and let
\begin{equation*}
  H(q)\,\omega =
  \tfrac{1}{2}
  \begin{bmatrix}
    -q_x\omega_x - q_y\omega_y - q_z\omega_z \\
    \phantom{-}q_w\omega_x + q_y\omega_z - q_z\omega_y \\
    \phantom{-}q_w\omega_y - q_x\omega_z + q_z\omega_x \\
    \phantom{-}q_w\omega_z + q_x\omega_y - q_y\omega_x
  \end{bmatrix}
\end{equation*}
denote the quaternion kinematics. With the gyroscopic ratio $\gamma = 5/9$ inherited from the inertia tensor used in our experiments,
\begin{equation*}
  f(x) =
  \begin{bmatrix}
    v \\[2pt]
    H(q)\,\omega \\[2pt]
    0 \\ 0 \\ -g \\[2pt]
    -\gamma\,\omega_y\omega_z \\
    \phantom{-}\gamma\,\omega_x\omega_z \\
    0
  \end{bmatrix},
  \qquad
  g(x) =
  \begin{bmatrix}
    0_{3\times 1}     & 0_{3\times 3} \\
    0_{4\times 1}     & 0_{4\times 3} \\
    R_{xz}(q)/m       & 0_{1\times 3} \\
    R_{yz}(q)/m       & 0_{1\times 3} \\
    R_{zz}(q)/m       & 0_{1\times 3} \\
    0_{3\times 1}     & I_{3}
  \end{bmatrix}.
\end{equation*}
The constraint function is a vertical cylindrical obstacle of radius $r$ about the world $z$-axis through the origin:
\begin{equation*}
  c(x) = \sqrt{p_x^{2} + p_y^{2}} - r.
\end{equation*}
See \cref{tab:exp_quadrotor} for the configuration and \cref{fig:slice_quadrotor} for a 2D CBF slice.

\begin{table}[ht]
  \centering
  \caption{Configuration --- Quadrotor.}
  \label{tab:exp_quadrotor}
  \footnotesize
  \begin{tabular}{@{}llr@{}}
    \toprule
    \multirow{5}{*}{Dynamics}
      & $x_{\min}$       & $[-3\!\cdot\!\mathbf{1}_3,\;-\mathbf{1}_4,\;-5\!\cdot\!\mathbf{1}_3,\;-5\!\cdot\!\mathbf{1}_3]$ \\
      & $x_{\max}$       & $[\phantom{-}3\!\cdot\!\mathbf{1}_3,\;\phantom{-}\mathbf{1}_4,\;\phantom{-}5\!\cdot\!\mathbf{1}_3,\;\phantom{-}5\!\cdot\!\mathbf{1}_3]$ \\
      & $u_{\min}$       & $[\phantom{-}0.0,\;-8.0,\;-8.0,\;-4.0]$ \\
      & $u_{\max}$       & $[20.0,\;\phantom{-}8.0,\;\phantom{-}8.0,\;\phantom{-}4.0]$ \\
      & Model params     & $m{=}1.0$,\; $C_T{=}1.0$,\; $g{=}9.81$ \\
    \midrule
    \multirow{1}{*}{Constraint}
      & $r$              & $0.67$ \\
    \midrule
    \multirow{7}{*}{Data}
      & Method           & Branch-and-bound \\
      & \# samples       & $1{,}000{,}000$ \\
      & $K$              & $30$ \\
      & $B$              & $150$ \\
      & $\Delta t$       & $0.1$\,s \\
      & extra params     & $n_{\mathrm{restarts}}=2$ \\
      & $T_{\text{data}}$ & $24$\,m\,$41$\,s \\
    \midrule
    \multirow{2}{*}{Model}
      & Hidden layers    & $4\times 128$, $\sin$ \\
      & Output           & softplus ($\beta=10$) \\
    \midrule
    \multirow{6}{*}{Training}
      & Epochs           & $10{,}000$ \\
      & Learning rate    & $10^{-3}$ \\
      & LR schedule      & $\times 0.1$ at epoch $9{,}000$ \\
      & PDE \# samples   & $2{,}000{,}000$ \\
      & PDE weight       & $0.15$ (fixed) \\
      & $T_{\text{train}}$ & $54$\,m\,$8$\,s \\
    \midrule
    \multirow{4}{*}{Validation}
      & Horizon $T$      & $4.0$\,s \\
      & $\Delta t$       & $0.01$\,s \\
      & \# samples       & $2{,}000{,}000$ \\
      & $T_{\text{valid}}$ & $55$\,s \\
    \bottomrule
  \end{tabular}
\end{table}

\end{document}